\documentclass[letterpaper, 10 pt, journal, twoside]{IEEEtran}

\IEEEoverridecommandlockouts                              

\usepackage{graphicx}
\graphicspath{{figs/}}
\usepackage{amsmath}
\usepackage{amssymb}
\usepackage{algorithm}
\usepackage{algpseudocode}
\usepackage{times}
\usepackage{multirow}
\usepackage{fourier}
\usepackage{array}
\usepackage{makecell}
\newcommand{\figref}[1]{Fig.~\ref{#1}}
\newcommand{\tabref}[1]{Table~\ref{#1}}

\newcommand{\argmin}{\operatornamewithlimits{argmin}}
\newcommand{\bm}[1]{\mbox{\boldmath{$#1$}}}

\title{Optimization-based Posture Generation for Whole-body Contact Motion by Contact Point Search on the Body Surface}

\author{Masaki Murooka$^{1}$, Kei Okada$^{1}$ and Masayuki Inaba$^{1}$%
  \thanks{Manuscript received: September, 8, 2019; Revised December, 29, 2019; Accepted February, 2, 2020.}
  \thanks{This paper was recommended for publication by Editor Hong Liu upon evaluation of the Associate Editor and Reviewers' comments.
    This work was supported by JSPS KAKENHI Grant Number JP19K20371.} 
  \thanks{$^{1}$M. Murooka, K. Okada and M. Inaba are with Department of Mechano-Infomatics,
    The University of Tokyo, 7-3-1 Hongo, Bunkyo-ku, Tokyo 113-8656, Japan
    {\tt\footnotesize murooka at jsk.t.u-tokyo.ac.jp}}%
  \thanks{Digital Object Identifier (DOI): see top of this page.}
}

\begin{document}

\maketitle

\begin{abstract}
Whole-body contact is an effective strategy for improving the stability and efficiency of the motion of robots.
For robots to automatically perform such motions,
we propose a posture generation method that employs all available surfaces of the robot links.
By representing the contact point on the body surface by two-dimensional configuration variables,
the joint positions and contact points are simultaneously determined through a gradient-based optimization.
By generating motions with the proposed method,
we present experiments in which robots manipulate objects effectively utilizing whole-body contact.
\end{abstract}


\newcommand\myswlang[2]{%
  \ifx\mypaperlang\empty%
#1%
  \else%
#2%
  \fi%
}
\def\mypaperlang{} 

\setlength{\floatsep}{4pt}
\setlength{\textfloatsep}{4pt}
\setlength{\abovecaptionskip}{2pt}

\section{Introduction}

\myswlang{%
  \IEEEPARstart{R}{obots}
  must perform dexterous and high-load operations autonomously
  to expand the range of applications to daily life support and disaster response.
  In the robot motions performed so far, the robots usually contact with the environment only through certain body points (e.g., gripper and foot),
  making the motions unstable and inefficient.

  In this study, to solve such a problem, we propose a method to generate motion through contact at any point on the robot's body surface.
  This is a versatile method
  in which the contact points on the body surface are expressed as configuration variables
  and searched together with joint positions by the conventional calculation procedure of inverse kinematics.
  We show the effectiveness of the proposed method by applying to a variety of whole-body contact motions.
}{%
  ロボットは，産業や生活の様々な場面で，繊細かつ高負荷な動作を自律して行うことが求められている．
  このために，ロボットが抱え上げや膝立ちなどのように，全身を接触させるような動作は重要である．
  全身接触動作を自律性に実現させるためには，
  ロボットや物体，環境の幾何モデルが与えられた時に，
  接触点を自動的に決定できることが，キー技術である．
  本研究では，このような，
  接触点を事前に指定しなくても自動的に決めることができるような，
  全身動作生成法を提案し，ヒューマノイドの動作生成に適用する．
  従来は，動作の指定には，
  IKで代表されるように，point-to-pointで与えられていることが多いが，
  これは人間にとって手間であり，さらに探索空間が狭くなり解が見つからなくなることが問題である．
  本論文により，
  従来のpoint-to-pointの目標で定義されていた動作生成の問題を，
  body-to-bodyの目標で定義することで，
  動作生成でユーザが指定する情報が削減され，
  全身接触動作生成のための，新たな動作生成が可能になると言える．
}%

\subsection{Related Works}
\subsubsection{Whole-body Contact Motion by Robot}
\myswlang{%
  Whole-body contact motions, that effectively use the robot's body, have been widely studied.
  Whole-body manipulation \cite{WholebodyPush:Murooka:ICRA2015} and getting up \cite{GetUp:Kanehiro:ICRA2003} are typical motions and realized by real robots.
  To generate these motions,
  the key postures are taught online \cite{Holding:Mittendorfer:AR2015}
  or the contact points are predetermined manually \cite{RobustHumanoid:Kakiuchi:IROS2017}.
  Many of these approaches follow human intervention or ad hoc rules,
  and the general method for automatically generating whole-body contact motion as proposed in this study is yet to be established.
  The proposed method is also effective in utilizing the whole-body contacts in the multi-contact motion of humanoid robots,
  which has been widely studied in recent years~\cite{MulticontactMotion:Bouyarmane:Humanoid2018}\cite{MulticontactTorque:Werner:IROS2016}.
}{%
  全身接触動作は，
  ロボットの多自由度を有効に活かすことが可能であり，
  広く研究されている．
  抱え上げ\cite{}や起き上がり\cite{}はその代表的な動作である．
  これらの研究では，
  センサフィードバックによって自動的に作られたり\cite{}，
  contact pointを指定していたりする\cite{}．
  また，多指ハンドによるパワーグラスプ\cite{}も同様の問題として捉えることができ，
  ***による動作計画法などが提案されている\cite{}．
  このように，
  多くの先行研究では，
  全身接触動作における  接触点の探索可能に，アドホックな手法を用いており，
  一般的な手法は未だ確立されていない．
  本研究では，これを可能にし，
  マニピュレーションからロコモーションまで，様々な全身接触動作を実現する．
  近年，盛んに研究されているヒューマノイドのマルチコンタクト動作\cite{}\cite{}において，
  全身の接触を活用するためにも，本論文の手法は有効である．
}%


\subsubsection{Grasp Planning of Multi-Fingered Hand}
\myswlang{%
  Power grasp \cite{WAMS:Salisbury:ICRA1988} with a multi-fingered hand belongs to the same kind of motion because the contact points are selected from the finger surface.
  In many of the grasp planning methods,
  the collision points are searched in the simulator by closing the fingers from the sampled initial posture \cite{GraspIt:Miller:RAM2004}\cite{PowerGraspPlan:Roa:ICRA2012}.
  Although this method is applicable for generating whole-body contact motion \cite{WholebodyPush:Murooka:ICRA2015},
  the sampling-based method has disadvantages including high computational costs due to repeated trials and difficulty in generating time-series postures or postures with good evaluation values (e.g., joint torque and nominal posture).
  In this study, we employ the relatively efficient gradient-based optimization.
}{%
}%

\subsubsection{Optimization-based Motion Generation}
\myswlang{%
  Recently, the optimization-based motion generation has been actively studied \cite{WholebodyMotionPlan:Dai:Humanoids2014}.
  Inverse kinematics (IK) is the fundamental but non-trivial problem in motion generation \cite{HierarchicalQP:Escande:IJRR2014}.
  Our method is positioned as an extension of inverse kinematics
  because the conventional point-to-point IK is generalized to a body-to-body IK.
  To apply the gradient-based optimization that is widely used in motion generation,
  we introduce configuration variables representing points on the body surface and derive the gradient with respect to the variables.
  Another approach has been proposed in which optimization problems are formulated on non-Euclidean manifolds to handle the surface contacts \cite{ManifoldIk:Brossette:TRO2018}.
}{%
  勾配を用いた最適化ベースの動作生成法が多く研究されている(LAAS, MIT)\cite{}.
  最も一般的な逆運動学計算も，
  最適化問題の一つであると見なすことができる\cite{}．
  さらに，
  時系列まで考慮した
  trajectory inverse-kinematics\cite{}や，
  静力学まで考慮した
  inverse kinematics-and-statics\cite{}もなされている．
  これらは，
  目的関数の勾配情報を利用して，比較的高速に動作を生成している．
  本研究に置いても，
  ボディ表面上の接触点をコンフィギュレーションとして，
  位置と法線方向に対する勾配を求められるようにすることで，
  これらの先行研究と同じ最適化ベースの汎用的な枠組みの上での動作生成を可能とする．
  また，局所最適に落ちる問題に対して，
  解発散成分という考え方を導入し緩和する．
  これは，上記の様々な先行研究にも適用出来うるアイディアである．
}%

\subsubsection{Geometric Model Representation}
\myswlang{%
  The problem with searching a point on the body surface by the gradient-based optimization is the difficulty in calculating
  the gradient of the point on the edge of the robot link represented by a polyhedron.
  A solution to this problem is
  approximating the shape of the robot link to a smooth shape represented by a sphere-torus-patch \cite{STPBV:Escande:TRO2014}
  or Catmull-Clark subdivision surface \cite{CatmullClark:Escande:ICRA2016}.
  However, implementation and calculation costs are involved to approximate the robot model.
  In this study, normal direction smoothing enables calculation of the gradient from the raw convex mesh of the robot model without conversion.
}{%
  ボディの幾何表現については，
  ロボットの干渉回避\cite{}やモデリング\cite{}の文脈で研究されている．
  コンタクトポイントを固定せずに，任意の接触点を探索するためには，
  サンプリングに基づく方法が提案されている\cite{}が効率が良くない．
  ボディ表面上の接触点を勾配を用いた最適化で考慮するためには，
  微分可能である必要が有り，
  一般に多面体で表されるロボットのリンクは，エッジで微分不可能であることが問題となる．
  ボディ表面を微分可能にするためには，
  STP-BV\cite{}やスプラインメッシュ表現\cite{}などが提案されている．
  本研究では，
  ロボットのメッシュ構造をそのまま利用可能な，
  実装が簡単な勾配の導出とアップデートの手順を提案し，
  これを最適化の動作生成で利用する．
}%

\subsection{Contributions of this Study}
\myswlang{%
  The contributions of this study are as follows:
  (i) an optimization problem that simultaneously searches joint positions and the contact points on the body surface is formulated,
  (ii) a smoothed normal direction on the body surface is introduced for gradient-based optimization,
  (iii) various whole-body contact motions are shown through simulation and real robots.

  In the following,
  Sec.~II describes the problem and provides an outline of the solution,
  and Sec.~III describes contact point searching on the body surface in detail.
  Numerical examples are given in Sec.~IV, and Sec.~V demonstrates applications of the whole-body contact motion.
}{%
  本論文の特徴は次の3点である．
  (i) ロボットや環境の接触点を事前に指定せずとも，リンク(ボディ)を指定したら，その表面上で接触点が自動的に決まるような姿勢生成法を提案する．
  (ii) ボディ表面上の接触点をコンフィギュレーションとして，これによる勾配を導出することで，従来の最適化による運動生成の汎用的枠組みに基づいた解法である．
  (iii) 実際のロボットでマニピュレーションからロコモーションまで様々な動作において，有効性があることを示す．

  以降では，
  2章で，動作生成問題を定義と解法の概要を説明し，
  3章で，ボディ表面の接触点の勾配の導出について説明し，
  さらに，
  4章では，解発散項について説明し，
  5章で数値計算例，6章でロボットの全身接触動作生成への適用例を示す．
}%

\section{Overview of Whole-body Contact\\* Motion Generation}

\subsection{Problem Definition}
\begin{figure}[thpb]
  \centering
  \includegraphics[width=\columnwidth]{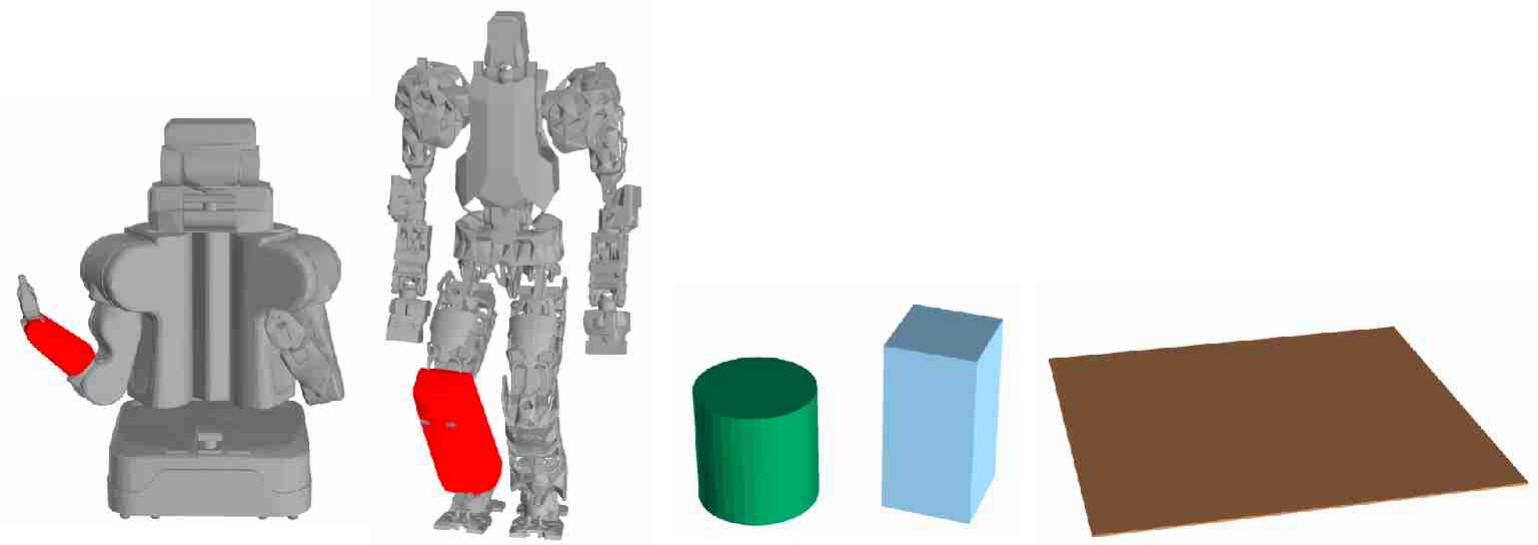}\\
  \vspace{-0.5mm}
  \begin{minipage}{0.3\columnwidth}
    \begin{center} \footnotesize robot link \end{center}
  \end{minipage}
  \begin{minipage}{0.3\columnwidth}
    \begin{center} \footnotesize object \end{center}
  \end{minipage}
  \begin{minipage}{0.25\columnwidth}
    \begin{center} \footnotesize ground \end{center}
  \end{minipage}
  \caption{Examples of body to be contact.
    \newline \footnotesize
    The body is modeled as a plane or convex polyhedron.
    For robot links,
    we use the convex polyhedron generated from the detailed mesh model automatically.
    Although the number of vertices in the body is high (about 50), the proposed optimization method based on normal smoothing works properly.
    We use simple primitive shapes (cylinder, cube, and sphere.) for the object and environment.
  }
  \label{fig:body}
\end{figure}

\begin{figure}[thpb]
  \centering
  \includegraphics[width=0.69\columnwidth]{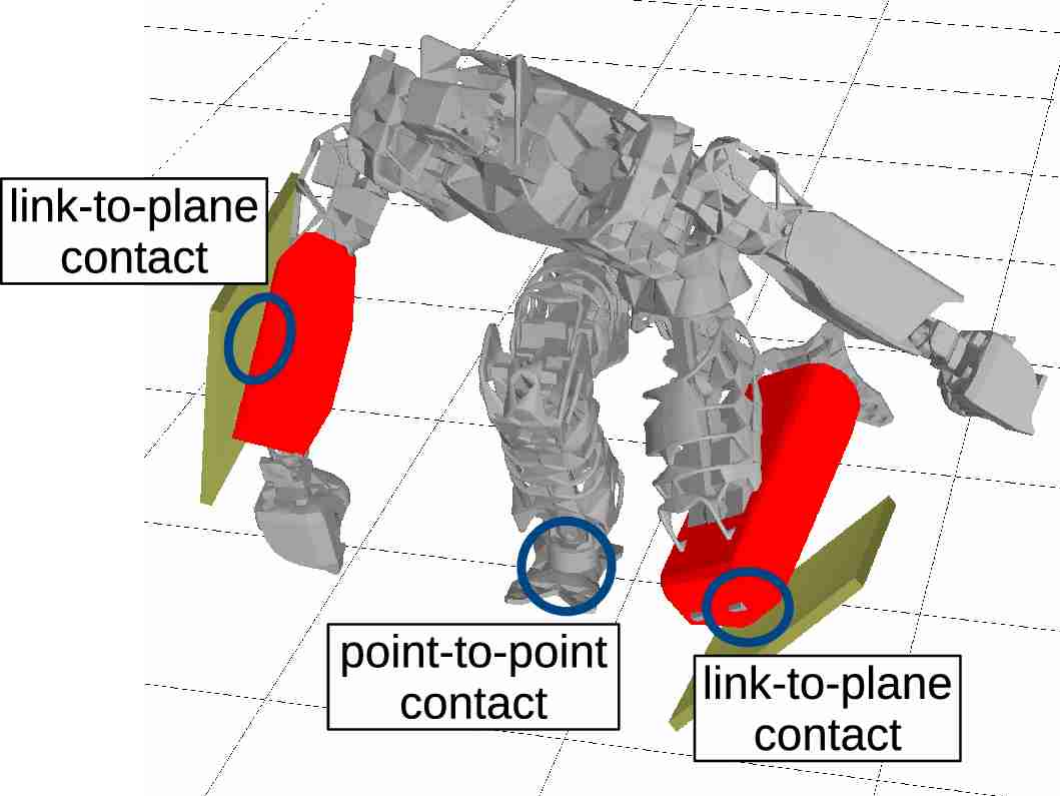}
  \includegraphics[width=0.29\columnwidth]{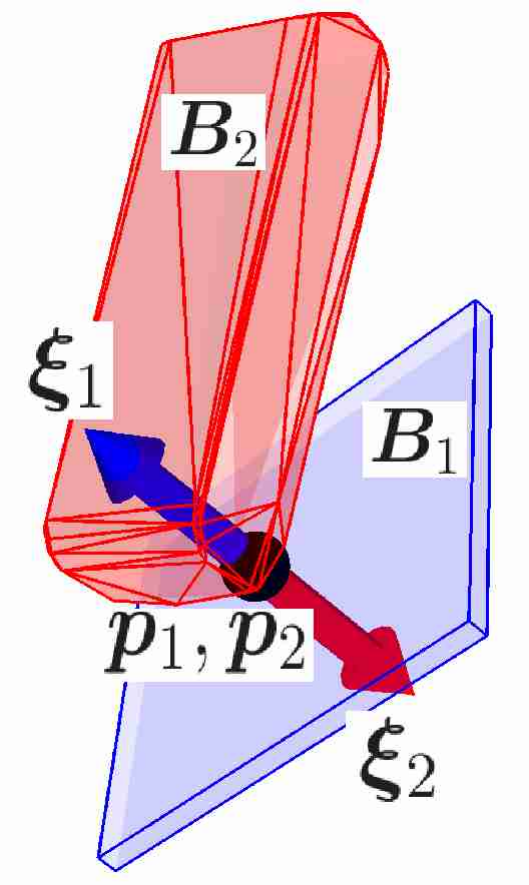}\\
  \vspace{-0.5mm}
  \begin{minipage}{0.69\columnwidth}
    \begin{center} \footnotesize (A) posture generation \end{center}
  \end{minipage}
  \begin{minipage}{0.29\columnwidth}
    \begin{center} \footnotesize (B) contact \end{center}
  \end{minipage}
  \caption{Example of posture generation.
    \newline \footnotesize
    (A)
    The figure shows an example of the whole-body contact posture generated by the proposed method.
    The robot is in contact with the environment in three places.
    The contacts of knee and elbow are designated as ``link-to-plane'' target, whereas the foot contact is designated as a conventional ``point-to-point'' target.
    (B)
    The figure also displays the enlarged view of contact at the knee.
    The contact positions of both bodies coincide, with reversed contact normal directions.
  }
  \label{fig:problem}
\end{figure}

\myswlang{%
  The simplest problem setting for whole-body contact motion generation is provided.
  The task goals are defined from the following information:
  \begin{itemize}
  \item Kinematics model of the robot, object, and environment.
  \item Pairs of bodies to be contacted at time index $t$, \\
    $\langle \bm{B}_{t,n,1}, \bm{B}_{t,n,2} \rangle\ (t=1,\cdots,T,\ n=1,\cdots,N(t))$.
  \end{itemize}
  where $T$ is the number of sampled time, and $N(t)$ is the number of contact pairs at time index $t$.
  The kinematics model must contain information of the links and joints.
  The body represents a robot link, object, or environment that is modeled as a plane or convex polyhedron.
  Examples of the body are shown in \figref{fig:body},
  and the problem task is defined through the following:
  \begin{eqnarray}
    &&\text{find \{ joint position, contact point \}$_{t=1,\cdots,T}$} \hspace{12mm} \label{eq:problem-def} \\
    &&\text{s.t. \ \ $\bm{B}_{t,n,1}$ and $\bm{B}_{t,n,2}$ contact.} \nonumber \\
    &&\phantom{\text{s.t. \ \ \ \ \ \ }} \text{$(t=1,\cdots,T,\ n=1,\cdots,N(t))$} \nonumber \\
    &&\phantom{\text{s.t. \ \ }} \text{The following constraints are satisfied:} \nonumber \\
    &&\phantom{\text{s.t. \ \ \ \ \ \ }} \text{joint position range and collision avoidance.} \nonumber
  \end{eqnarray}
  This problem is termed posture generation when $T = 1$, and motion generation when $T \geq 2$.
  An example of the problem of posture generation is displayed in \figref{fig:problem} (A).
}{%
  Whole-body contact motion generationのもっともシンプルな問題設定を以下で示す．

  タスクの目標は情報から定義される．
  \begin{itemize}
  \item ロボット，物体，環境などの幾何モデル．
  \item 接触すべきボディのペア．
  \end{itemize}
  ボディは，ロボットのリンクや物体を表し，凸多面体の剛体としてモデル化されるものとする．
  ボディの例を図に示す．

  タスクは以下で表される．
  \begin{eqnarray}
    &&{\rm find} \{{\rm joint pos, contact point}\}_{i=1,\cdots,T} \\
    &&{\rm s.t.} {\rm \bm{B}_i and \bm{B}_j contact} \\
    &&\phantom{\rm s.t.} {\rm within joint range, no collision, etc}
  \end{eqnarray}
  $T=1$のとき，posture generation,$T>2$のときmotion generationと言える．
  問題設定を表す図．
}%

\subsection{Problem Representation by Optimization}
\myswlang{%
  The problem \eqref{eq:problem-def} is formulated as an optimization problem as follows:
  \begin{eqnarray}
    &&\min_{\bm{\hat{q}}} \frac{1}{2} \| \bm{\hat{e}} (\bm{\hat{q}}) \|^2 \label{eq:opt-def} \\
    &&{\rm s.t.} \ \ \bm{c}(\bm{\hat{q}}) \geq \bm{0}
  \end{eqnarray}
  where $\bm{c}(\bm{\hat{q}})$ represents the conditions for joint position range and collision avoidance,
  while $\bm{\hat{e}}(\bm{\hat{q}})$ represents the conditions for contacts.

  The configuration $\bm{\hat{q}}$ is a time series of the joint position $\bm{\theta}_t$ and contact point $\bm{u}_t$ and is given by:
  \begin{eqnarray}
    &&\bm{\hat{q}} = \begin{pmatrix} \bm{q}_1^T & \cdots & \bm{q}_T^T \end{pmatrix}^T \label{eq:config-all} \\
    &&\bm{q}_t = \begin{pmatrix} \bm{\theta}_t^T & \bm{u}_t^T \end{pmatrix}^T \ \ t=1,\cdots,T \label{eq:config-t}
  \end{eqnarray}
  The $\bm{\theta}_t \in \mathbb{R}^M$ is a joint position vector where $M$ is the number of joints.
  The parameter $\bm{u}_t$ is a variable representing the point on the body surface,
  and is described in more detail in Sec.~III.

  The task function $\bm{\hat{e}}(\bm{\hat{q}})$ representing the contact of bodies is defined as follows:
  \begin{eqnarray}
    \bm{\hat{e}}(\bm{\hat{q}}) = \begin{pmatrix} \bm{e}_{1,1}^T(\bm{\hat{q}}) & \cdots & \bm{e}_{t,n}^T(\bm{\hat{q}}) & \cdots & \bm{e}_{T,N}^T(\bm{\hat{q}}) \end{pmatrix}^T
  \end{eqnarray}
  As the element task function $\bm{e}_{t,n}(\bm{\hat{q}})$, various kinds of kinematics tasks can be specified and combined.
  In common inverse kinematics problems,
  the tasks on position only ($\mathbb{R}^3$) and position and rotation ($\rm{SE}(3)$) are widely used \cite{ResolvedMotion:Whitney:TMS1969}.
  In addition, we introduce the following position and normal task (PN-task):
  \begin{eqnarray}
    \bm{e}(\bm{\hat{q}})
    &=&
    \begin{pmatrix} \bm{e}_{\mathit{pos}}(\bm{\hat{q}}) \\ w_{\mathit{nrm}} \bm{e}_{\mathit{nrm}}(\bm{\hat{q}}) \end{pmatrix} \label{eq:pn-const} \\
    {\rm where}&& \bm{e}_{\mathit{pos}}(\bm{\hat{q}}) = \bm{p}_1(\bm{\hat{q}}) - \bm{p}_2(\bm{\hat{q}}) \\
    {\phantom {\rm where}}&& \bm{e}_{\mathit{nrm}}(\bm{\hat{q}}) = \bm{\xi}_1^T(\bm{\hat{q}}) \, \bm{\xi}_2(\bm{\hat{q}}) + 1
  \end{eqnarray}
  The subscript ${t,n}$ is omitted while
  $\bm{p}_1(\bm{\hat{q}}), \bm{p}_2(\bm{\hat{q}}) \in \mathbb{R}^3$ represent the positions of contact points on the surface of bodies $\bm{B}_1, \bm{B}_2$, respectively.
  $\bm{\xi}_1(\bm{\hat{q}}), \bm{\xi}_2(\bm{\hat{q}}) \in \mathbb{R}^3$ represent the normal directions of the body surface at contact points, whereas
  $w_{\mathit{nrm}}$ is a constant to adjust the weight of the position and normal tasks.
  The norm of this task function becomes zero if and only if the contact positions coincide, and the contact normal directions are reverse
   (i.e., $\bm{p}_1=\bm{p}_2, \bm{\xi}_1=-\bm{\xi}_2$),
  with an example shown in \figref{fig:problem} (B).
  As another candidate for the formulation of the normal task,
  the following expression based on the angle between the normal directions can be considered:
  \mbox{$\arccos (\bm{\xi}_1^T(\bm{\hat{q}}) \, \bm{\xi}_2(\bm{\hat{q}})) - \pi$}.
  However, this expression was not used
  because it has the disadvantage that
  $\arccos$ becomes indifferentiable around the point where the task function becomes zero (i.e., at the point $\arccos(-1)$).
}{%
  この問題は，以下のような最適化問題として定式化される．
  \begin{eqnarray}
    &&\min_{\bm{q}} \frac{1}{2} \| \bm{e} (\bm{q}) \|^2 \\
    &&{\rm s.t.} \ \ \bm{c}(\bm{q}) \geq \bm{0}
  \end{eqnarray}

  ただし，探索コンフィギュレーションは以下のように
  時系列の関節位置と接触位置である．
  \begin{eqnarray}
    &&\bm{q} = \begin{pmatrix} \bm{q}_1^T & \cdots & \bm{q}_T^T \end{pmatrix}^T \\
    &&\bm{q}_t = \begin{pmatrix} \bm{\theta}_t^T & \bm{u}_t^T \end{pmatrix}^T
  \end{eqnarray}
  $\bm{u}_t \in \mathbb{R}^2$は，
  ボディ表面上を表す変数である．
  具体的な設定については次章で説明する．

  また，タスク関数は次式で定義される．
  \begin{eqnarray}
    &&\bm{e}(\bm{q}) = \begin{pmatrix} \bm{e}_1^T(\bm{q}) & \cdots & \bm{e}_N^T(\bm{q}) \end{pmatrix}^T \\
  \end{eqnarray}
  $\bm{e}_j(\bm{q})$としては，様々な制約を組み合わせることができる．
  位置のみの制約，位置姿勢の制約(姿勢にはaxis-angle表現を使う)が広く用いられているが，
  本研究では，これに加えて，以下の接触拘束を用いる．
  \begin{eqnarray}
    &&\bm{e}_j(\bm{q}) = \begin{pmatrix} \bm{u}^{\mathit{kin\mathchar`-trg}} - \bm{u}(\bm{u}) \\ \bm{\xi}^{\mathit{kin\mathchar`-trg},T} \bm{\xi}(\bm{u}) + 1 \end{pmatrix} \in \mathbb{R}^{4}
  \end{eqnarray}
  $\bm{\xi}^{\mathit{kin\mathchar`-trg}}, \bm{\xi}(\bm{u})$はそれぞれ，目標点，着目点の法線ベクトルを表す．
}%

\subsection{Overview of the Problem Solution}
\myswlang{%
  The optimization problem \eqref{eq:opt-def} is a non-linear optimization with constraints.
  Sequential quadratic programming (SQP) is a major method available to this kind of optimization problem \cite{NumericalOptimization:SWright:Springer1999}.
  In the SQP,
  the configuration $\bm{\hat{q}}$ is updated by the following equations:
  \begin{eqnarray}
    \!\!\!\!\!\!\!\! \bm{\hat{q}}^{[k+1]} \!\!\!\! &=& \!\!\!\! \bm{\hat{q}}^{[k]} + \Delta \bm{\hat{q}}^{*[k]} \label{eq:sqp-update} \\
    \!\!\!\!\!\!\!\! \Delta \bm{\hat{q}}^{*[k]} \!\!\!\! &=& \!\!\!\! \argmin_{\Delta \bm{\hat{q}}} \frac{1}{2} \Delta \bm{\hat{q}}^T \bm{J}_k^T \bm{J}_k \Delta \bm{\hat{q}} + \bm{\hat{e}}(\bm{\hat{q}}^{[k]})^T \bm{J}_k \Delta \bm{\hat{q}} \label{eq:sqp-qp} \\
    &&\ \ {\rm s.t.} \ \ \bm{c}(\bm{\hat{q}}^{[k]}) + \nabla \bm{c}(\bm{\hat{q}}^{[k]})^T \Delta \bm{\hat{q}} \geq \bm{0} \label{eq:sqp-constraint}
  \end{eqnarray}
  where $k$ is the iteration index.
  The iteration starts with some initial configuration $\bm{\hat{q}}^{[1]}$ corresponding to the nominal posture in this study,
  and terminates when the norm of the task function gets small enough or the maximum number of iterations is attained.
  The minimization in eq. \eqref{eq:sqp-qp} is a quadratic programming (QP) that is efficiently solved by the QP solver libraries.
  If the contact point is fixed to the robot or environment and
  the configuration $\bm{q}_t$ is without $\bm{u}_t$,
  the formulae~\eqref{eq:sqp-update}-\eqref{eq:sqp-constraint} provide an optimization-based solution for conventional IK \cite{HierarchicalQP:Escande:IJRR2014}.

  $\bm{J}_k$ is a Jacobian matrix of the task function given by:
  \begin{eqnarray}
    \bm{J}_k &=& \left. \frac{\partial \bm{\hat{e}}}{\partial \bm{\hat{q}}} \right|_{\bm{\hat{q}}=\bm{\hat{q}}^{[k]}}
  \end{eqnarray}
  and $\bm{J}_k$ consists of Jacobian matrices of element tasks $\frac{\partial \bm{e}_{t,n}}{\partial \bm{\theta}_t}, \frac{\partial \bm{e}_{t,n}}{\partial \bm{u}_t}$ as a block matrix.

  In the case the element task is PN-task \eqref{eq:pn-const},
  the Jacobian matrix with respect to the joint position $\bm{\theta}$ becomes as follows:
  \begin{eqnarray}
    \frac{\partial \bm{e}}{\partial \bm{\theta}} &=&
    \begin{pmatrix}
      \bm{j}_{\mathit{pos},1} & \cdots & \bm{j}_{\mathit{pos},M} \\
      w_{\mathit{nrm}} \bm{j}_{\mathit{nrm},1} & \cdots & w_{\mathit{nrm}} \bm{j}_{\mathit{nrm},M}
    \end{pmatrix} \label{eq:theta-jacobi} \ \ \ \ \\
    \bm{j}_{\mathit{pos},m} &=& \frac{\partial \bm{p}_1}{\partial \theta_m} - \frac{\partial \bm{p}_2}{\partial \theta_m} \ \ \ \ \ \ \ \ \ \ m=1,\cdots,M \\
    \bm{j}_{\mathit{nrm},m} &=& \bm{\xi}_1^T \frac{\partial \bm{\xi}_2}{\partial \theta_m} + \bm{\xi}_2^T \frac{\partial \bm{\xi}_1}{\partial \theta_m}
  \end{eqnarray}
  The subscript ${t,n}$ is omitted.
  $\frac{\partial \bm{p}_i}{\partial \theta_m}, \frac{\partial \bm{\xi}_i}{\partial \theta_m} (i \in \{1,2\})$ can be calculated from the kinematics model easily.
  For example, if the pose of the body $\bm{B}_1$ depends on the $m$th joint, whose type is revolute,
  \begin{eqnarray}
    \frac{\partial \bm{p}_1}{\partial \theta_m} &=& \bm{\bar{a}}_m \times (\bm{p}_1 - \bm{\bar{p}}_m) \\
    \frac{\partial \bm{\xi}_1}{\partial \theta_m} &=& \bm{\bar{a}}_m \times \bm{\xi}_1
  \end{eqnarray}
  $\bm{\bar{a}}_m$ and $\bm{\bar{p}}_m$ represent the axis and the position of the $m$th joint, respectively.

  $\frac{\partial \bm{e}_{t,n}}{\partial \bm{u}_t}$ is a Jacobian matrix with respect to the body surface point
  that is key in this study.
  This is explained in detail in the next section.
}{%
  この問題をSQPで解く\cite{}．
  \begin{eqnarray}
    &&\bm{q}^{[k+1]} = \bm{q}^{[k]} + \Delta \bm{q}^{[k]} \\
    &&\min_{\Delta \bm{q}} \frac{1}{2} \Delta \bm{q}^T \bm{J}^T \bm{J} \Delta \bm{q} + \bm{e}^T \bm{J} \Delta \bm{q} \\
    &&{\rm s.t.} \ \ \bm{c} + \nabla \bm{c}^T \Delta \bm{q} \geq \bm{0}
  \end{eqnarray}
  $\bm{J} = \frac{\partial \bm{e}(\bm{q})}{\partial \bm{q}}$である．
  これは，
  $\frac{\partial \bm{e}_j}{\partial \bm{\theta}_t}, \frac{\partial \bm{e}_j}{\partial \bm{u}_t}$をブロック行列として構成される．
  式*で示す接触幾何拘束の場合は，関節角のヤコビ行列は次式で表される．
  \begin{eqnarray}
    \frac{\partial \bm{e}_j}{\partial \bm{\theta}_t} =
    - \bm{\xi}_{m}^{\mathit{kin\mathchar`-trg}, T}(\bm{q}) \frac{\partial \bm{\xi}_{m}^{\mathit{kin\mathchar`-att}}(\bm{q})}{\partial \bm{\phi}} - \bm{\xi}_{m}^{\mathit{kin\mathchar`-att}, T}(\bm{q}) \frac{\partial \bm{\xi}_{m}^{\mathit{kin\mathchar`-trg}}(\bm{q})}{\partial \bm{\phi}}
    \bm{\bar{a}} \times \bm{n}
  \end{eqnarray}
  $\frac{\partial \bm{e}_j}{\partial \bm{u}_t}$は，
  ボディ表面上のヤコビ行列に相当しており，
  次章で提案する．
}%

\section{Optimization of Contact Point\\* on Body Surface}

\subsection{Contact Configuration}
\begin{figure}[thpb]
  \centering
  \includegraphics[width=0.8\columnwidth]{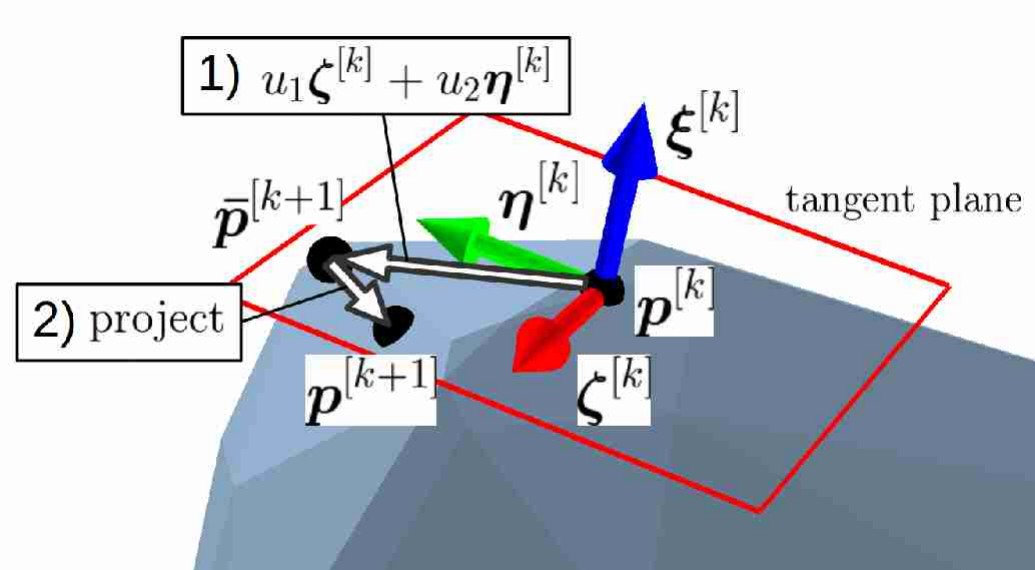}
  \caption{Mapping from the contact configuration to the contact point.
    \newline \footnotesize
    The contact point of the next iteration is calculated by first moving locally on the tangent plane and then projecting it onto the body.
  }
  \label{fig:contact-config}
\end{figure}

\myswlang{%
  In this section,
  we define configuration variables representing the contact points on the body surface.

  In eq.~\eqref{eq:config-t}, we introduced the configuration $\bm{u}_t$, that represents contact points at time index $t$.
  The parameter $\bm{u}_t$ consists of two-dimensional configurations $\bm{u}_{t,n,i}$ corresponding to each contact point
  where $n$ is the index of the contact pair and $i \in \{1,2\}$ is the index of the body.

  From here,
  we focus on one contact point and omit the subscripts ${t,n,i}$.
  Assuming that the contact point of the $k$th iteration $\bm{p}^{[k]} \in \mathbb{R}^3$ is given,
  we map the contact configuration $\bm{u} \in \mathbb{R}^2$ to the contact point of the next iteration $\bm{p}^{[k+1]}$ as follows (\figref{fig:contact-config}):
  \begin{enumerate}
  \item Move locally on the tangent plane.
  \begin{eqnarray}
    \bm{\bar{p}}^{[k+1]}(\bm{u})
    &=& u_1 \bm{\zeta}^{[k]} + u_2 \bm{\eta}^{[k]} + \bm{p}^{[k]} \\
    &=& \begin{pmatrix} \bm{\zeta}^{[k]} & \bm{\eta}^{[k]} \end{pmatrix} \bm{u} + \bm{p}^{[k]} \label{eq:contact-config-1} \\
    {\rm where}&& - u_{\mathit{lim}} \leq u_1, u_2 \leq u_{\mathit{lim}} \label{eq:contact-config-lim}
  \end{eqnarray}
  \item Project from the tangent plane onto the body.
  \begin{eqnarray}
    \bm{p}^{[k+1]}(\bm{u}) = \mathit{project} (\, \bm{\bar{p}}^{[k+1]}(\bm{u}), \bm{B} \,) \label{eq:contact-config-2}
  \end{eqnarray}
  \end{enumerate}
  The parameters $u_1, u_2$ are elements of $\bm{u}$.
  For each iteration, we define a coordinate system with the contact point as the origin
  where $\bm{\zeta}^{[k]}, \bm{\eta}^{[k]} \in \mathbb{R}^3$ are the tangential axes
  and $\bm{\xi}^{[k]} \in \mathbb{R}^3$ is the normal axis of the body surface.
  We use a smoothed normal direction as $\bm{\xi}^{[k]}$, that is described in Sec.~III.C.
  The upper and lower limits $u_{\mathit{lim}}$ are set in the contact configuration (eq.~\eqref{eq:contact-config-lim})
  to prevent the error between the tangent plane and convex body becomes too large;
  $u_{\mathit{lim}}$ of life-sized robots is set to be in the range 10-20 mm.
  The function $\mathit{project} ( \bm{p}, \bm{B} )$ returns the point on the body $\bm{B}$ that is closest to the external point $\bm{p}$.
  For this projection, we employed an algorithm that exhaustively compares distances between $\bm{p}$ and each convex face of $\bm{B}$.
}{%
  本章では，
  ボディ表面の接触点を表すコンフィギュレーション$\bm{u}_t$を定義し，
  その勾配$\frac{\partial \bm{e}_j}{\partial \bm{u}_t}$を導出する．

  あるkイテレーション目の，ボディ表面上の位置が
  $\bm{p}^{[k]} \in \mathbb{R}^3$であるとする．
  このとき$\bm{u}$は，接平面上の二次元座標系の座標を表す（ボディの接平面の図）．
  このあと，射影する．

  $\bm{p}(\bm{u}) \in \mathbb{R}^3$
  \begin{eqnarray}
    \bm{\zeta}^{[k]}, \bm{\eta}^{[k]}, \bm{\xi}^{[k]} \in \mathbb{R}^3
  \end{eqnarray}
  \begin{eqnarray}
    \bm{\bar{p}}^{[k]}(\bm{u})
    &=& u_1 \bm{\zeta}^{[k]} + u_2 \bm{\eta}^{[k]} + \bm{p}^{[k]} \\
    &=& \begin{pmatrix} \bm{\zeta}^{[k]} & \bm{\eta}^{[k]} \end{pmatrix} \bm{u} + \bm{p}^{[k]}
  \end{eqnarray}
  \begin{eqnarray}
    \frac{\partial \bm{e}}{\partial \bm{u}}
    &=&
    \begin{pmatrix}
      \frac{\partial \bm{p}}{\partial \bm{u}} \\
      \bm{\xi}^T \frac{\partial \bm{\xi}}{\partial \bm{u}}
    \end{pmatrix} \\
    &\approx&
    \begin{pmatrix}
      \frac{\partial \bm{\bar{p}}}{\partial \bm{u}} \\
      - \bm{\xi}^T \frac{\partial \bm{\xi}}{\partial \bm{u}}
    \end{pmatrix} \\
  \end{eqnarray}
  \begin{eqnarray}
    \frac{\partial \bm{\xi}}{\partial \bm{u}}
    &=&
    \begin{pmatrix}
      \frac{\partial \bm{\xi}}{\partial u_1} & \frac{\partial \bm{\xi}}{\partial u_2}
    \end{pmatrix} \\
    &=&
    \begin{pmatrix}
      \frac{\bm{\xi}^{1+} -  \bm{\xi}^{1-}}{2 \Delta u_1} & \frac{\bm{\xi}^{2+} -  \bm{\xi}^{2-}}{2 \Delta u_2}
    \end{pmatrix}
  \end{eqnarray}
  \begin{eqnarray}
    \bm{p}^{[k+1]} = \mathit{project} ( \bm{\bar{p}}^{[k]} )
  \end{eqnarray}
}%

\subsection{Optimization of Contact Point}
\myswlang{%
  By solving QP \eqref{eq:sqp-qp}, we can obtain the optimal contact configuration $\bm{u}^{*[k]}$ and then the optimal contact point $\bm{p}^{*[k]}$ in each iteration.
  We derive the Jacobian matrix $\frac{\partial \bm{e}_{t,n}}{\partial \bm{u}_{t,n,i}}$ of PN-task \eqref{eq:pn-const}, that is necessary to formulate QP, as follows:
  \begin{eqnarray}
    \frac{\partial \bm{e}}{\partial \bm{u}}
    &=&
    \begin{pmatrix}
      \frac{\partial \bm{p}_1}{\partial \bm{u}} \\
      w_{\mathit{nrm}} \bm{\xi}_2^T \frac{\partial \bm{\xi}_1}{\partial \bm{u}}
    \end{pmatrix} \in \mathbb{R}^{4 \times 2} \label{eq:contact-jacobi} \\
    \frac{\partial \bm{p}_1}{\partial \bm{u}}
    &=&
    \begin{pmatrix}
      \bm{\zeta} & \bm{\eta}
    \end{pmatrix} \in \mathbb{R}^{3 \times 2} \label{eq:contact-jacobi-position} \\
    \frac{\partial \bm{\xi}_1}{\partial \bm{u}}
    &=&
    \begin{pmatrix}
      \frac{\bm{\xi}_1^{1+} -  \bm{\xi}_1^{1-}}{2 \varepsilon} & \frac{\bm{\xi}_1^{2+} -  \bm{\xi}_1^{2-}}{2 \varepsilon}
    \end{pmatrix} \in \mathbb{R}^{3 \times 2} \label{eq:contact-jacobi-normal}
  \end{eqnarray}
  The subscripts ${t,n,i}$ and superscript $k$ are omitted.
  Eqs.~\eqref{eq:contact-jacobi}-\eqref{eq:contact-jacobi-normal} describe the case when $\bm{u}$ represents the contact point on body $\bm{B}_1$.
  The formula for a contact point on $\bm{B}_2$ can be derived similarly.
  The Jacobian of normal direction is derived by finite difference (eq.~\eqref{eq:contact-jacobi-normal}).
  $\bm{\xi}_1^{i\pm} (i \in \{1,2\})$ represents the normal direction at the contact point when $u_i = \pm \varepsilon$.

  Since the contact configuration is defined only in a locally linearized coordinate system, there is no global configuration for the contact point.
  Therefore, the contact configuration is not updated in eq.~\eqref{eq:sqp-update} in the SQP.
  Instead, the new contact point $\bm{p}^{[k+1]}$ is saved and used as the origin for the linearization of the next iteration.

  In summary, each iteration of contact optimization is performed as follows:
  \begin{enumerate}
  \setcounter{enumi}{-1}
  \item The contact point of the initial iteration $\bm{p}^{[1]}$ is determined by some rule (e.g., random or closest to the target).
  \item The Jacobian matrix at the contact point $\bm{p}^{[k]}$ is calculated from eq.~\eqref{eq:contact-jacobi}.
  \item The optimal contact configuration $\bm{u}^{*[k]}$ is obtained by solving QP \eqref{eq:sqp-qp}.
    The point on the tangent plane $\bm{\bar{p}}^{[k+1]}$ is simultaneously obtained (eq.~\eqref{eq:contact-config-1}).
  \item A new contact point $\bm{p}^{[k+1]}$ is calculated by projection from eq.~\eqref{eq:contact-config-2}.
  \end{enumerate}
  The joint position $\bm{\theta}$ is also updated in each iteration.
  By this optimization-based method,
  a robot posture is generated while automatically searching contact points.
}{%
}%

\subsection{Normal Smoothing}
\begin{figure}[thpb]
  \centering
  \includegraphics[width=0.35\columnwidth]{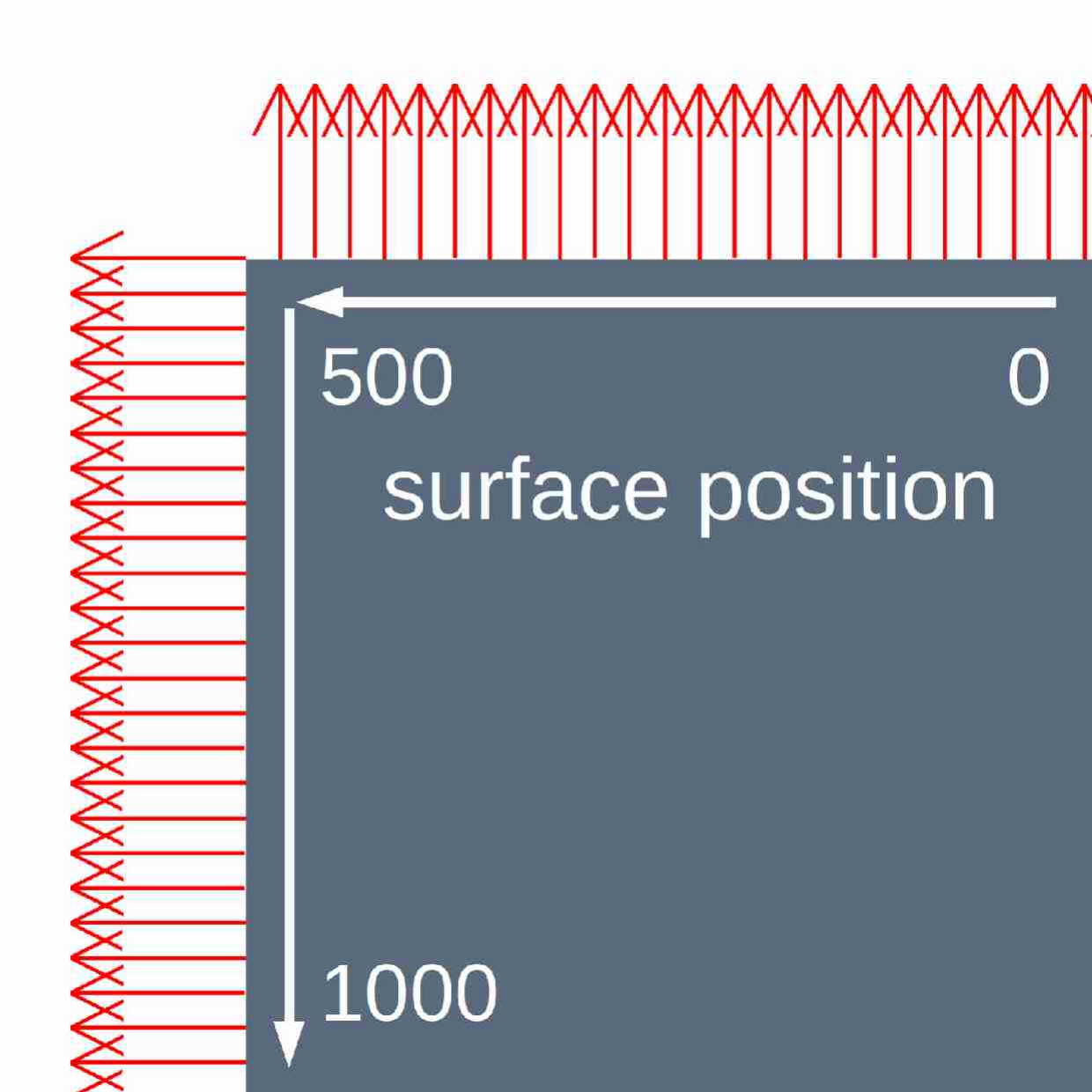}
  \hspace{10mm}
  \includegraphics[width=0.35\columnwidth]{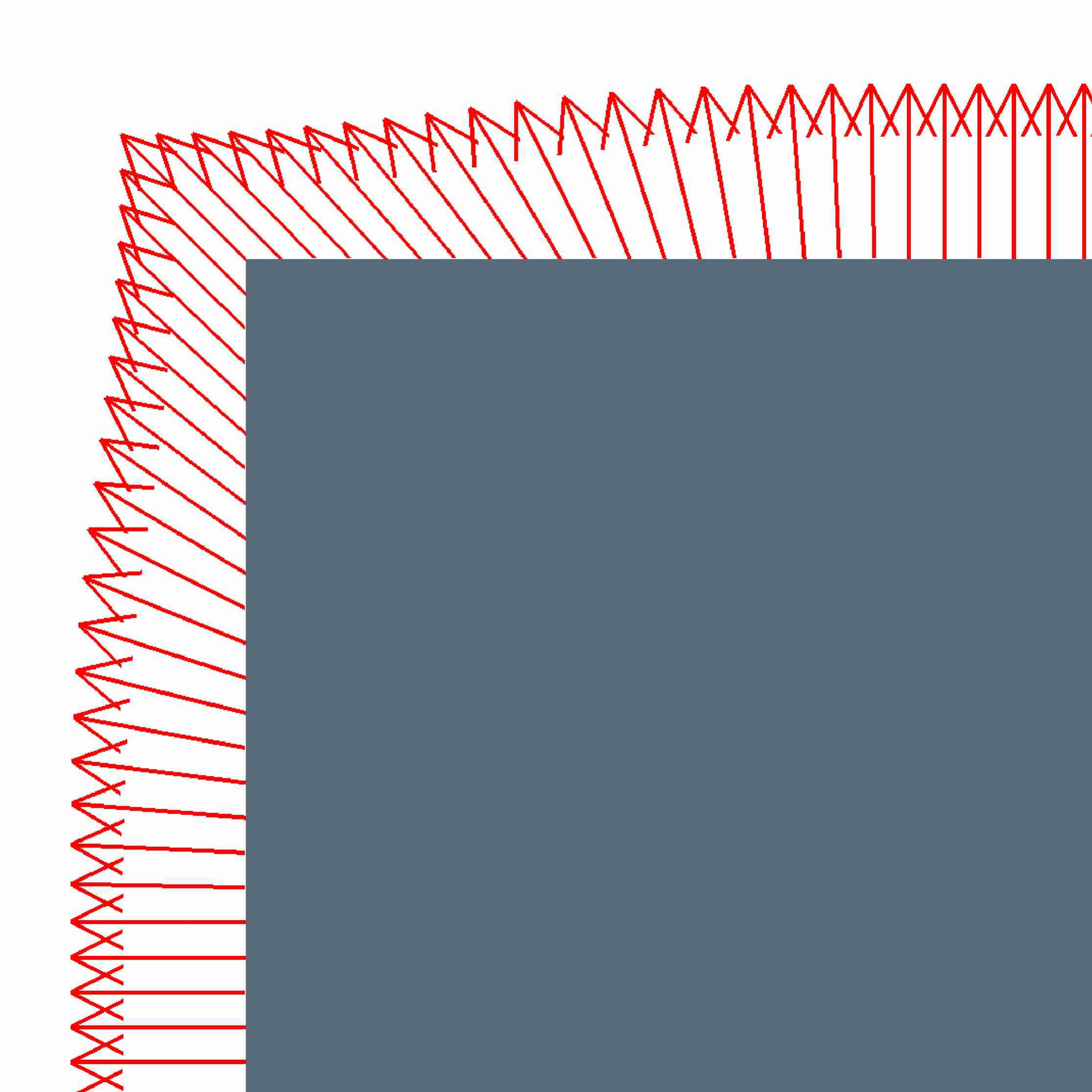}\\
  \vspace{2mm}
  \includegraphics[width=0.49\columnwidth]{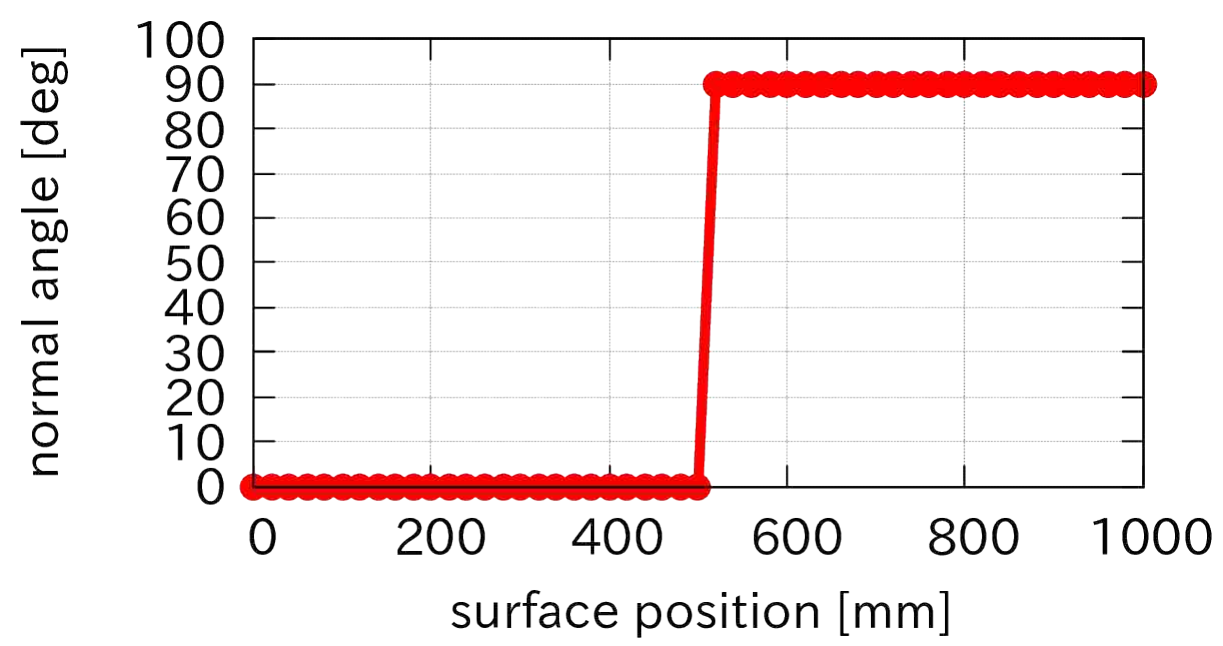}
  \includegraphics[width=0.49\columnwidth]{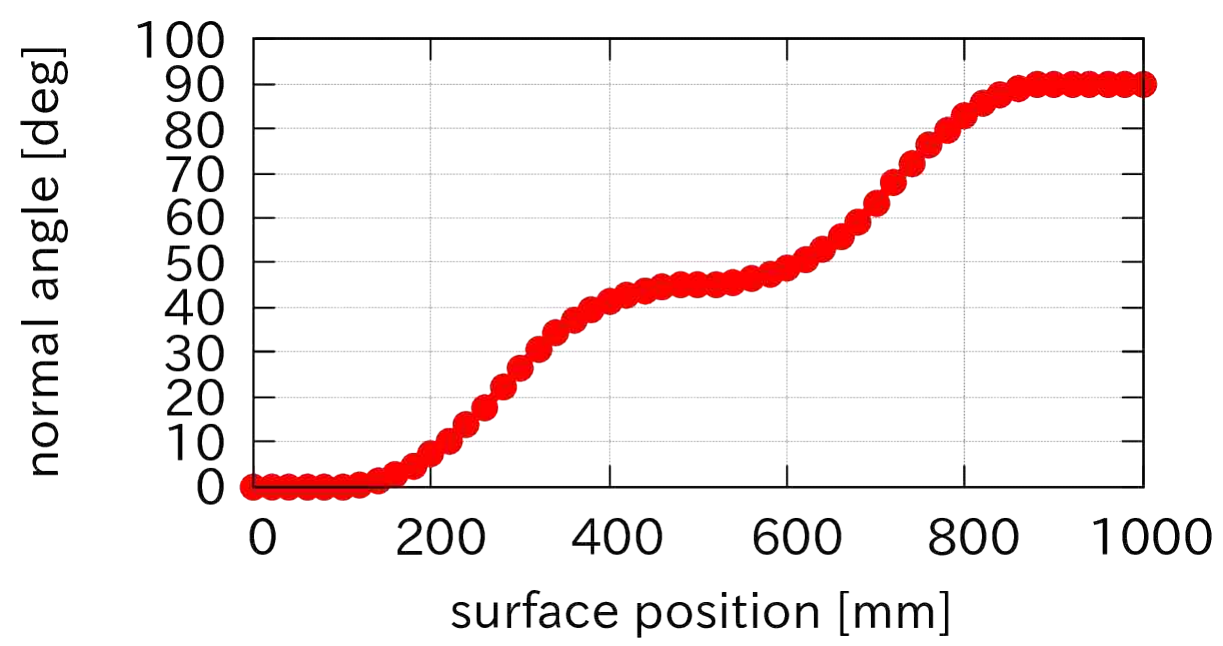}\\
  \vspace{-0.5mm}
  \begin{minipage}{0.49\columnwidth}
    \begin{center} \footnotesize (A) conventional normal direction \end{center}
  \end{minipage}
  \begin{minipage}{0.49\columnwidth}
    \begin{center} \footnotesize (B) smoothed normal direction \end{center}
  \end{minipage}
  \caption{Example of a normal calculation of the body surface.
    \newline \footnotesize
    The smoothing range ($R$ in eq.~\eqref{eq:normal-interpolate-func}) is $400$ mm.
  }
  \label{fig:smooth-normal}
\end{figure}

\begin{figure}[thpb]
  \centering
  \includegraphics[width=0.49\columnwidth]{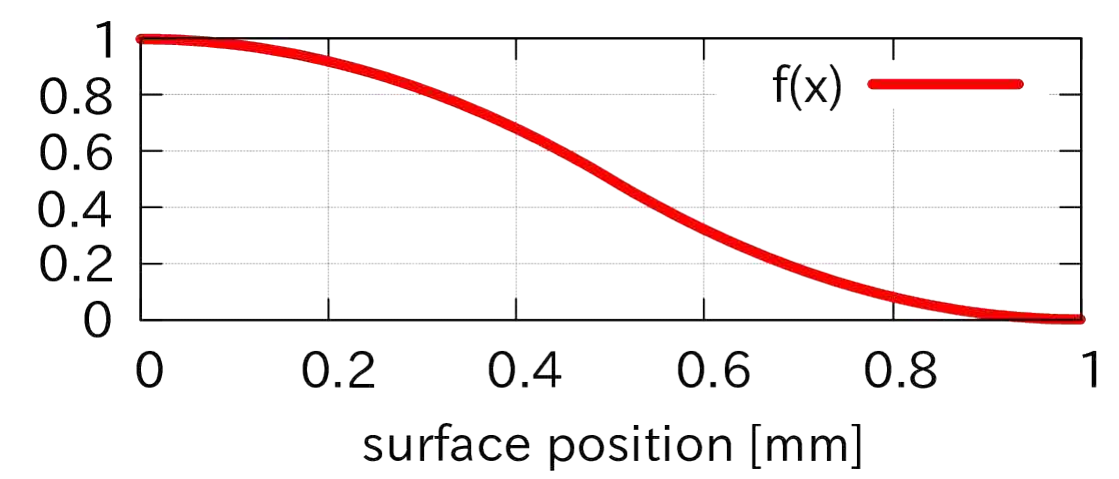}
  \includegraphics[width=0.49\columnwidth]{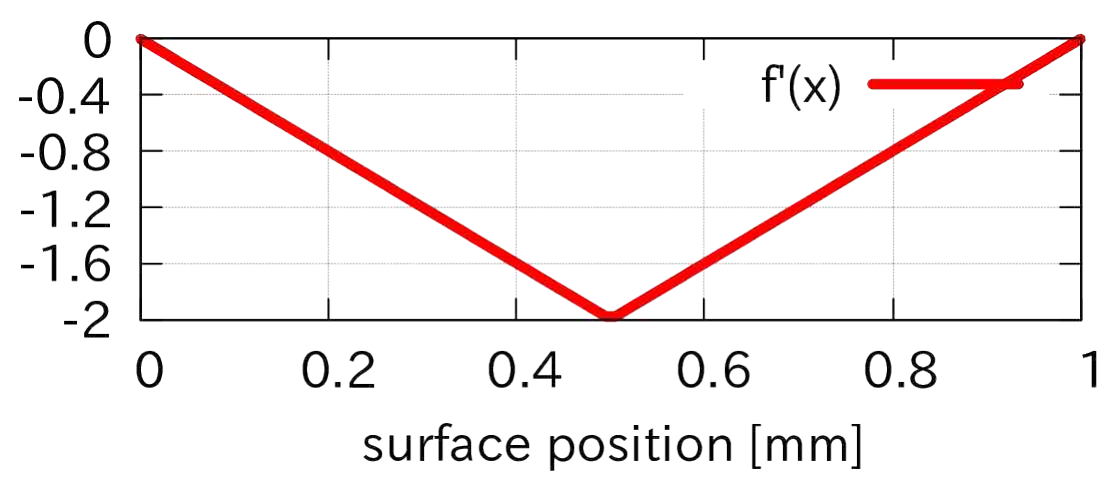}\\
  \vspace{-0.5mm}
  \begin{minipage}{0.49\columnwidth}
    \begin{center} \footnotesize (A) $y = f(x)$ (eq.~\eqref{eq:normal-interpolate-func}) \end{center}
  \end{minipage}
  \begin{minipage}{0.49\columnwidth}
    \begin{center} \footnotesize (B) $y = \frac{d f(x)}{d x}$ \end{center}
  \end{minipage}
  \caption{Smooth interpolation function ($R = 1$).
  }
  \label{fig:normal-interpolate-weight}
\end{figure}

\begin{figure}[thpb]
  \centering
  \includegraphics[width=0.8\columnwidth]{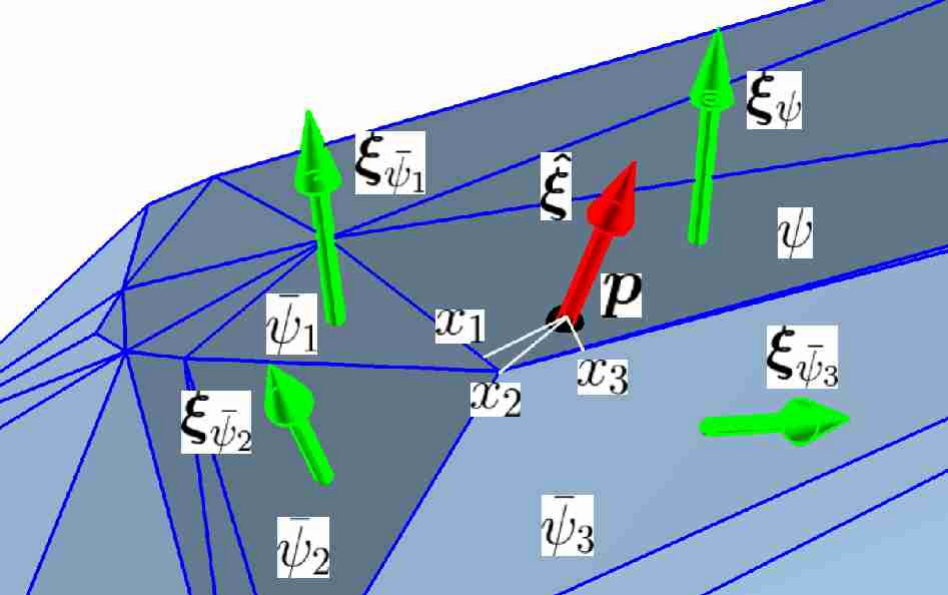}
  \caption[]{Example of normal calculation of the 3D convex mesh.
    \newline \footnotesize
    \begin{minipage}{0.95\linewidth}
      \vspace{0.75mm}
      The mesh represents the knee link of the robot in \figref{fig:problem}.
      $x_i$ represents an abbreviation for $x(\bm{p}, \bar{\psi}_i)$.
      The smoothing range $R$ is $30$ mm.
      In this range, three adjacent faces are found: $\bar{\Psi} = \{ \bar{\psi}_1, \bar{\psi}_2, \bar{\psi}_3 \}$.
      The distance to the faces and the normal weight are as follows:
      {
        \vspace{-1.5mm}
        \setlength{\jot}{1pt}
        \begin{align*}
          &x(\bm{p}, \bar{\psi}_1) = 20.4 {\rm \ mm}, &w_{\bar{\psi}_1} = 0.162 \\
          &x(\bm{p}, \bar{\psi}_2) = 26.4 {\rm \ mm}, &w_{\bar{\psi}_2} = 0.002 \\
          &x(\bm{p}, \bar{\psi}_3) = \phantom{0}9.3 {\rm \ mm}, &w_{\bar{\psi}_3} = 0.707
        \end{align*}
      }
    \end{minipage}
  }
  \label{fig:normal-smooth-weight}
\end{figure}

\myswlang{%
  To apply a gradient-based optimization like the SQP,
  a differentiable task function is required.
  When the body is represented by a convex polyhedron, the normal direction of the body surface is indifferentiable on the edges (\figref{fig:smooth-normal}(A)).
  This also renders the tangent direction indifferentiable, so that all components of the PN-task \eqref{eq:pn-const} become indifferentiable.

  We propose to define the smoothed normal $\bm{\hat{\xi}}$ on the body surface point $\bm{p}$ as follows:
  \begin{eqnarray}
    \bm{\hat{\xi}}(\bm{p}) = \mathit{normalize} \left( \frac{\bm{\xi}_{\psi} + \sum_{\bar{\psi} \in \bar{\Psi}(\psi)} w_{\bar{\psi}}(\bm{p}) \bm{\xi}_{\bar{\psi}}}{1 + \sum_{\bar{\psi} \in \bar{\Psi}(\psi)} w_{\bar{\psi}}(\bm{p})} \right)
  \end{eqnarray}
  and use it instead of the conventional normal $\bm{\xi}$.
  This formula represents the normalized vector of the weighted average of normals.
  $\psi$ represents a face to which the surface point $\bm{p}$ belongs.
  $\bar{\Psi}(\psi)$ represents a set of faces adjacent to $\psi$,
  with $\bm{\xi}_{\psi}$ representing a normal of the face $\psi$.

  The parameter $w_{\bar{\psi}}(\bm{p})$ is a normal weight of adjacent faces calculated as follows:
  \begin{eqnarray}
    && w_{\bar{\psi}}(\bm{p}) = f(x(\bm{p}, \bar{\psi})) \\
    && {\rm where} \ \ f(x) = \left\{ \begin{array}{ll}
      -2 \left( \frac{x}{R} \right)^2 + 1 & (0 \leq x \leq \frac{R}{2}) \ \ \ \ \\
      2 \left( \frac{x}{R} - 1 \right)^2 & (\frac{R}{2} \leq x \leq R)
    \end{array} \right. \label{eq:normal-interpolate-func}
  \end{eqnarray}
  $x(\bm{p}, \bar{\psi})$ is the distance on the body surface from point $\bm{p}$ to the adjacent face $\bar{\psi}$;
  $f(x)$ is a smooth function for interpolation from $1$ to $0$ (\figref{fig:normal-interpolate-weight});
  and $R$ is the smoothing range, within which adjacent faces are searched.
  Since this function is smooth and its gradient is zero at $x=0, \, R$, the smoothed normal $\bm{\hat{\xi}}$ becomes differentiable (\figref{fig:smooth-normal}(B)).
  Taking into account the discrete change in a set of adjacent faces $\bar{\Psi}(\psi)$,
  the finite difference was used to calculate the derivative in eq.~\eqref{eq:contact-jacobi-normal}.

  This method does not require prior conversion of the body shape and can handle various 3D shapes from detailed convex meshes to primitive shapes.
  \figref{fig:normal-smooth-weight} depicts an example of a 3D convex mesh.
}{%
}%

\section{Numerical Examples}

We implemented the proposed motion generation algorithm with EusLisp \cite{EusLisp},
that is an extension of the Common Lisp for robot programming.
We used external libraries qpOASES \cite{qpOASES:Ferreau:MPC2014} and Bullet \cite{Bullet:Coumanns}
for QP and collision detection, respectively.

\subsection{Gradient-based Contact Point Searching}
\begin{figure}[thpb]
  \centering
  \includegraphics[height=0.32\columnwidth]{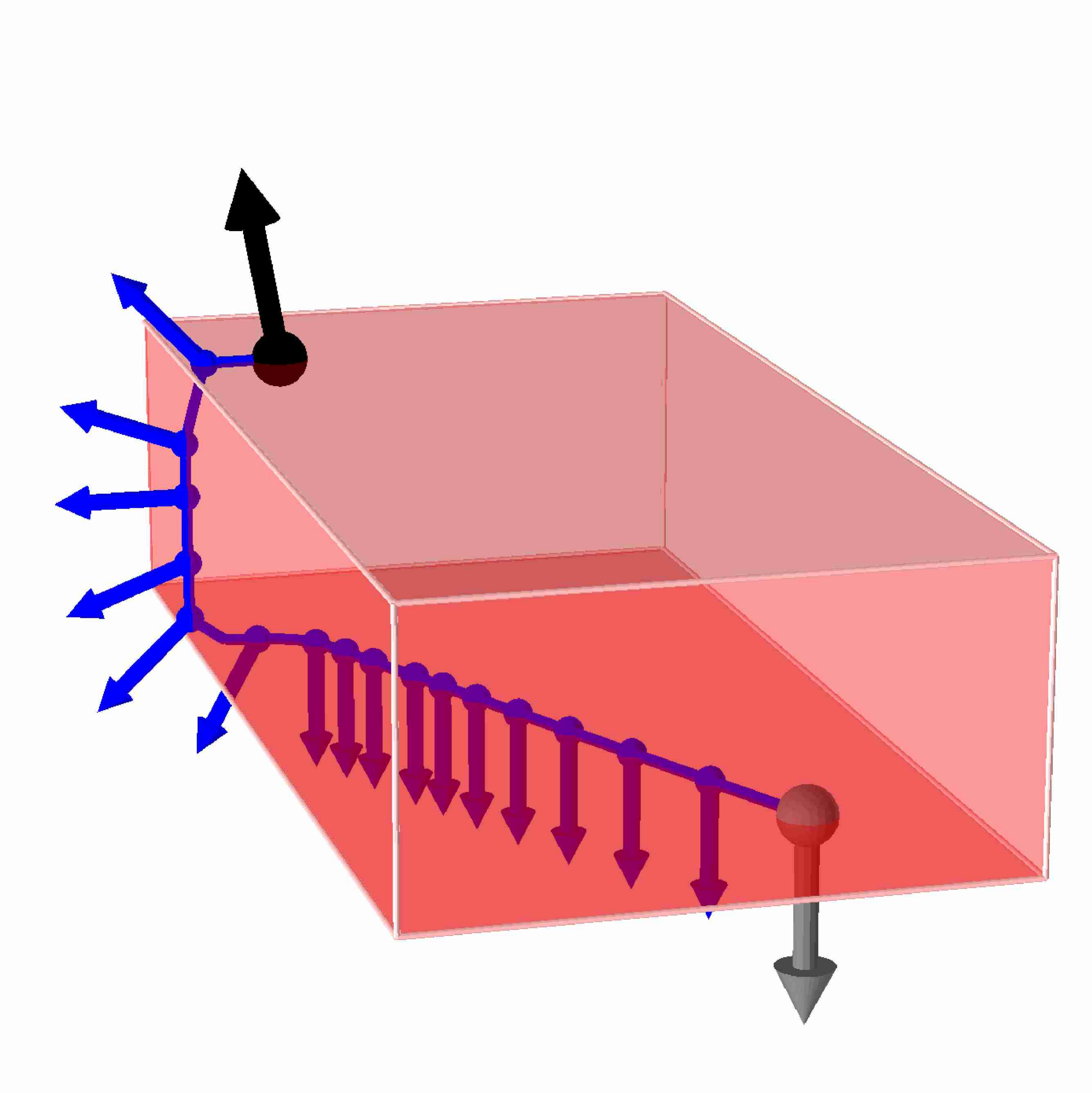}
  \includegraphics[height=0.32\columnwidth]{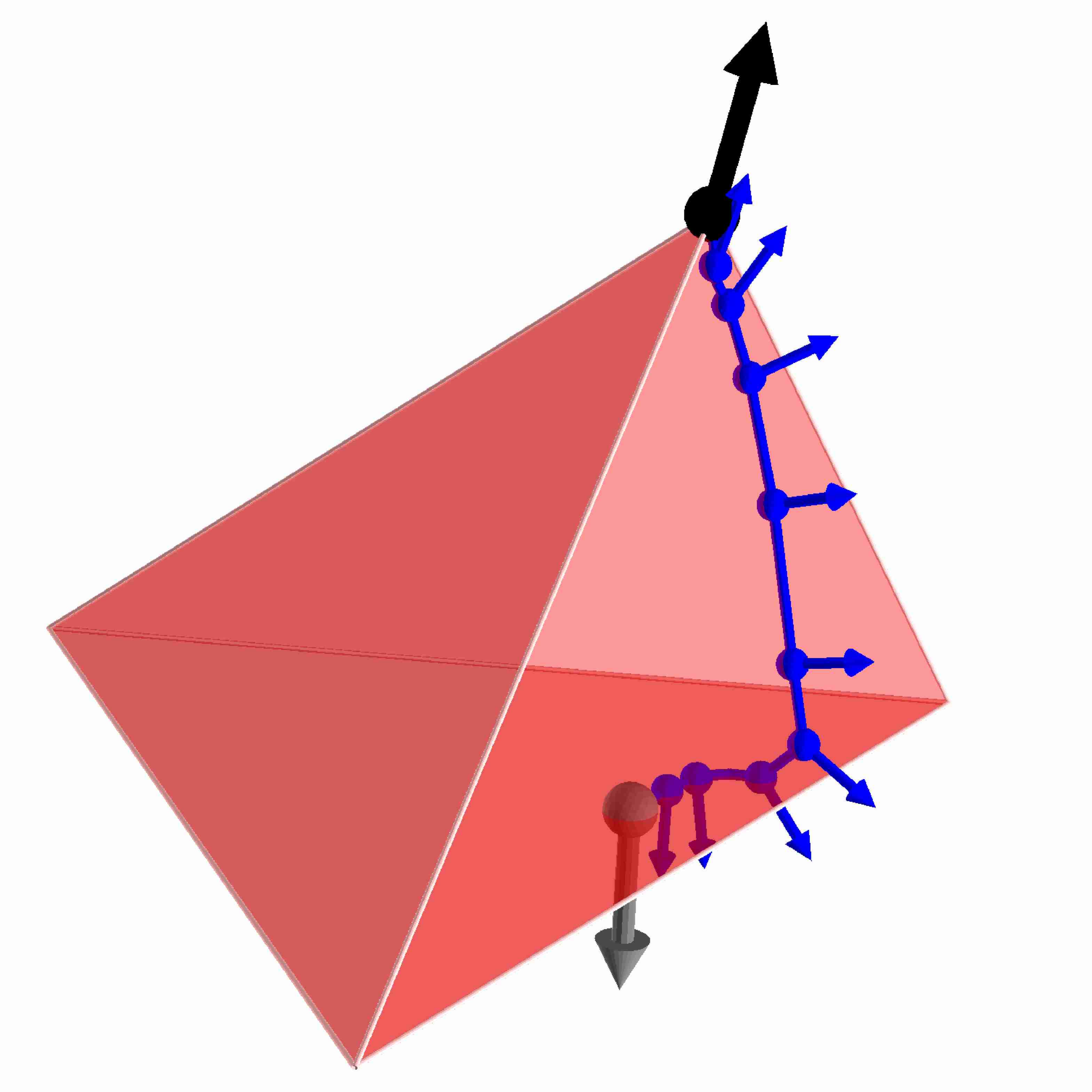}
  \includegraphics[height=0.32\columnwidth]{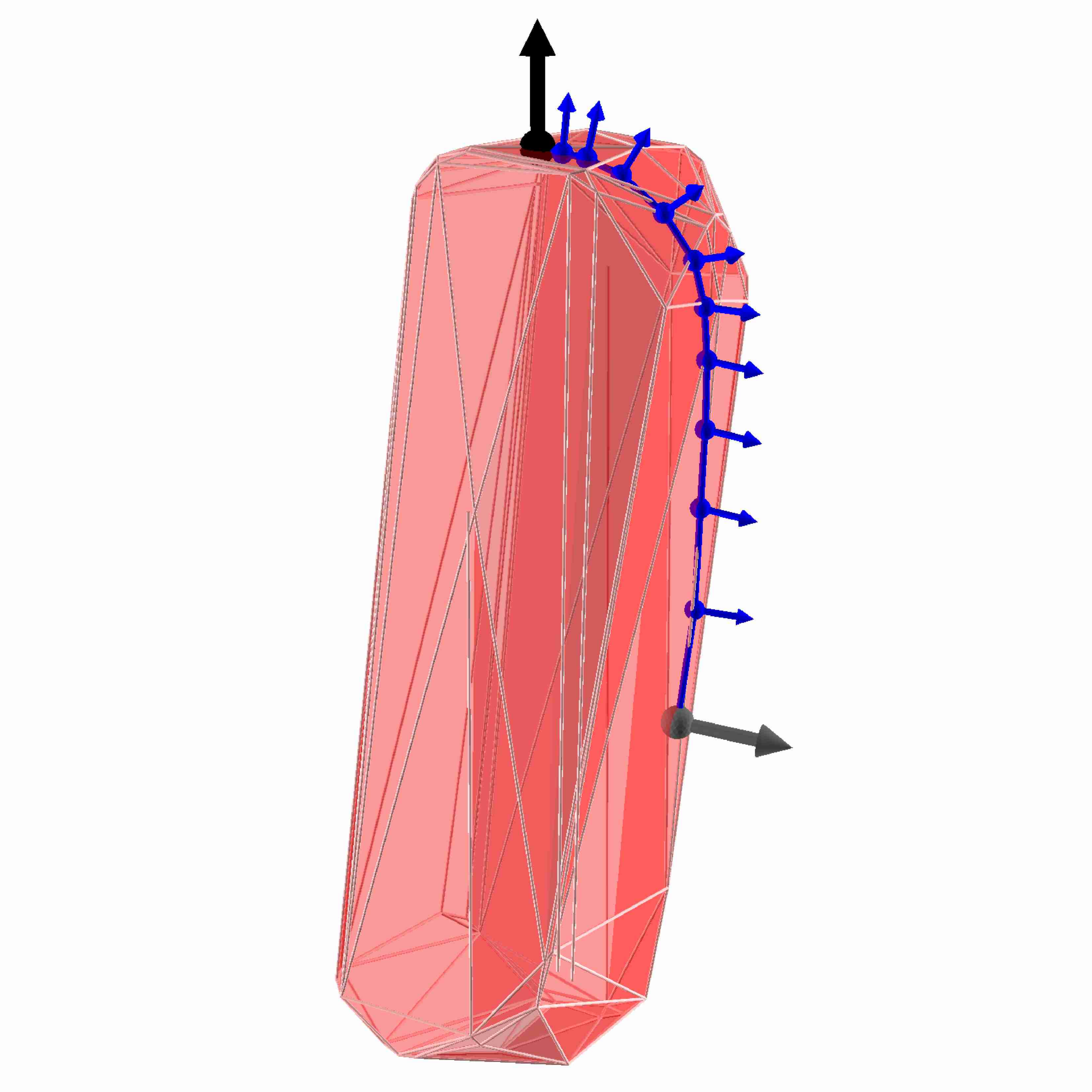}\\
  \vspace{-1.75mm}
  \begin{minipage}{0.34\columnwidth}
    \begin{center} \footnotesize (A1) box \end{center}
  \end{minipage}
  \begin{minipage}{0.28\columnwidth}
    \begin{center} \footnotesize (A2) tetrahedron \end{center}
  \end{minipage}
  \begin{minipage}{0.34\columnwidth}
    \begin{center} \footnotesize (A3) mesh \end{center}
  \end{minipage}\\
  \includegraphics[height=0.32\columnwidth]{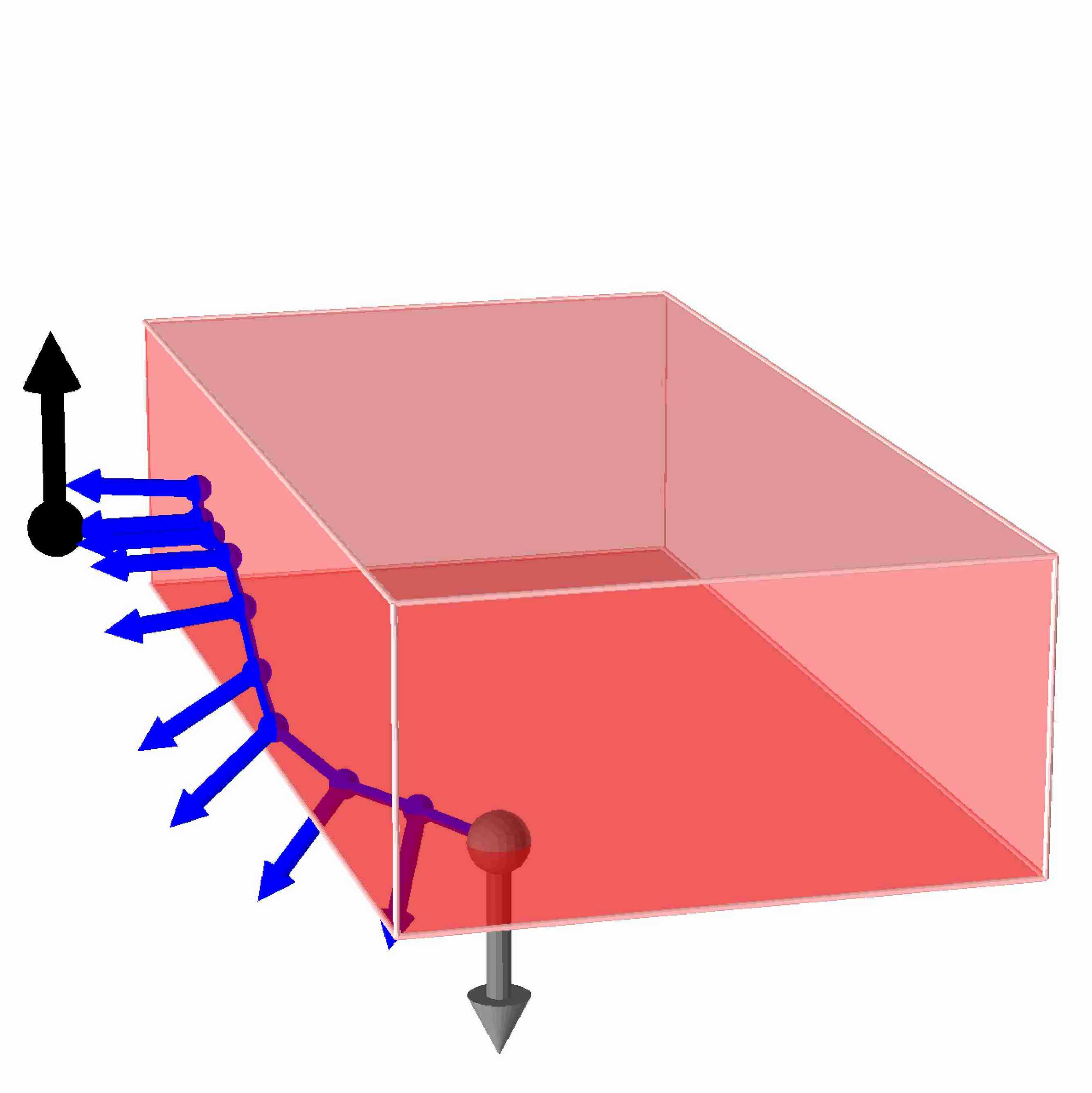}
  \includegraphics[height=0.32\columnwidth]{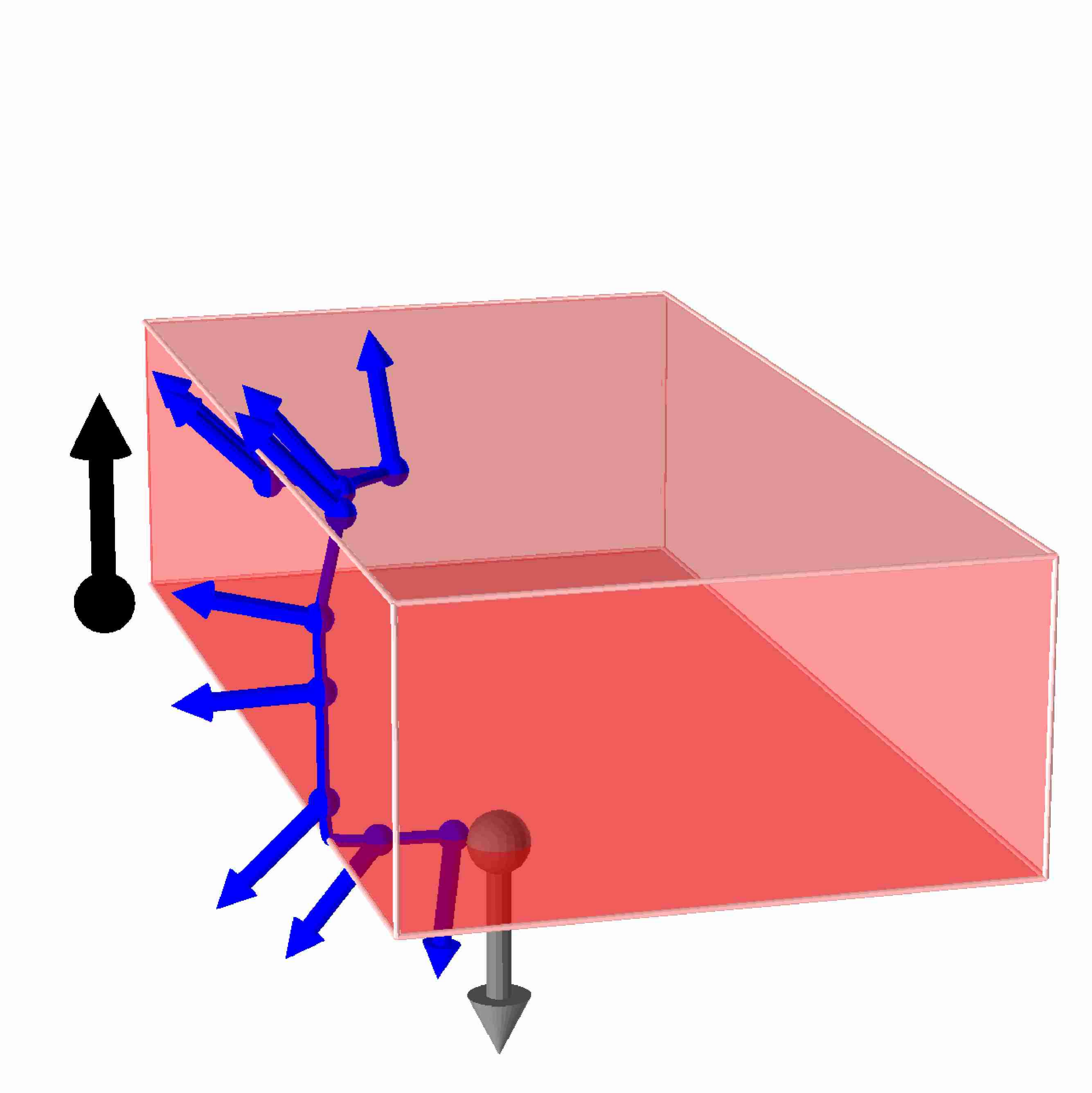}
  \includegraphics[height=0.32\columnwidth]{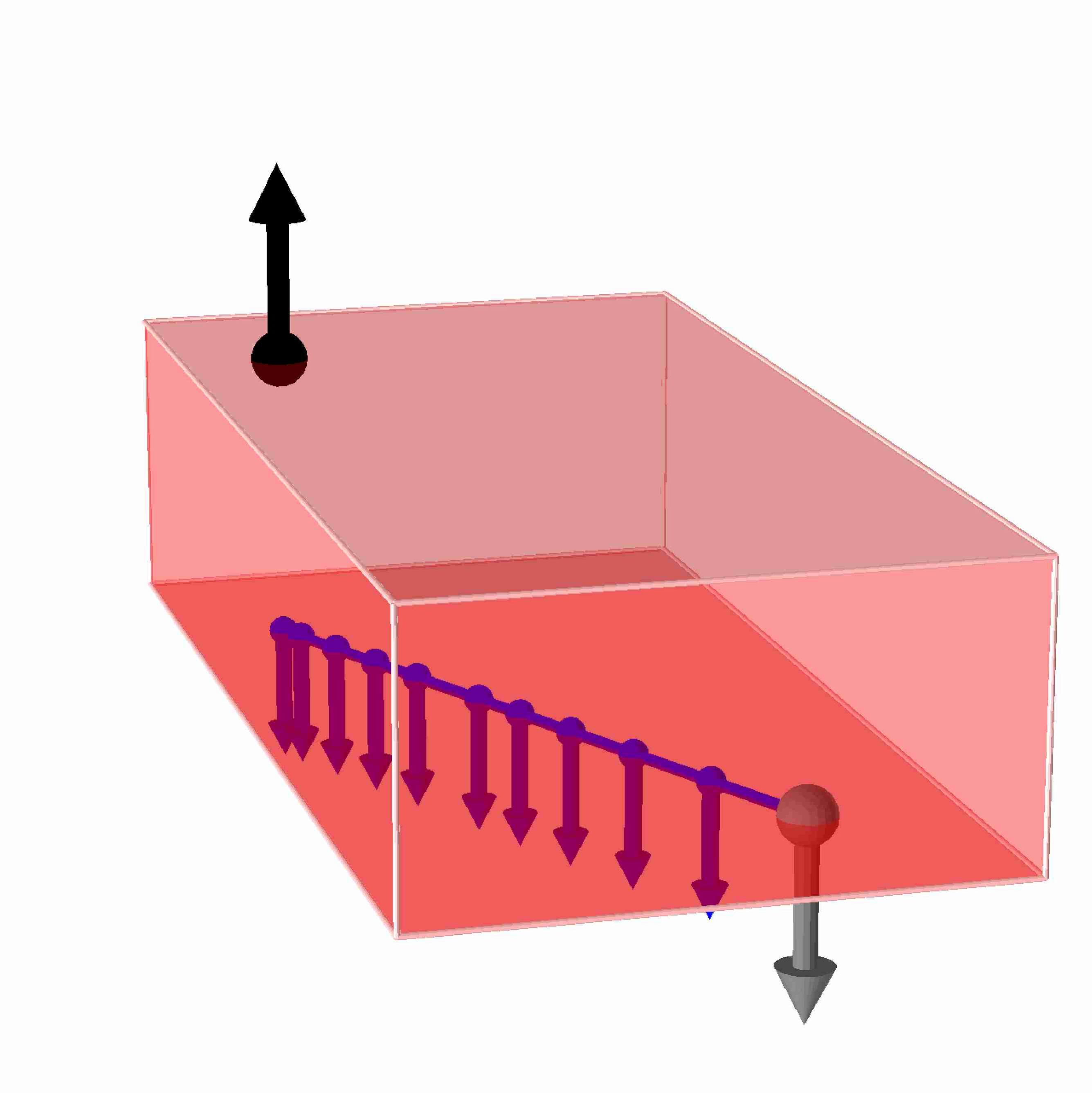}\\
  \vspace{-1mm}
  \begin{minipage}{0.34\columnwidth}
    \begin{center} \footnotesize (B1) small normal weight \end{center}
  \end{minipage}
  \begin{minipage}{0.28\columnwidth}
    \begin{center} \footnotesize (B2) large normal weight \end{center}
  \end{minipage}
  \begin{minipage}{0.34\columnwidth}
    \begin{center} \footnotesize (C) without normal smoothing \end{center}
  \end{minipage}
  \caption{Gradient-based contact point searching.
    \newline \footnotesize
    Gray and black markers show the initial and target contact point, respectively.
    Blue markers show the history of contact points in the optimization
    while the arrows show the smoothed normal.
    In (B1) and (B2), $w_{\mathit{nrm}}$ is $1$ and $10$, respectively.
  }
  \label{fig:sample-surface}
\end{figure}

The gradient-based contact point searching by SQP \eqref{eq:sqp-update}-\eqref{eq:sqp-constraint} is validated (\figref{fig:sample-surface}).
We deal with the problem of finding the contact point on the body surface closest to the target position and the normal direction for a single body without joints.
This corresponds to the case that $T=1$ and $\bm{\theta}$ is empty in eqs.~\eqref{eq:config-all},\eqref{eq:config-t}.
\figref{fig:sample-surface} (A1)-(A3) confirmed the convergence to a contact point that matches the target.
From an initial state with an error of about $2$ m, the error decreased to less than $0.1$ mm and $0.01$ degree after 200 iterations.

Even when the target was not on the body surface,
the closest contact point to the target was found as in \figref{fig:sample-surface} (B1),(B2).
The error value of the PN-task~\eqref{eq:pn-const} depends on the normal weight $w_{\mathit{nrm}}$.
In (B2), the weight is 10 times larger than (B1), and the contact point converges to a point where the normal is close to the target.

\figref{fig:sample-surface}~(C) shows the case where the conventional normal is used instead of the smoothed normal.
In this case,
the contact point fell to the local optimal without proceeding to the next face across the edge.
This is because the tangent plane beyond the edge can be predicted when using a smoothed normal that considers the normals of adjacent faces.

\subsection{Inverse Kinematics with Body Surface Contact}

\begin{figure}[thpb]
  \centering
  \includegraphics[width=0.32\columnwidth]{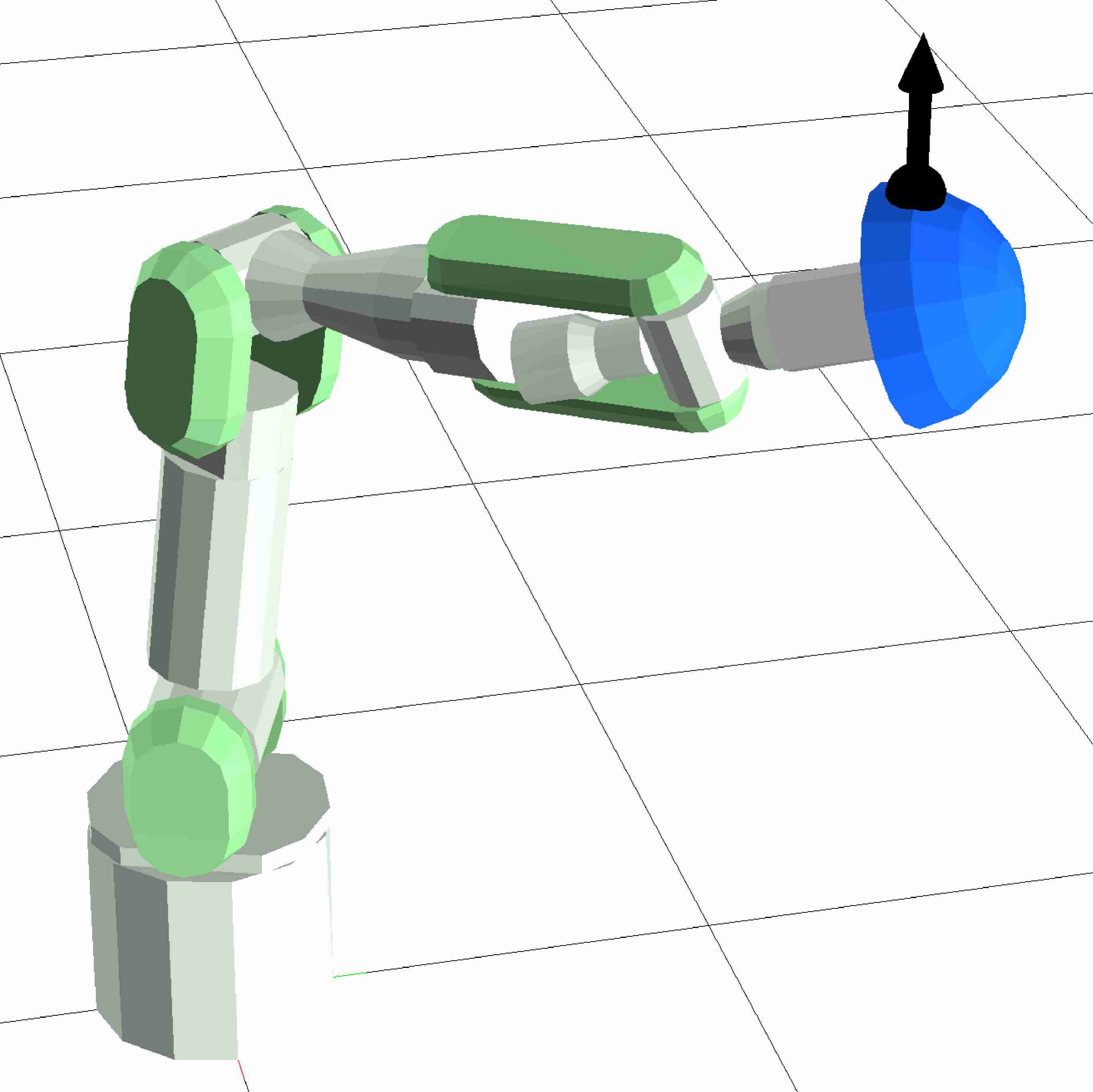}
  \includegraphics[width=0.32\columnwidth]{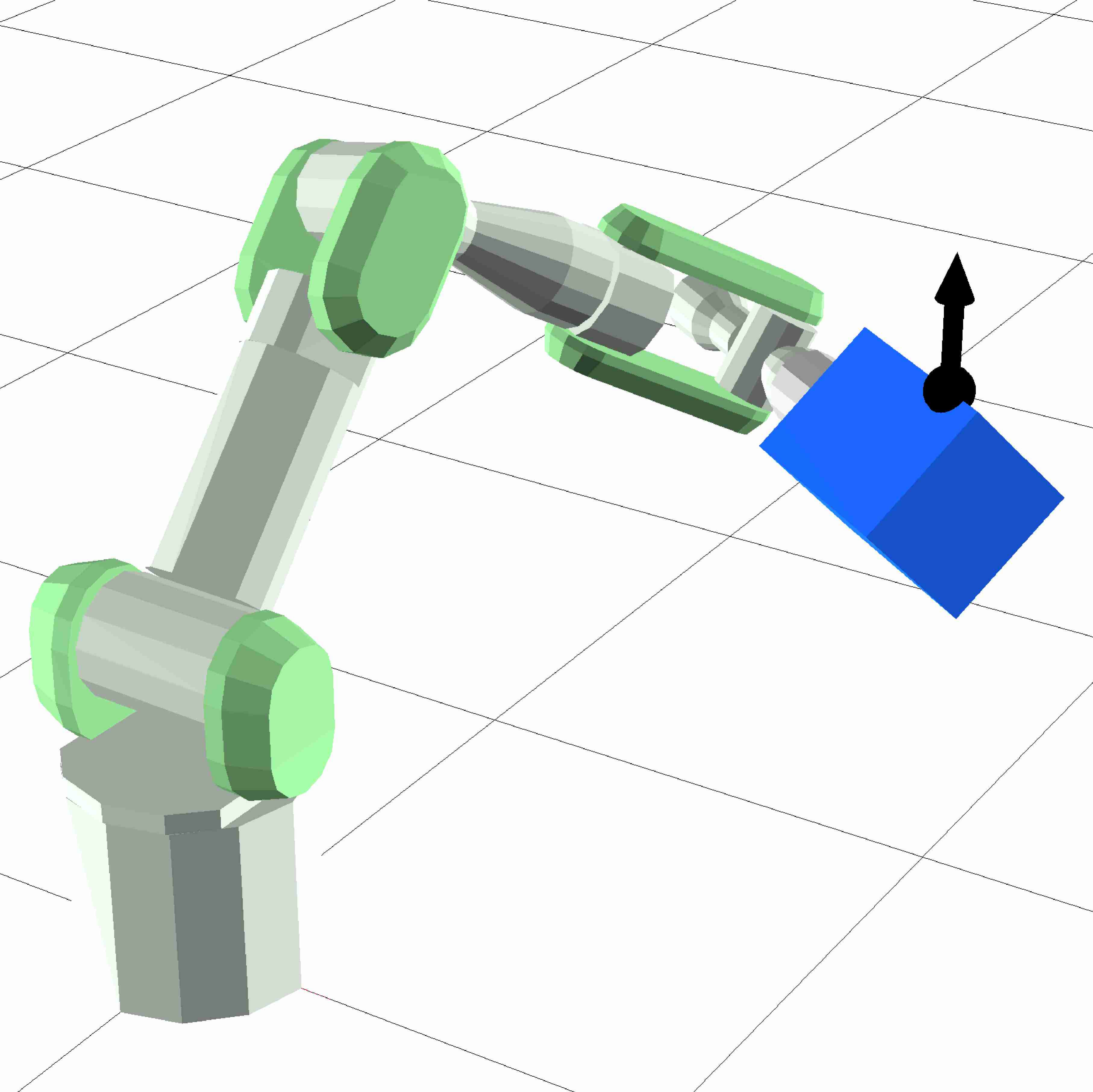}
  \includegraphics[width=0.32\columnwidth]{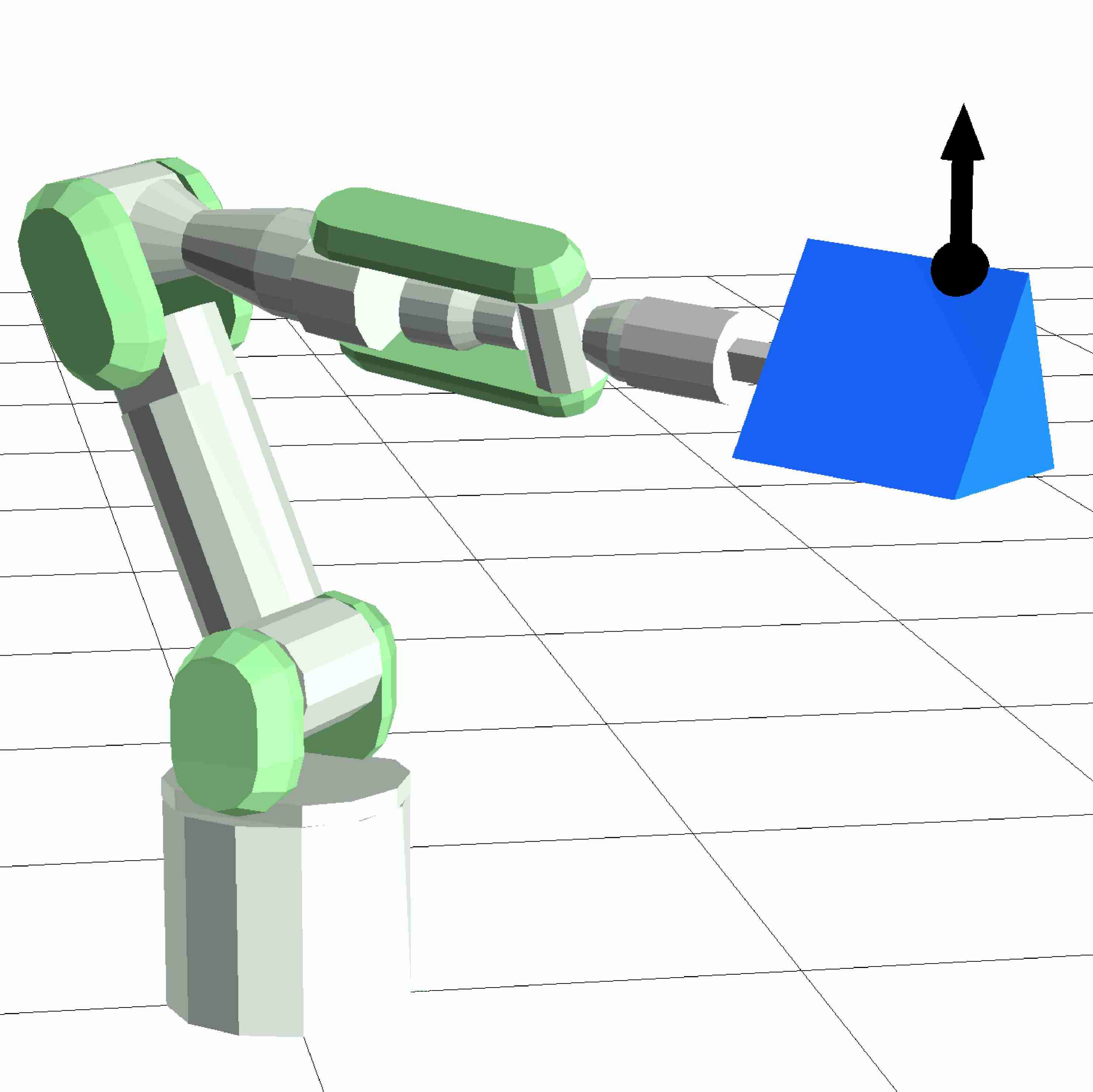}\\
  \vspace{-0.25mm}
  \begin{minipage}{0.32\columnwidth}
    \begin{center} \footnotesize hemisphere \end{center}
  \end{minipage}
  \begin{minipage}{0.32\columnwidth}
    \begin{center} \footnotesize cube \end{center}
  \end{minipage}
  \begin{minipage}{0.32\columnwidth}
    \begin{center} \footnotesize triangular prism \end{center}
  \end{minipage}
  \caption{Inverse kinematics with surface contact on tip body.
    \newline \footnotesize
    The tip body is shown in blue.
    Black markers show the target position and normal direction.
  }
  \label{fig:sample-ik-various-tip}
\end{figure}

The posture generation with contact point searching is validated by examples of the simple manipulator.
A body with various shapes was added to the tip of the manipulator PA-10, and a posture satisfying the PN-task~\eqref{eq:pn-const} was generated.
The joint position and contact point (\figref{fig:sample-ik-various-tip}) were calculated such that the error is less than $1$ mm and $0.1$ degree after 100 iterations.
The optimization time for 100 iterations is presented in \tabref{tab:computation-time}.
As presented in \tabref{tab:computation-time}, it took several milliseconds per iteration,
but when the body shape was complicated, it took additional time to calculate the gradient of eq. \eqref{eq:contact-jacobi-normal}.

\begin{figure}[thpb]
  \centering
  \includegraphics[width=0.40\columnwidth]{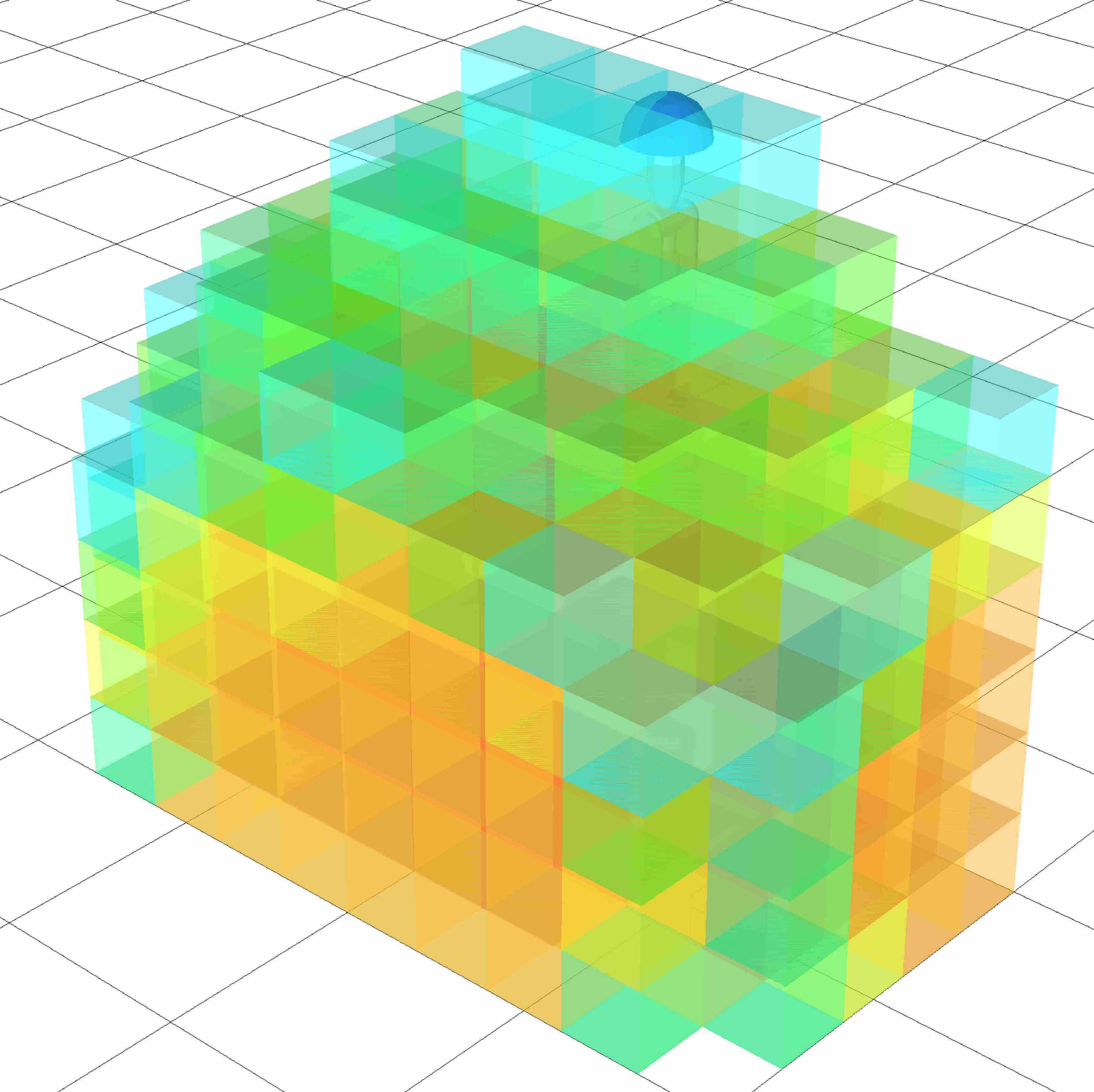}
  \includegraphics[width=0.40\columnwidth]{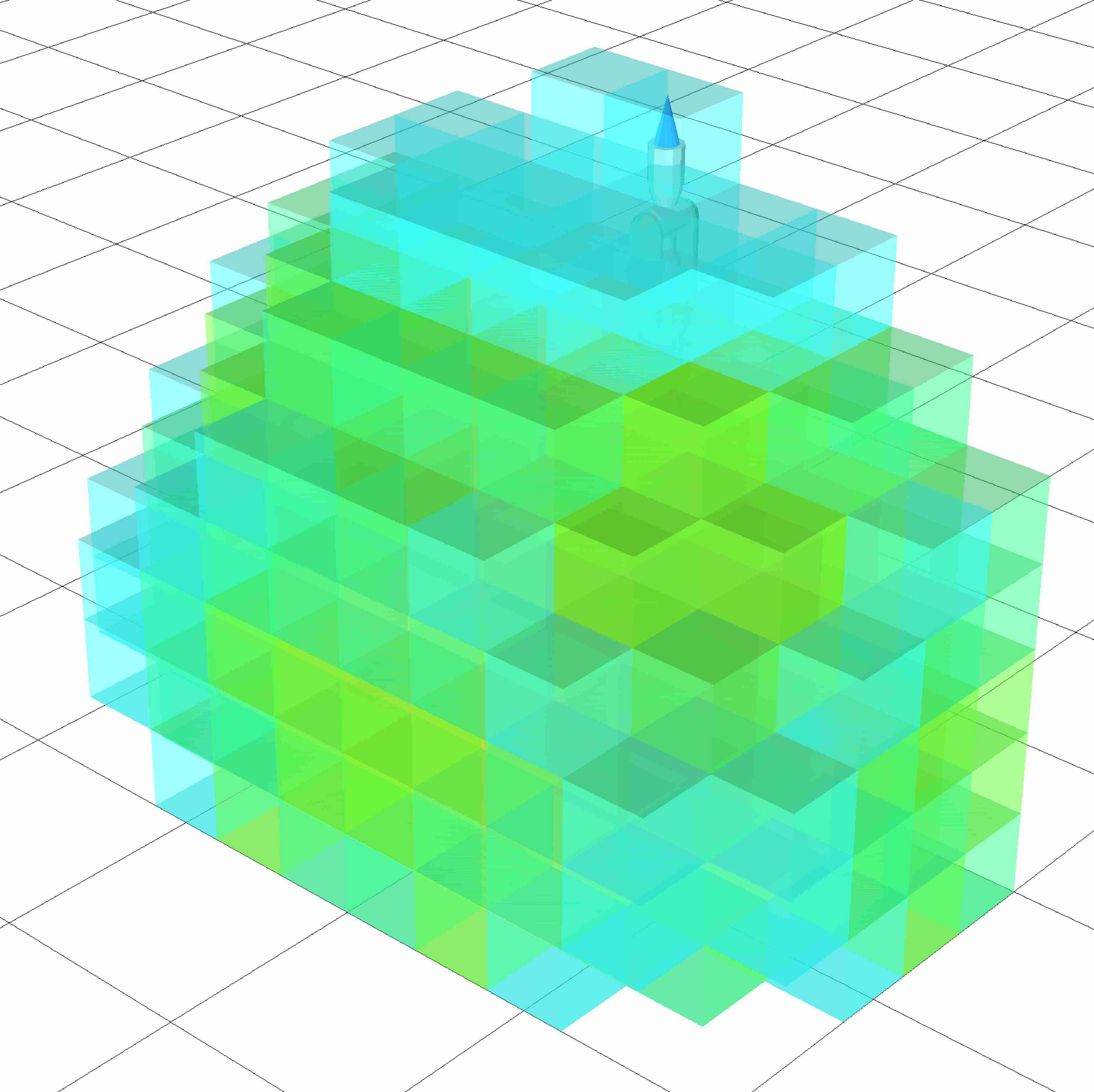}
  \begin{minipage}{0.12\columnwidth}
    \begin{center} \ \  \end{center}
  \end{minipage}\\
  \includegraphics[width=0.40\columnwidth]{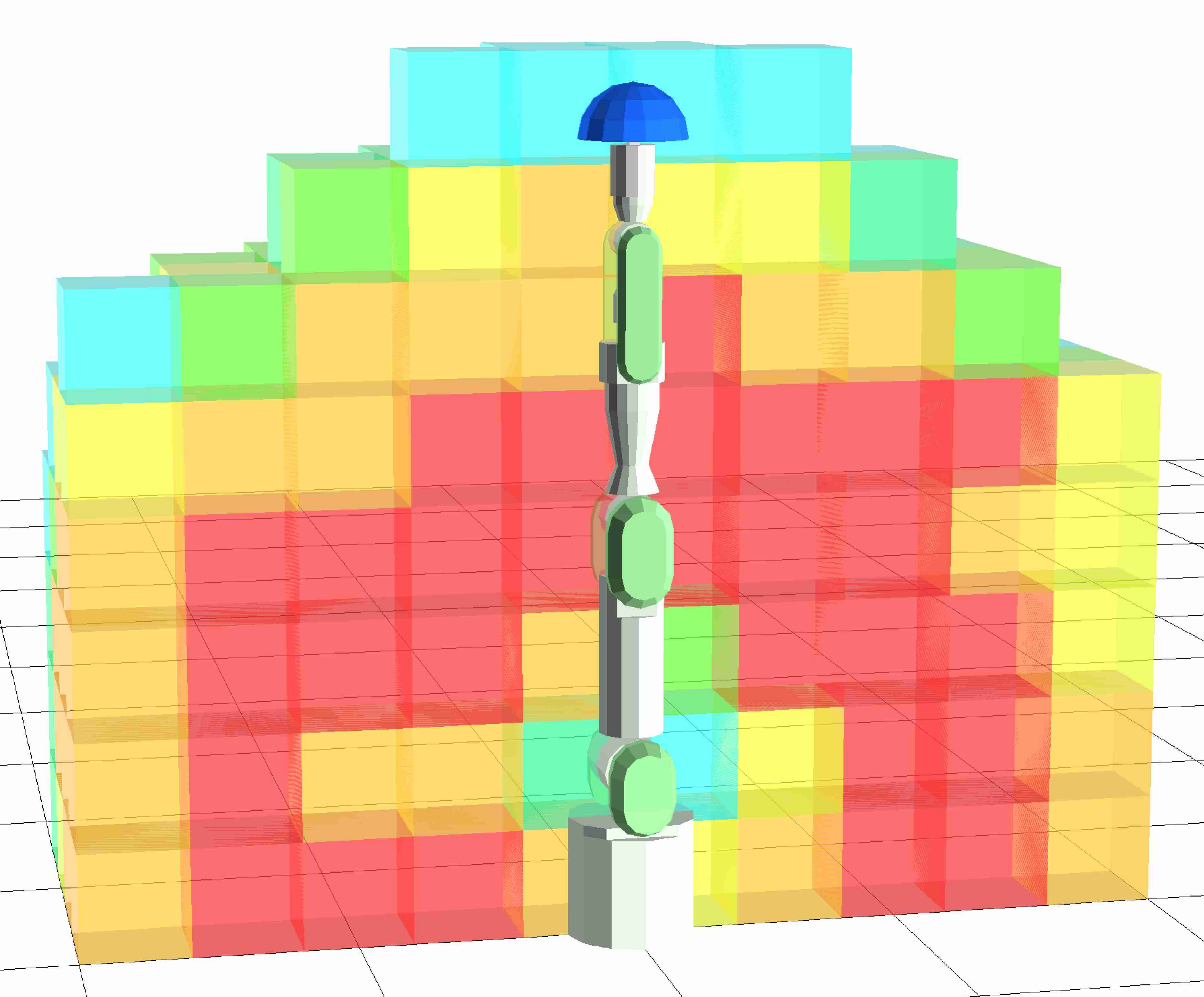}
  \includegraphics[width=0.40\columnwidth]{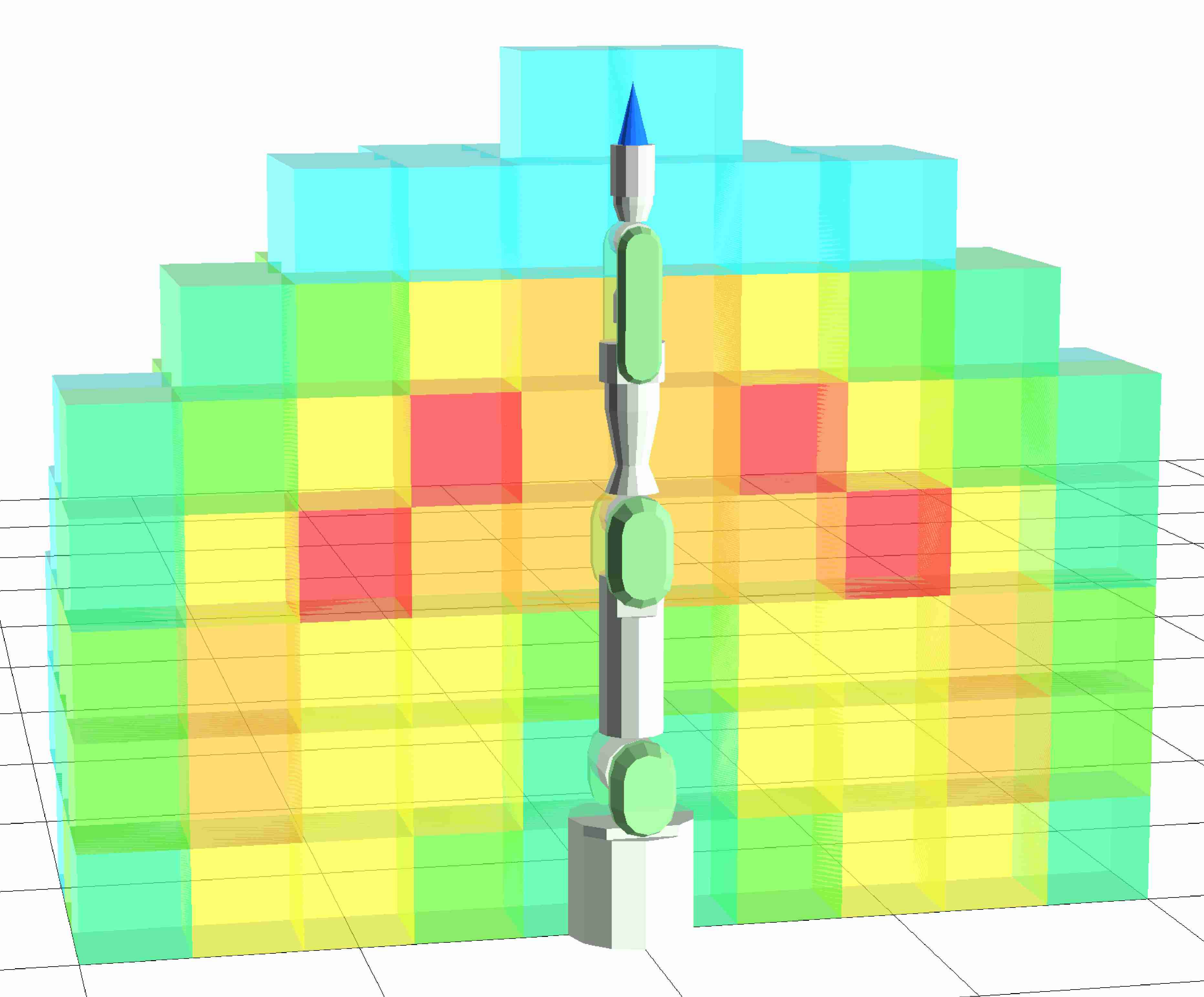}
  \includegraphics[width=0.12\columnwidth]{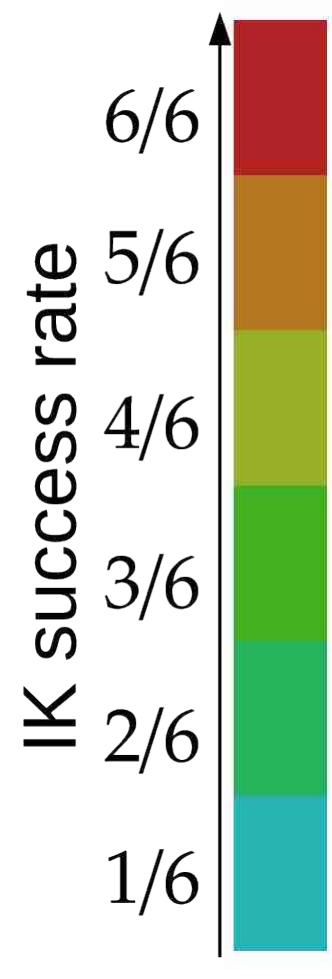}\\
  \vspace{-0.25mm}
  \begin{minipage}{0.40\columnwidth}
    \begin{center} \footnotesize with tip body \end{center}
  \end{minipage}
  \begin{minipage}{0.40\columnwidth}
    \begin{center} \footnotesize without tip body \end{center}
  \end{minipage}
  \begin{minipage}{0.12\columnwidth}
    \begin{center} \footnotesize \ \  \end{center}
  \end{minipage}
  \caption{Comparison of solvability with and without a tip body.
    \newline \footnotesize
    For each grid center, the inverse kinematics is calculated with six target normal directions (i.e., $\pm {\rm x}, \pm {\rm y}, \pm {\rm z}$).
  }
  \label{fig:sample-ik-grid-test}
\end{figure}

We compared the solvability range and ratio with and without the contact body attached to the tip of the manipulator.
As shown in \figref{fig:sample-ik-grid-test}, the region with high solvability is larger when contacting with the hemispherical body surface.
This shows that the body surface contact is effective when, for example, the robot needs to push buttons placed at various positions and orientations.

\begin{figure}[thpb]
  \centering
  \includegraphics[width=0.49\columnwidth]{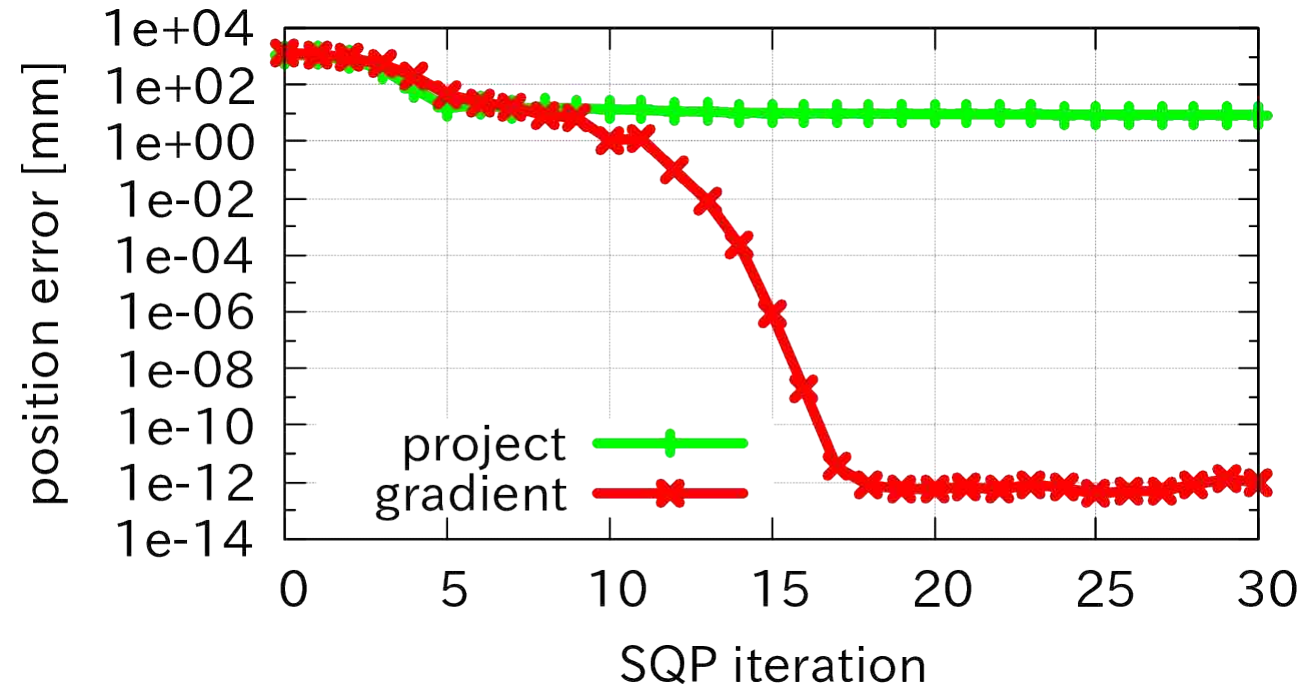}
  \includegraphics[width=0.49\columnwidth]{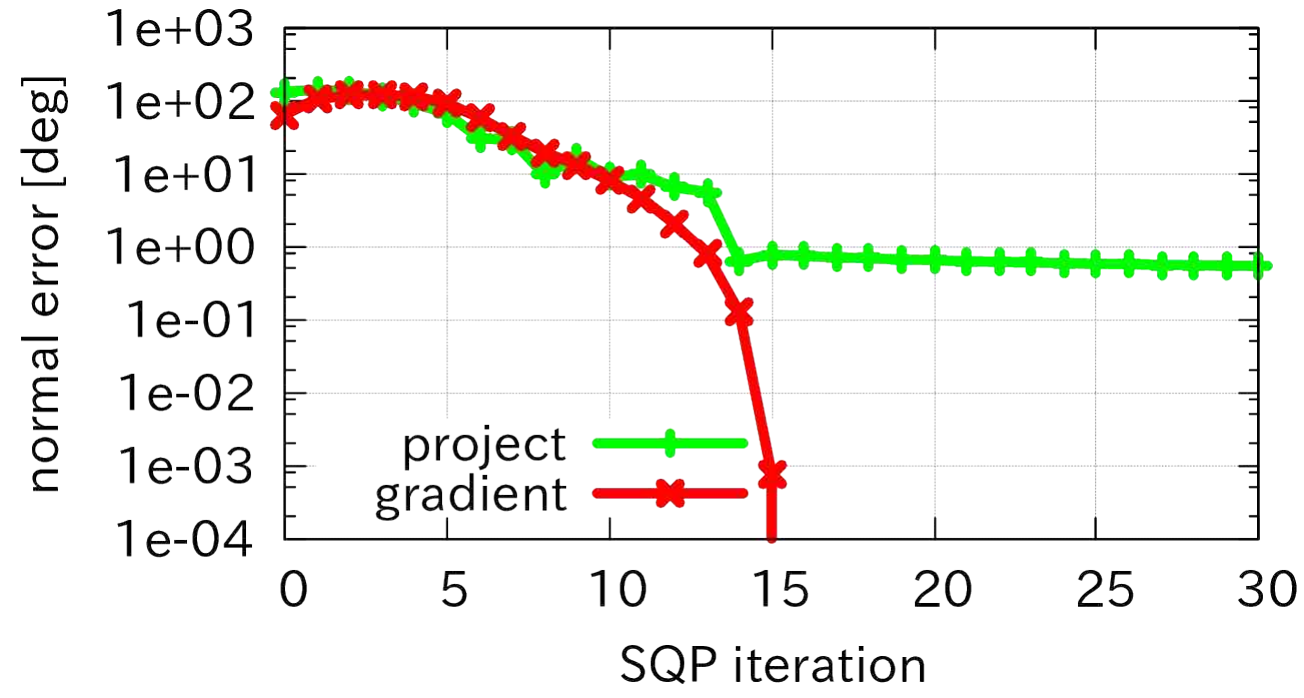}
  \vspace{-6mm}
  \caption{Error in position and normal direction.
    \newline \footnotesize
    The baseline method (project) and the proposed method (gradient) are compared.
    The normal direction error of the proposed method turned to zero after 15 iterations because of computer calculation accuracy.
  }
  \label{fig:sample-ik-compare-projection}
\end{figure}

We also verified the convergence performance,
with \figref{fig:sample-ik-compare-projection} depicting the position and normal errors for the proposed and baseline methods.
In the baseline method, the contact point on the tip body is set as the point closest to the target position for every SQP iteration without introducing contact configurations.
The proposed method that explicitly considers the gradient of the contact point converges to a smaller error value (\figref{fig:sample-ik-compare-projection}).
This is because the baseline method, in which the contact point is projected at each iteration, renders the gradient inaccurate without considering the projection effect.

\begin{table}[tbh]
  \begin{center}
    \caption{Computation Time of Inverse Kinematics in \figref{fig:sample-ik-various-tip}}
    \label{tab:computation-time}
    \renewcommand{\arraystretch}{1.3}
    \begin{tabular}{l||p{0.18\columnwidth}|p{0.18\columnwidth}}
      \hline
       & hemisphere & triangular prism \\ \hline
      total &
      0.66\,s [100\,\%] & 0.23\,s [100\,\%] \\
      \ \ calculate matrices in QP \eqref{eq:sqp-qp} &
      0.54\,s [\ 82\,\%] & 0.20\,s [\ 87\,\%] \\
      \ \ \ \ calculate $\frac{\partial \bm{e}}{\partial \bm{\theta}}$ in eq. \eqref{eq:theta-jacobi} &
      0.00\,s [\ \ 1\,\%] & 0.00\,s [\ \ 2\,\%] \\
      \ \ \ \ calculate $\frac{\partial \bm{e}}{\partial \bm{u}}$ in eq. \eqref{eq:contact-jacobi} &
      0.39\,s [\ 59\,\%] & 0.06\,s [\ 25\,\%] \\
      \ \ \ \ calculate $\bm{c}(\bm{\hat{q}}), \nabla \bm{c}(\bm{\hat{q}})$ in eq. \eqref{eq:sqp-constraint} &
      0.06\,s [\ \ 9\,\%] & 0.06\,s [\ 28\,\%] \\
      \ \ solve QP \eqref{eq:sqp-qp} &
      0.12\,s [\ 18\,\%] & 0.03\,s [\ 12\,\%] \\
      \hline
    \end{tabular}
  \end{center}
\end{table}

\section{Application to Whole-body Contact Motion}

\subsection{Object Holding with Dual Arms}
\begin{figure}[thpb]
  \centering
  \includegraphics[height=0.48\columnwidth]{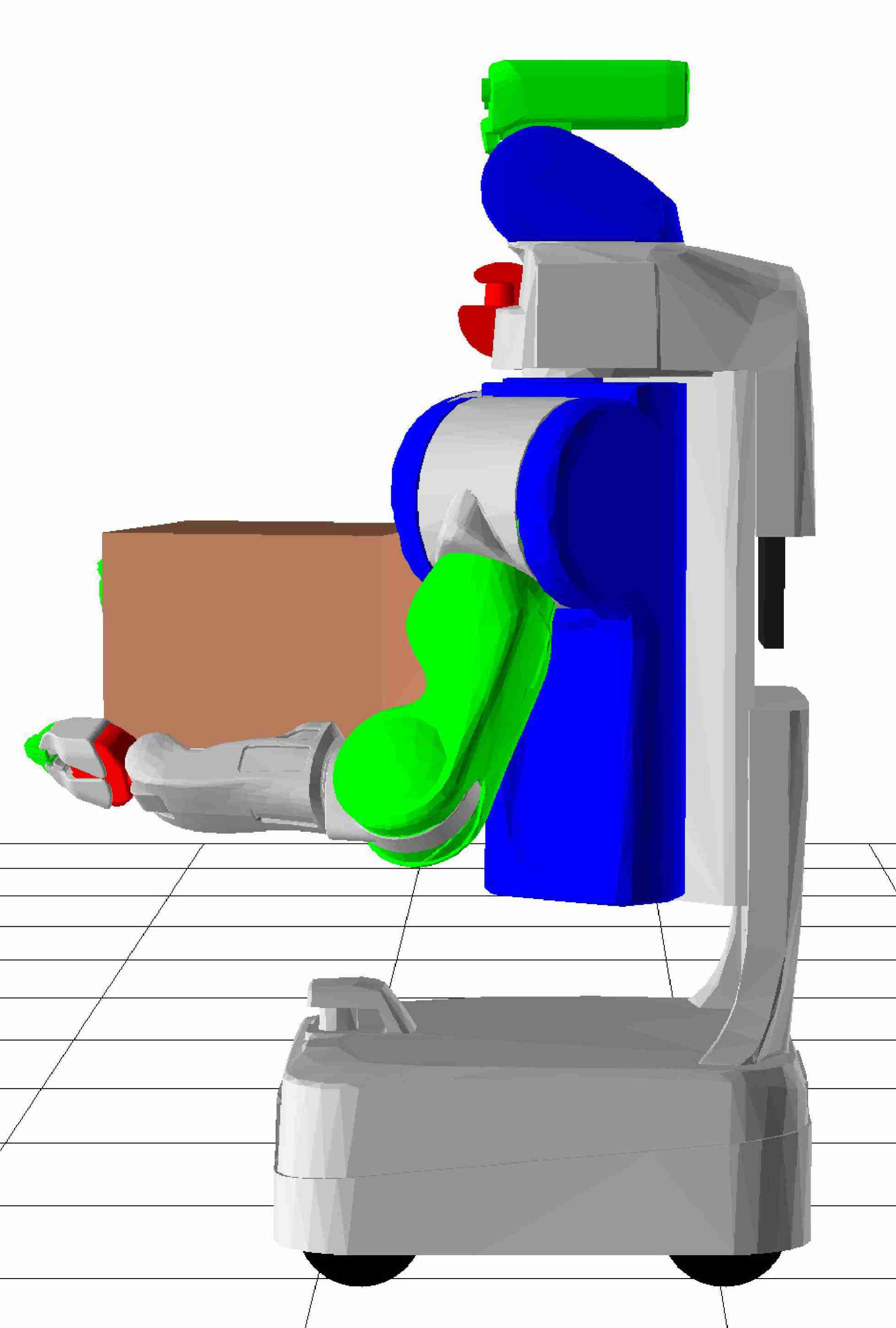}
  \includegraphics[height=0.48\columnwidth]{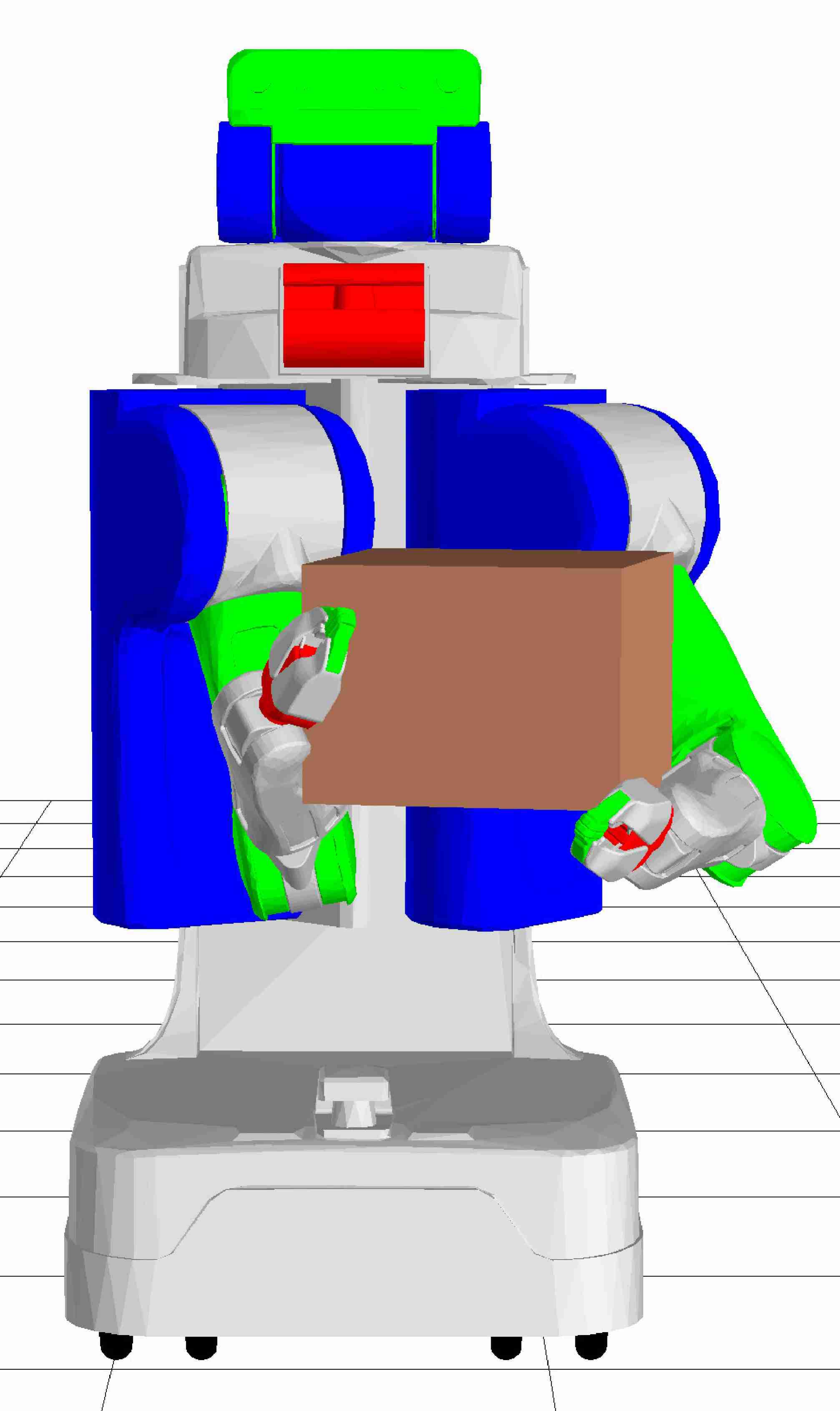}
  \includegraphics[height=0.48\columnwidth]{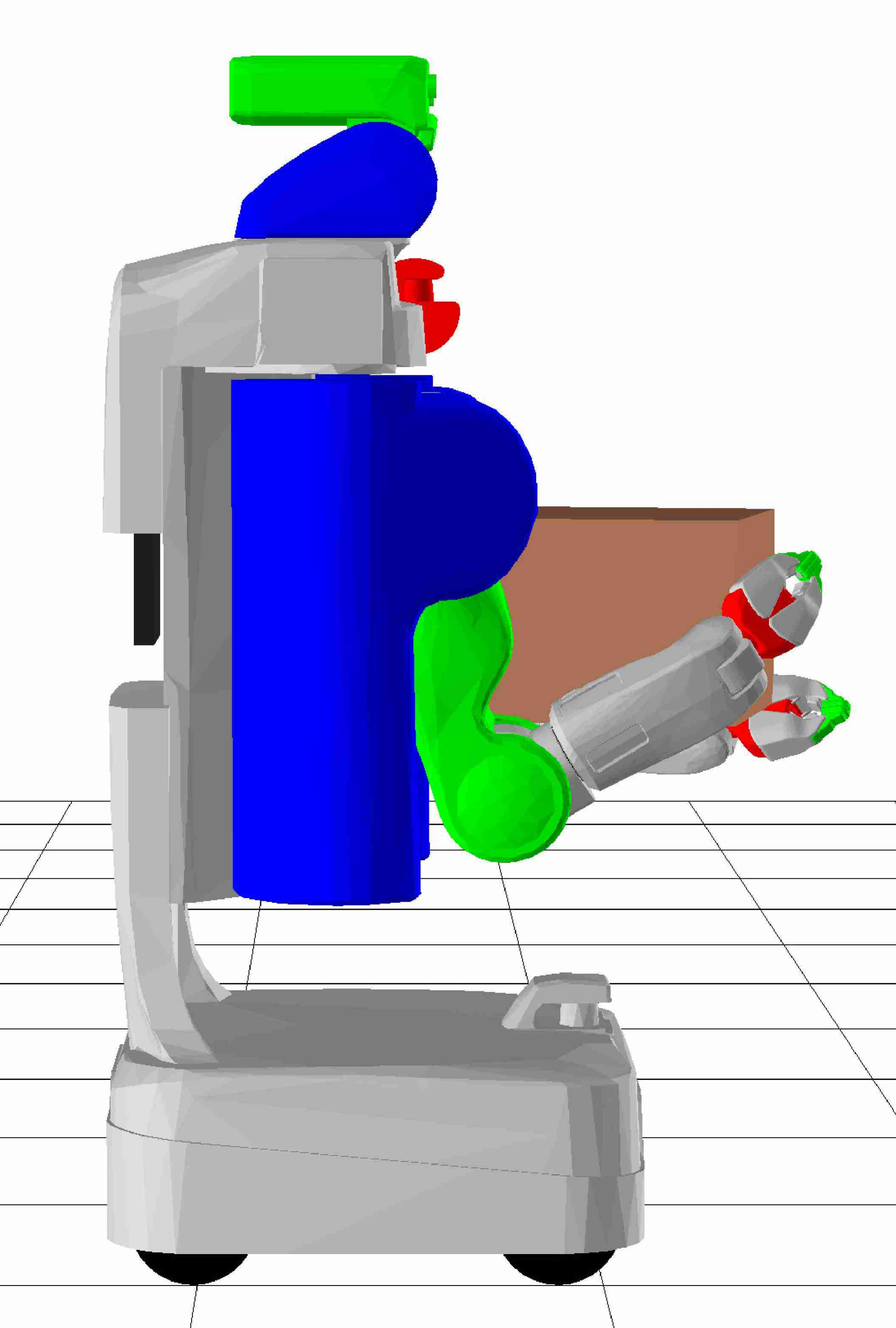}\\
  \includegraphics[height=0.48\columnwidth]{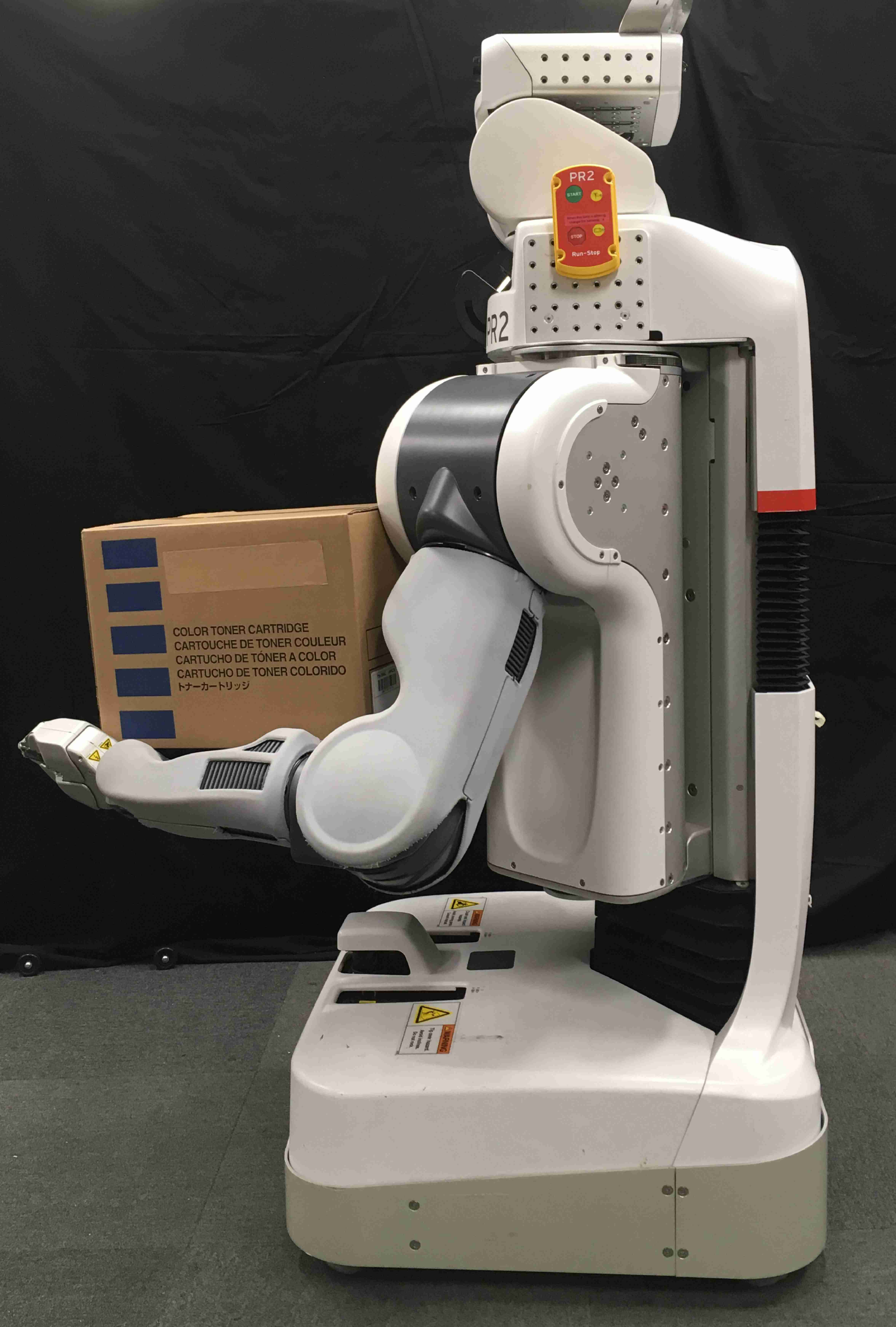}
  \includegraphics[height=0.48\columnwidth]{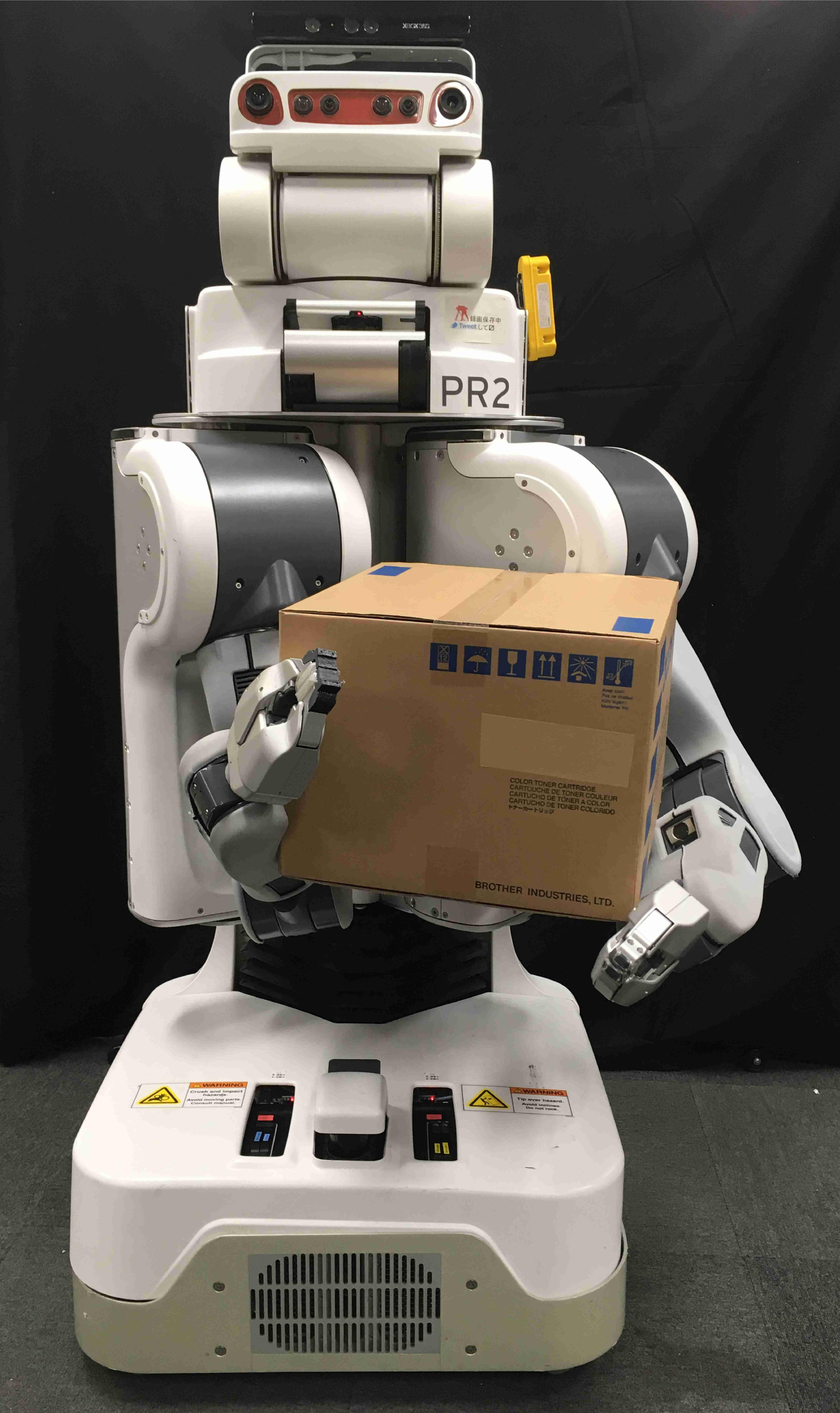}
  \includegraphics[height=0.48\columnwidth]{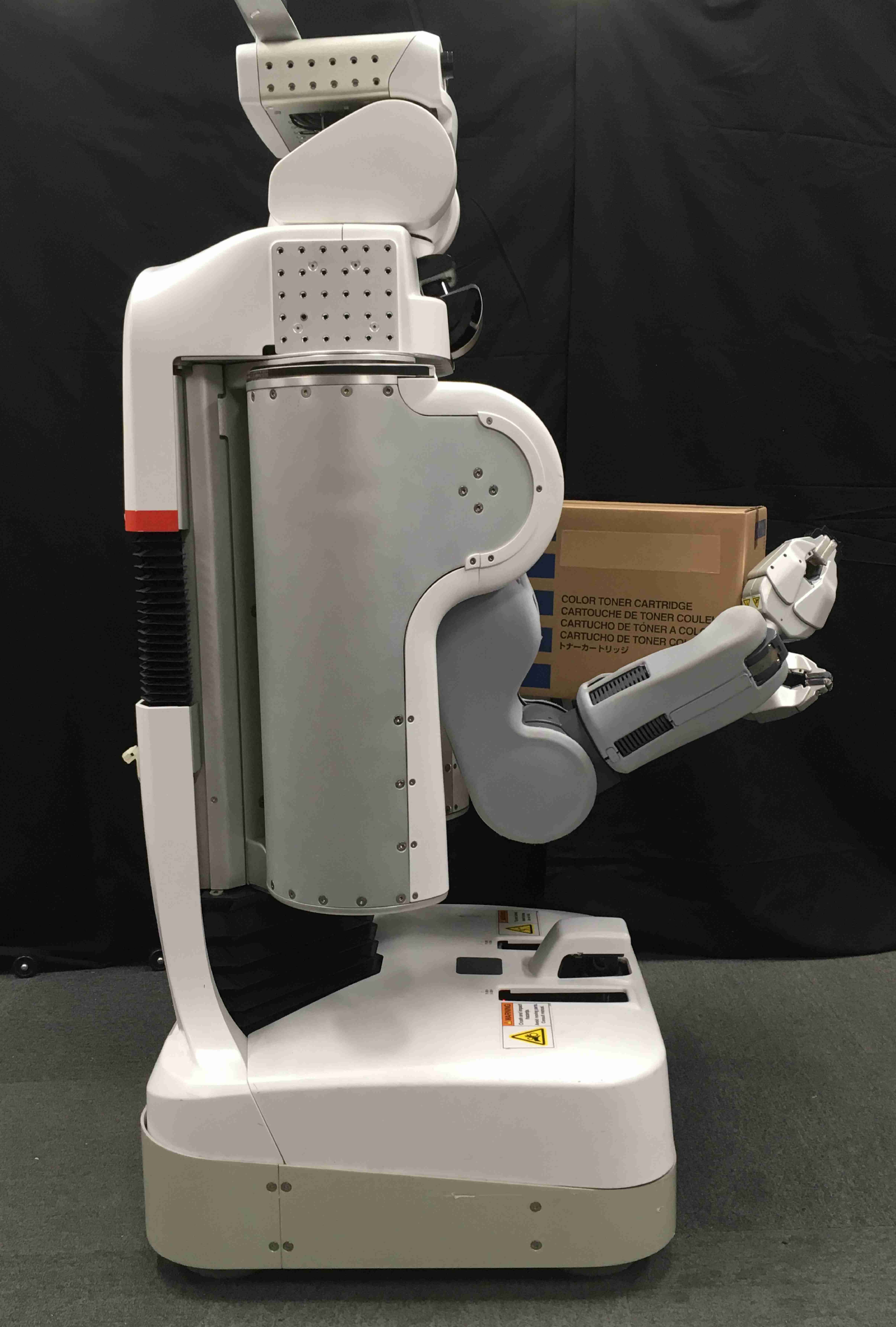}\\
  \vspace{-0.5mm}
  \begin{minipage}{0.34\columnwidth}
    \begin{center} \footnotesize left side \end{center}
  \end{minipage}
  \begin{minipage}{0.28\columnwidth}
    \begin{center} \footnotesize front \end{center}
  \end{minipage}
  \begin{minipage}{0.34\columnwidth}
    \begin{center} \footnotesize right side \end{center}
  \end{minipage}
  \caption{Object holding with dual arms.
    \newline \footnotesize
    The mobile manipulator PR2 holds the object in contact with six links of dual arms.
    The top is an automatically generated posture while the bottom shows the execution with the real robot.
  }
  \label{fig:demo-pr2}
\end{figure}

We generated a posture of whole-body manipulation, in which a mobile manipulator PR2 holds a box-shaped object with dual arms (\figref{fig:demo-pr2}).
Since the weight of the object is $3$ kg,
insufficient friction makes it difficult to grasp with just the tip of both grippers, and it is therefore necessary to grasp with whole-arm links.

\begin{figure}[thpb]
  \centering
  \begin{tabular}{ccc}
    \multirow{3}{*}[15mm]{\includegraphics[height=0.5\columnwidth]{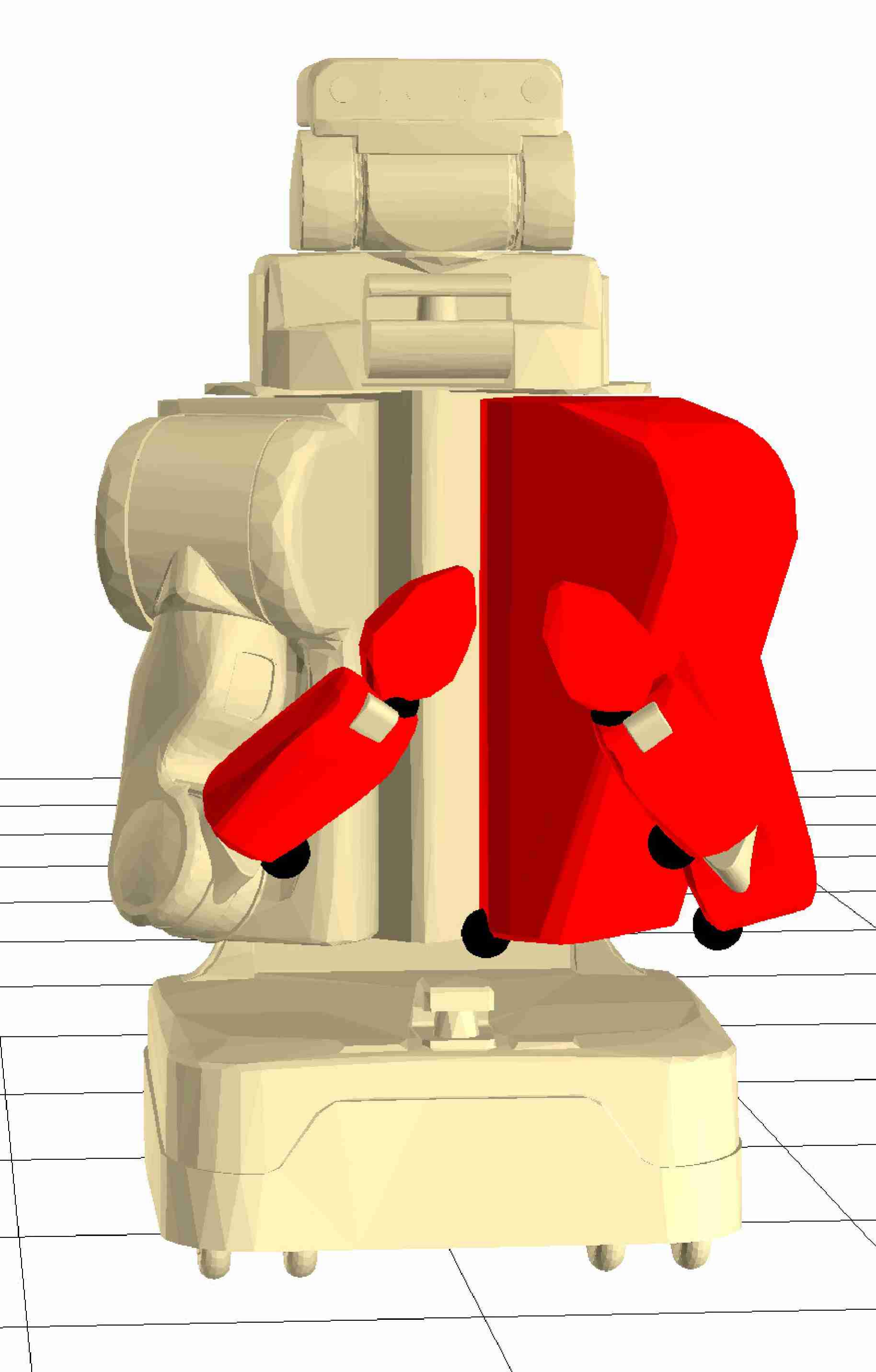}} &
    \multirow{3}{*}[15mm]{\includegraphics[height=0.5\columnwidth]{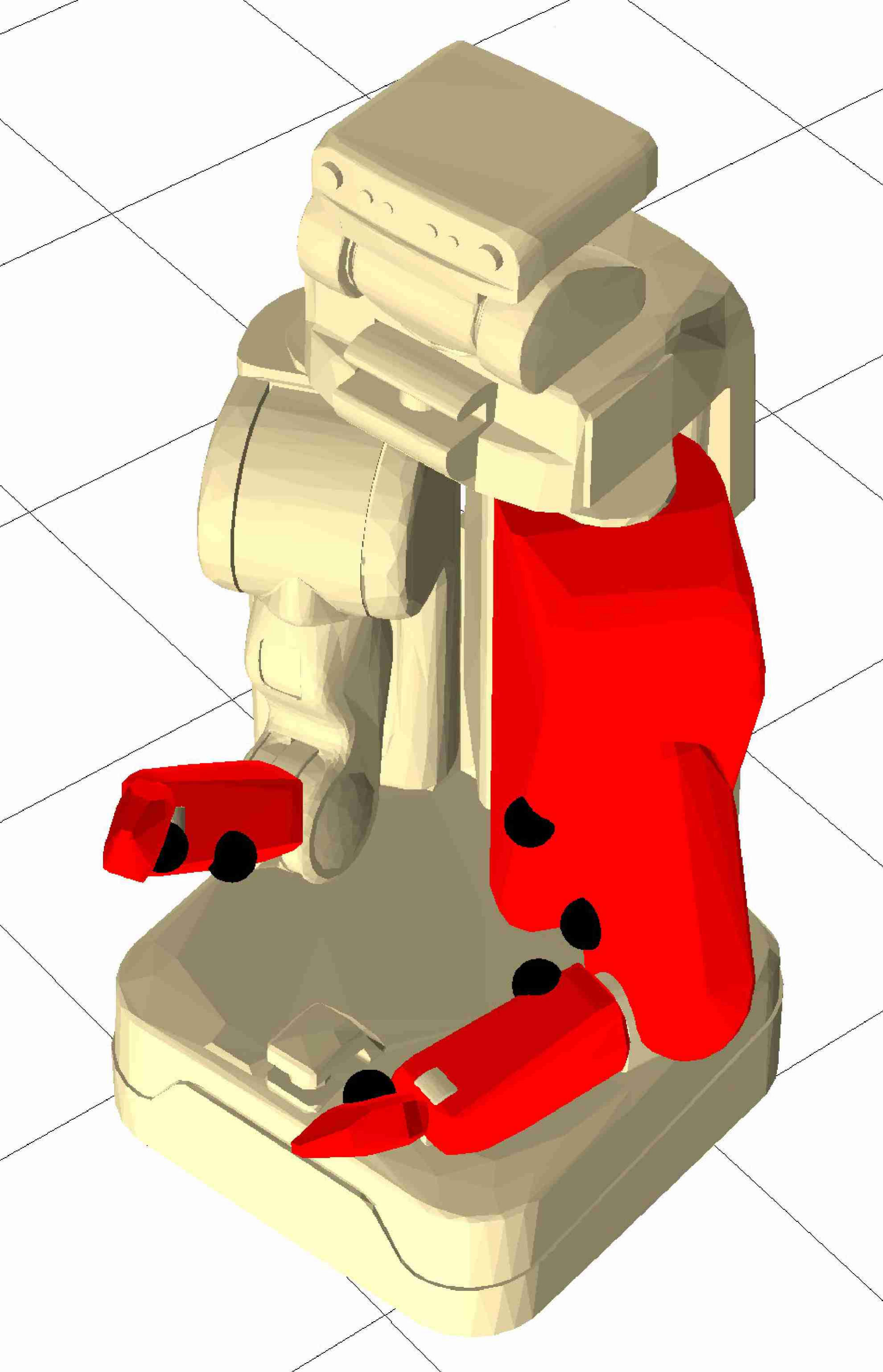}} &
    \includegraphics[height=0.22\columnwidth]{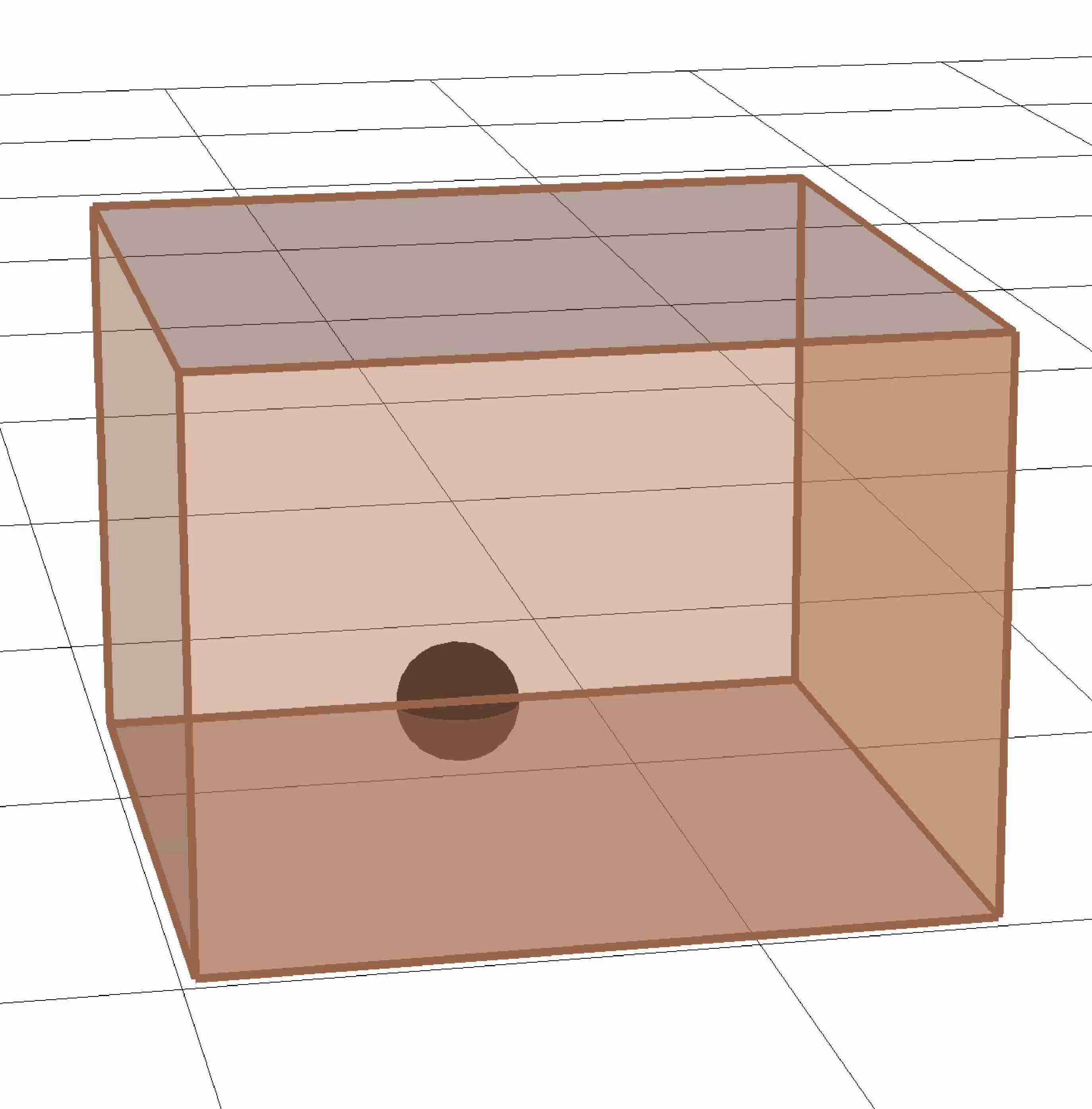} \\
    && \footnotesize \makecell{(C) object initial\\ contact} \\
    &&
    \includegraphics[height=0.22\columnwidth]{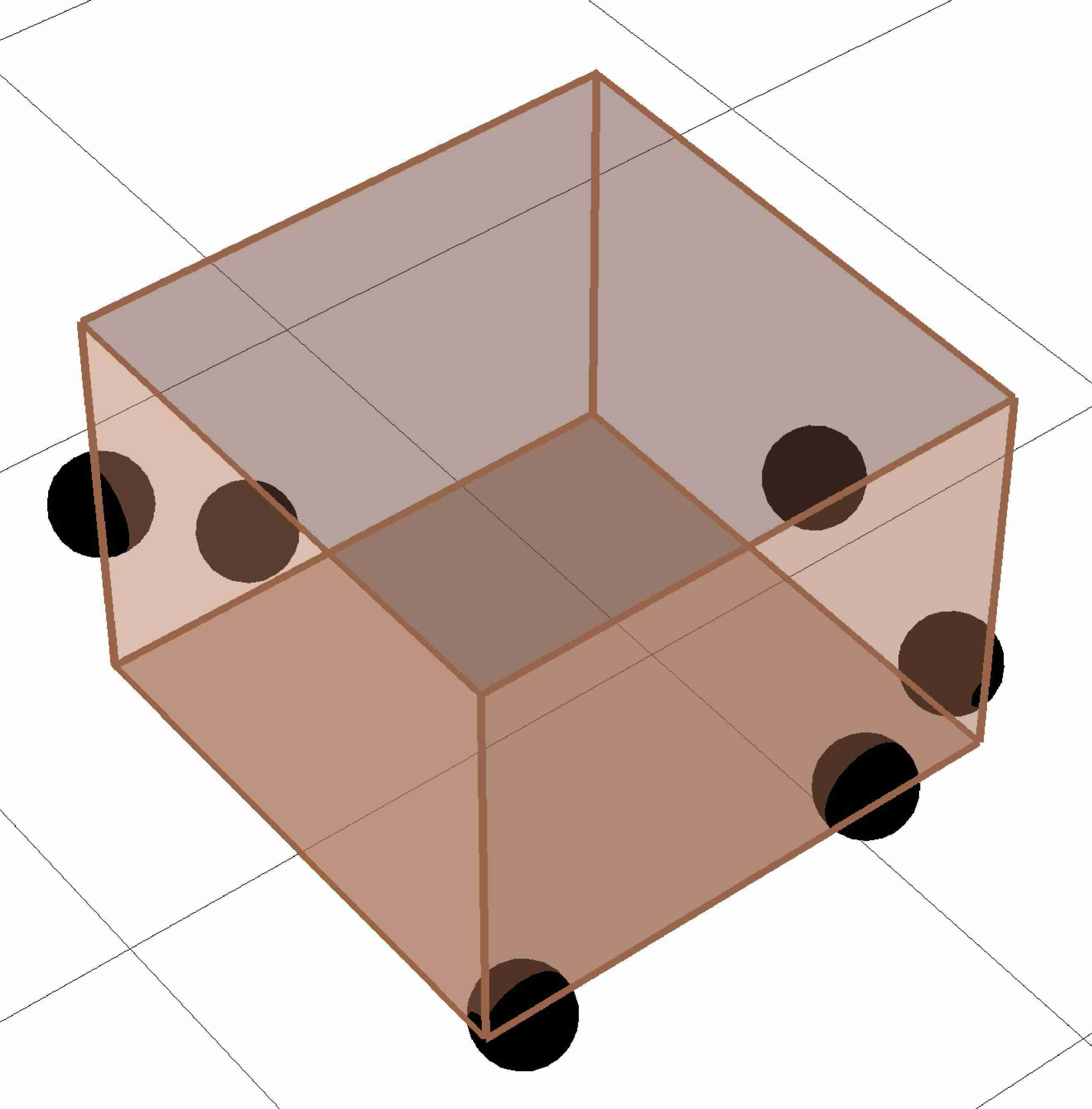} \\
    \footnotesize (A) robot initial contact & \footnotesize (B) robot final contact & \footnotesize \makecell{(D) object final\\ contact}
  \end{tabular}
  \caption{Contact points in dual-arm holding posture.
    \newline \footnotesize
    The figure shows the robot posture and contact points (shown as black points) in the initial iteration and 200 iterations when generating the posture of \figref{fig:demo-pr2}.
    The contact links are drawn in red.
  }
  \label{fig:demo-pr2-surface}
\end{figure}

\begin{figure}[thpb]
  \centering
  \includegraphics[width=0.22\columnwidth]{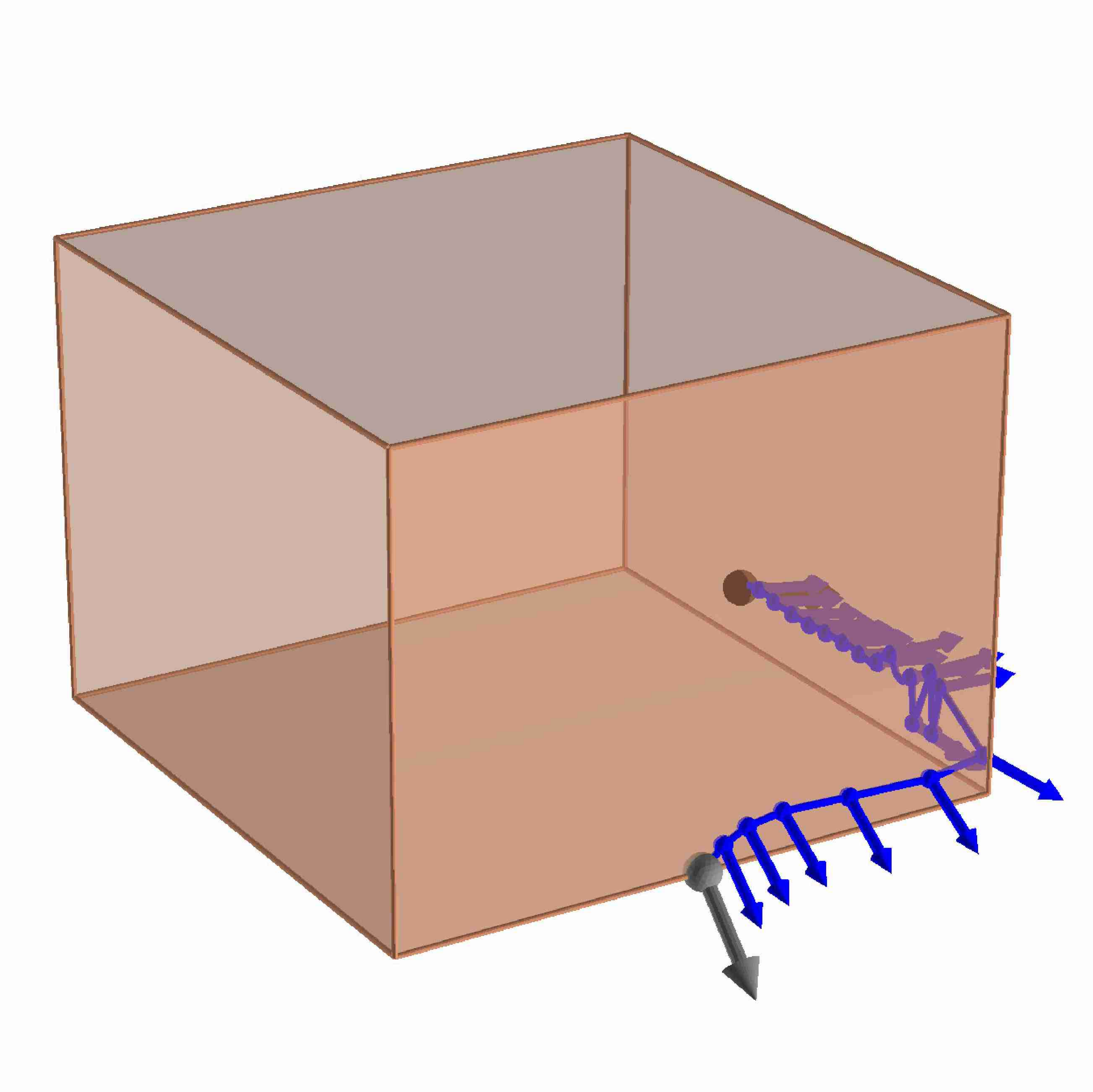}
  \includegraphics[width=0.24\columnwidth]{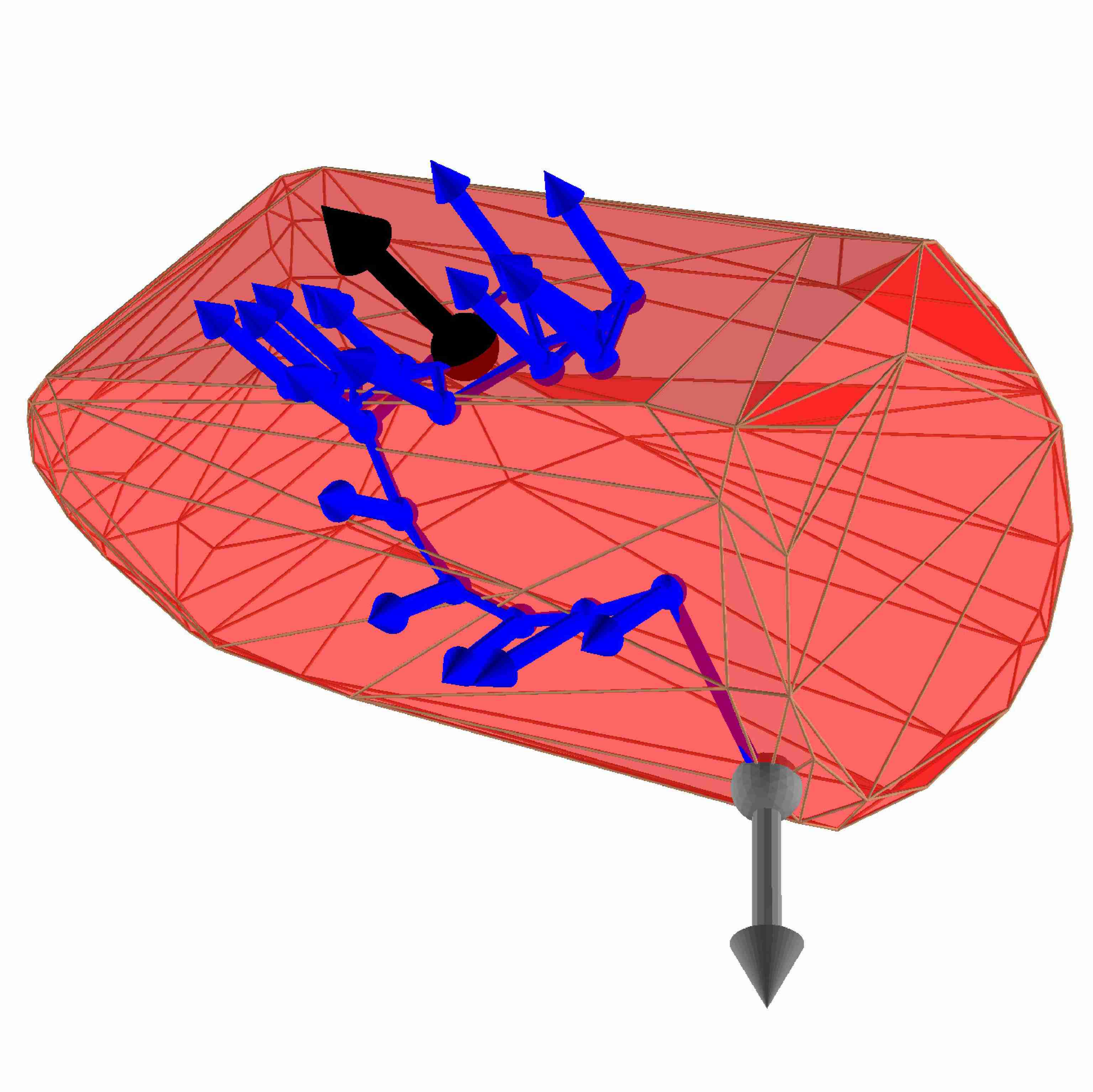}
  \hspace{2mm}
  \includegraphics[width=0.22\columnwidth]{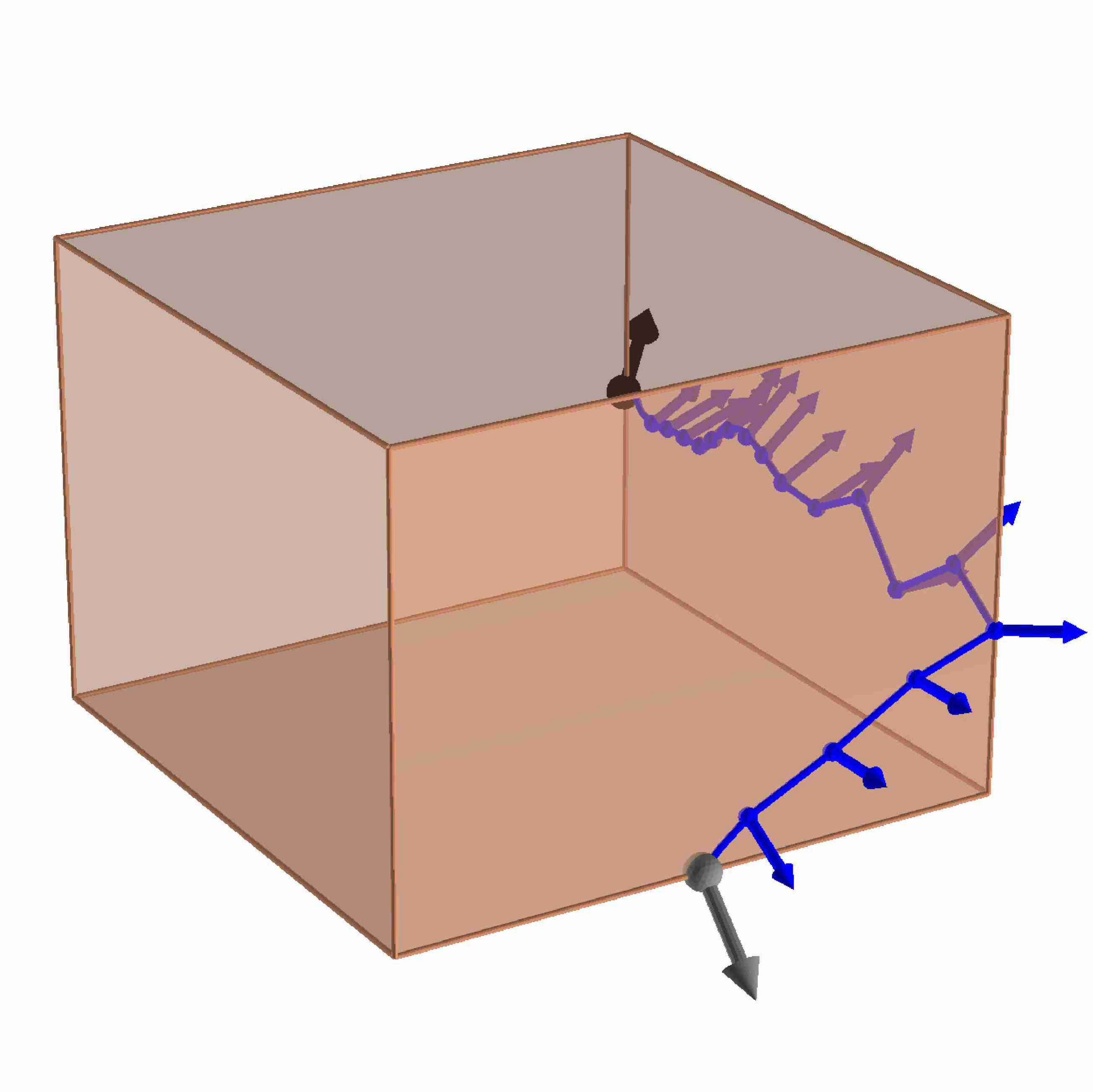}
  \includegraphics[width=0.24\columnwidth]{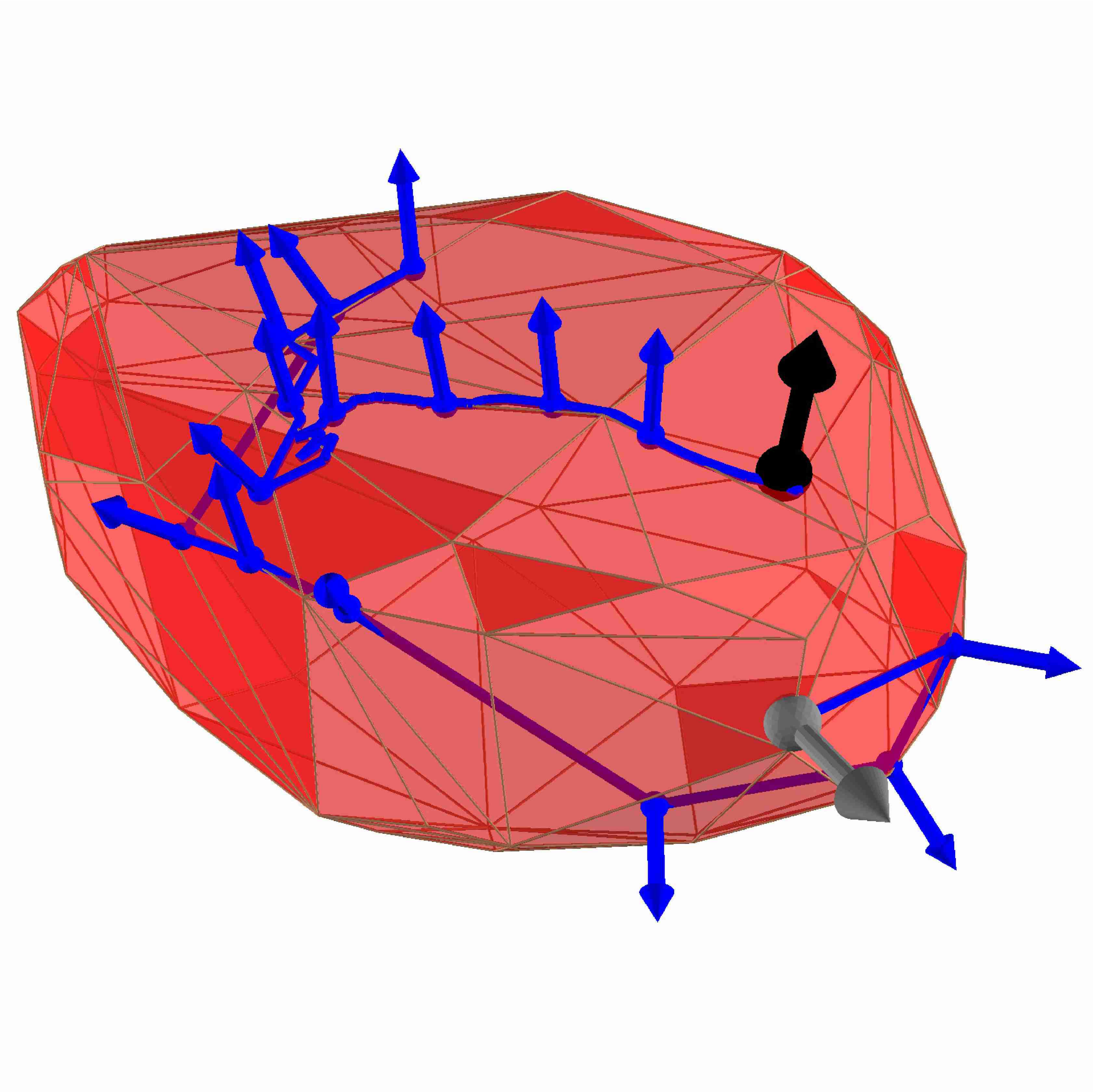}\\
  \vspace{-2.0mm}
  \begin{minipage}{0.49\columnwidth}
    \begin{center} \footnotesize (A) right forearm \end{center}
  \end{minipage}
  \begin{minipage}{0.49\columnwidth}
    \begin{center} \footnotesize (B) right gripper \end{center}
  \end{minipage}\\
  \vspace{0.5mm}
  \includegraphics[width=0.22\columnwidth]{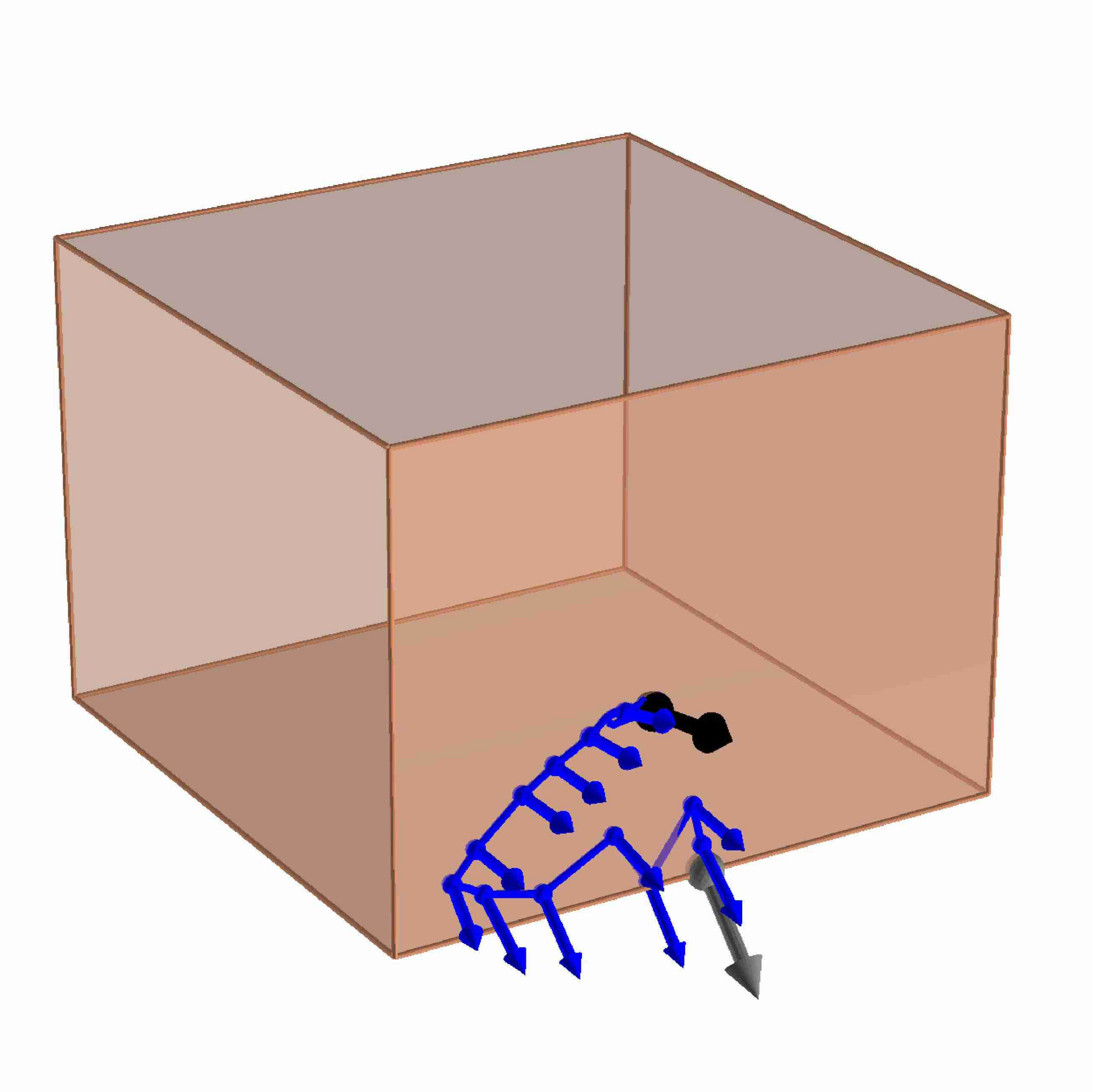}
  \includegraphics[width=0.24\columnwidth]{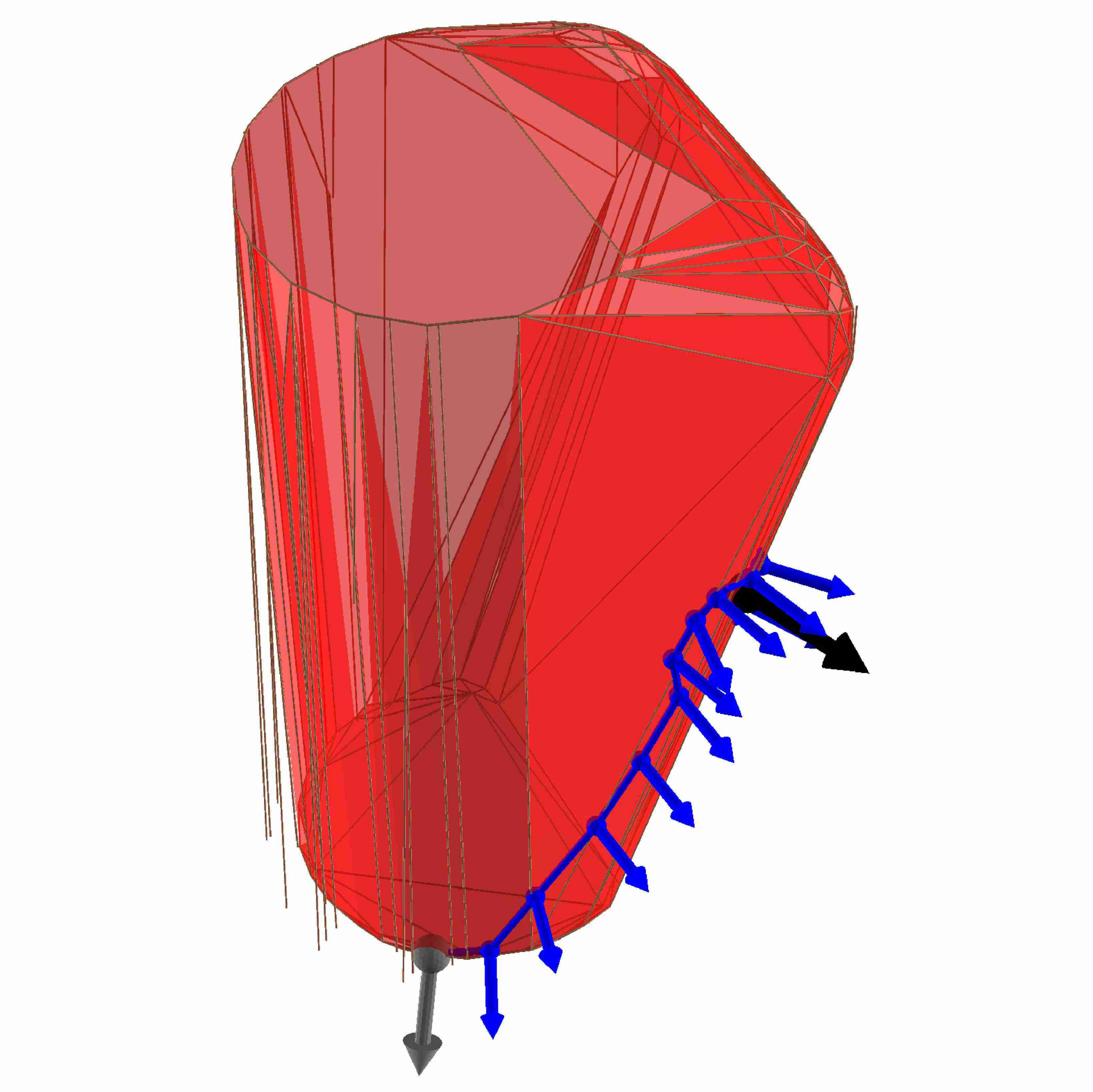}
  \hspace{2mm}
  \includegraphics[width=0.22\columnwidth]{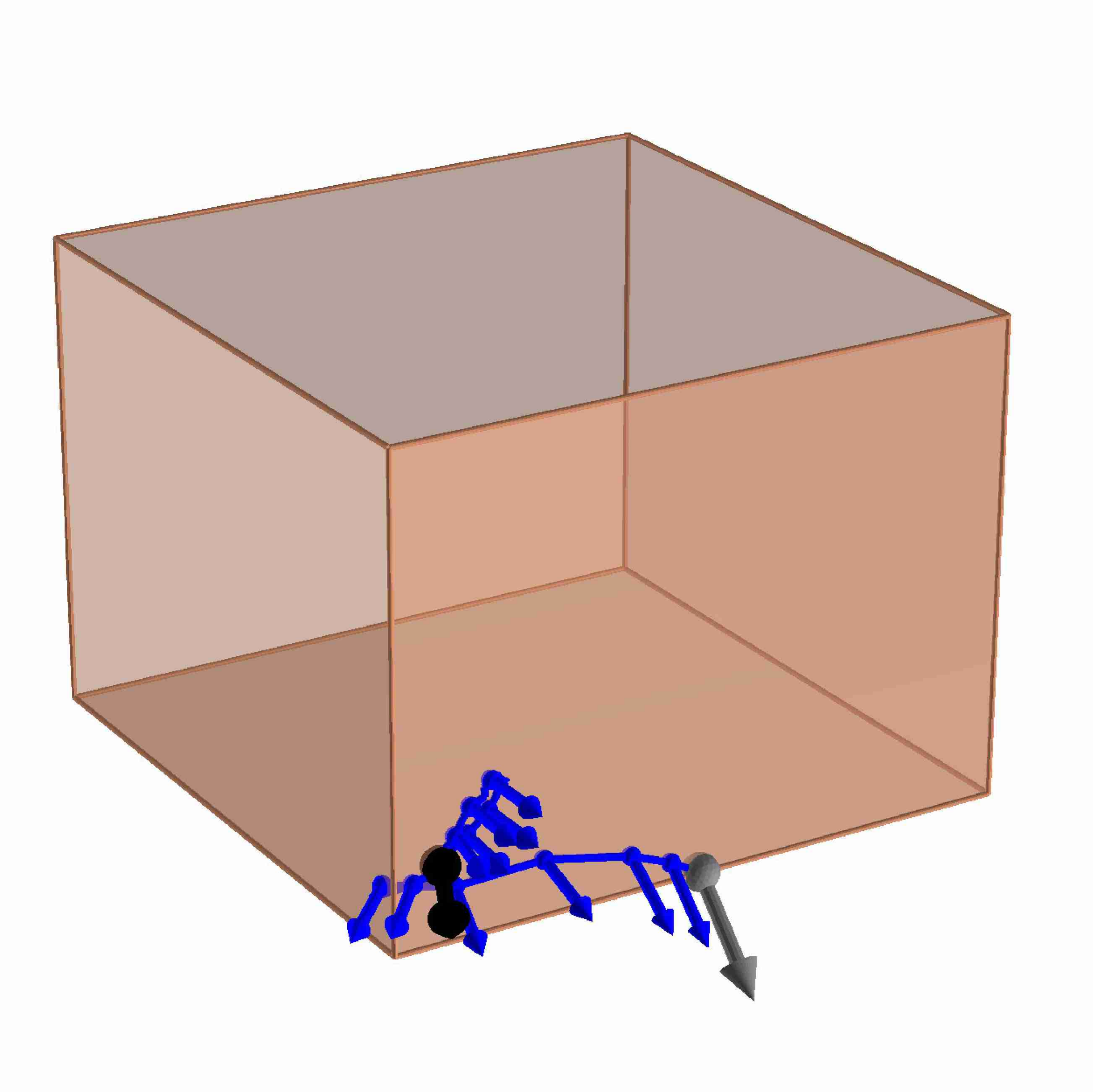}
  \includegraphics[width=0.24\columnwidth]{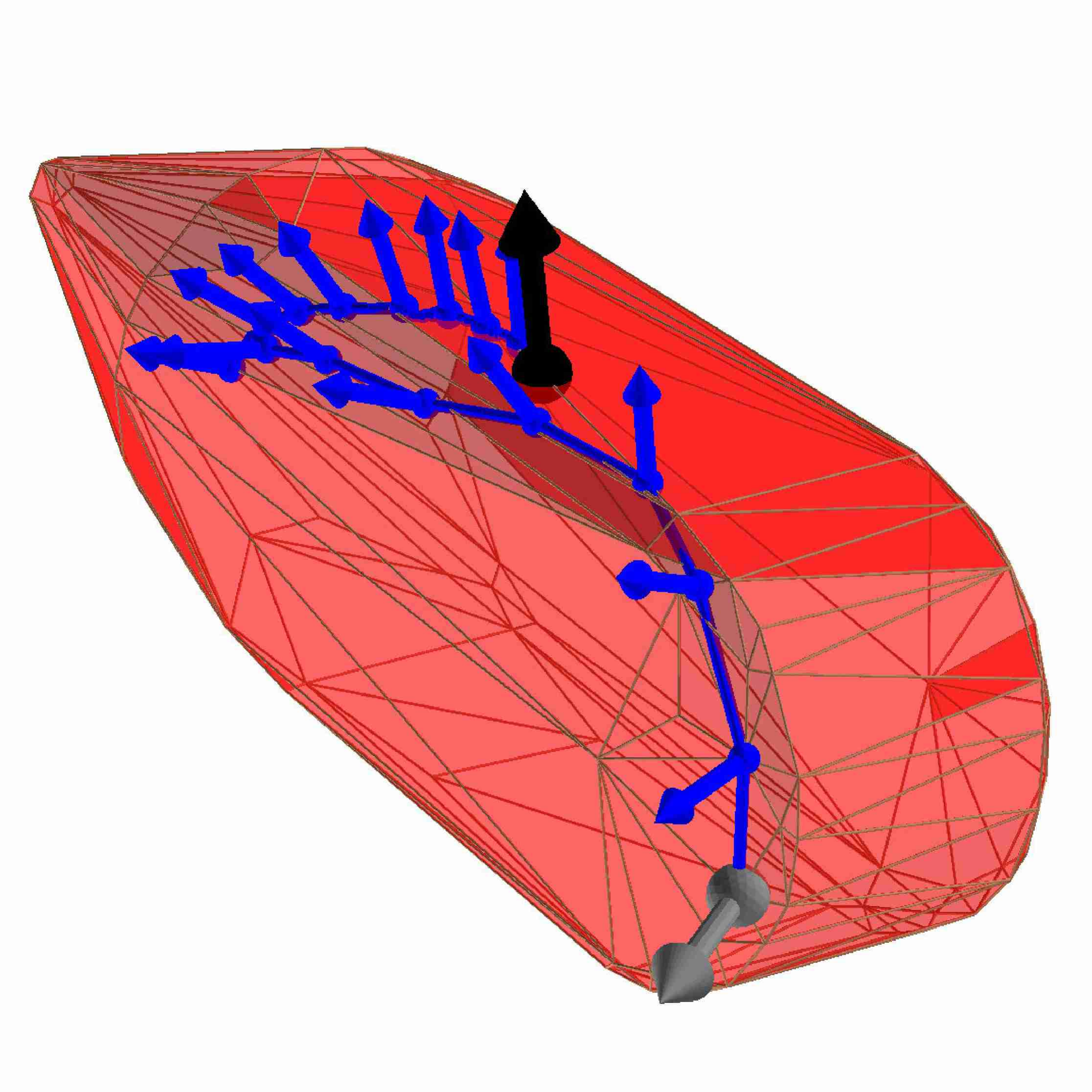}\\
  \vspace{-2.0mm}
  \begin{minipage}{0.49\columnwidth}
    \begin{center} \footnotesize (C) left sholder \end{center}
  \end{minipage}
  \begin{minipage}{0.49\columnwidth}
    \begin{center} \footnotesize (D) left upper arm \end{center}
  \end{minipage}\\
  \vspace{0.5mm}
  \includegraphics[width=0.22\columnwidth]{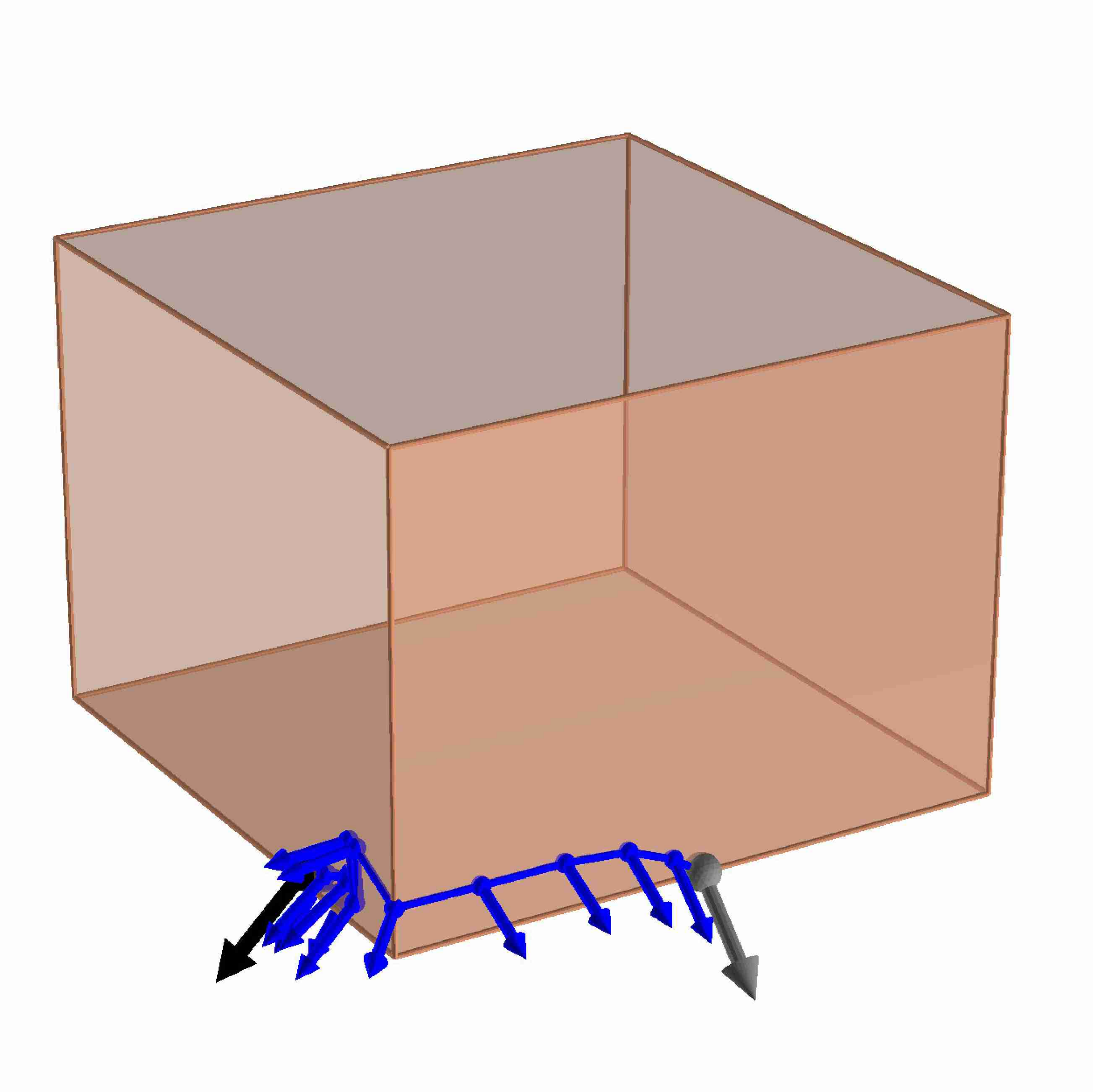}
  \includegraphics[width=0.24\columnwidth]{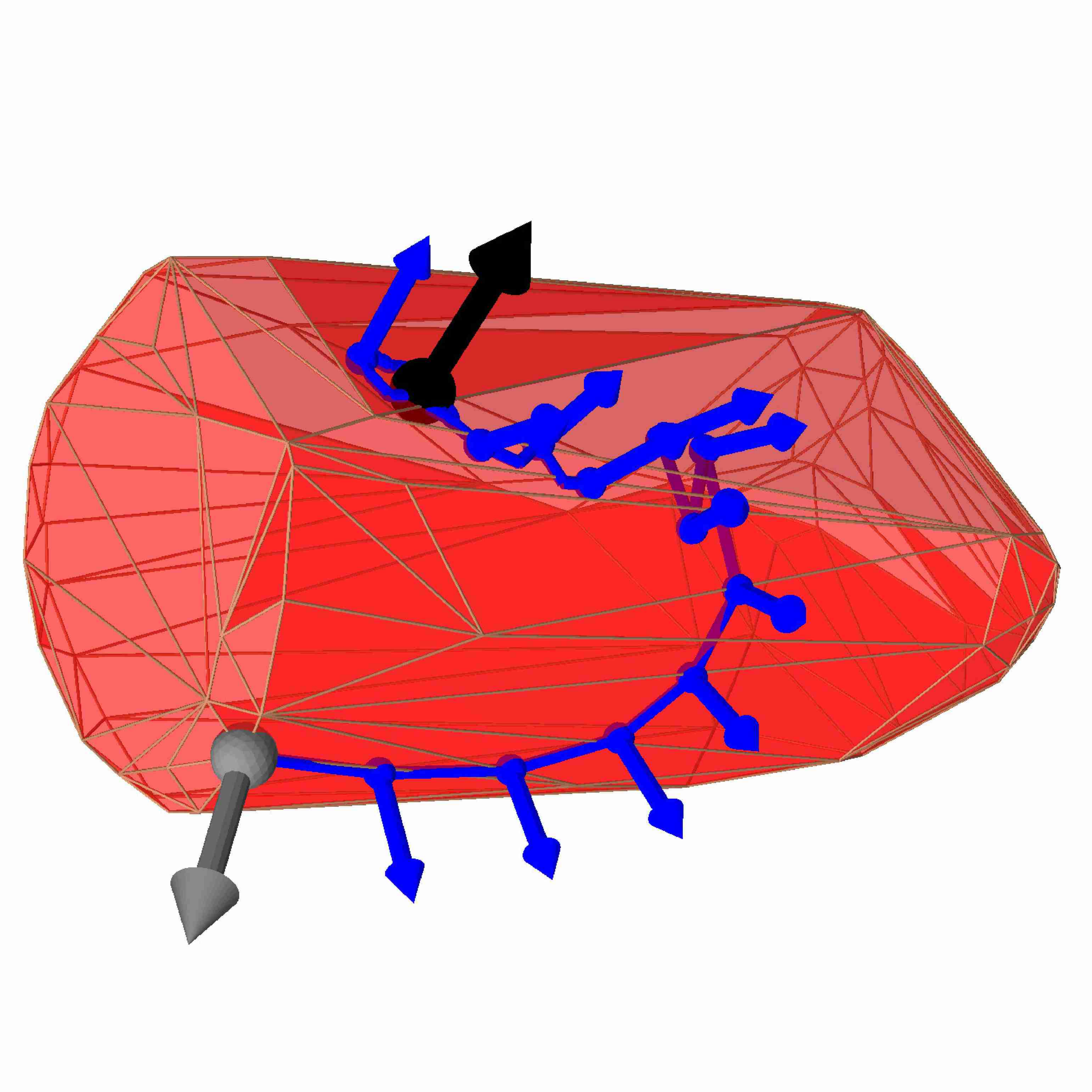}
  \hspace{2mm}
  \includegraphics[width=0.22\columnwidth]{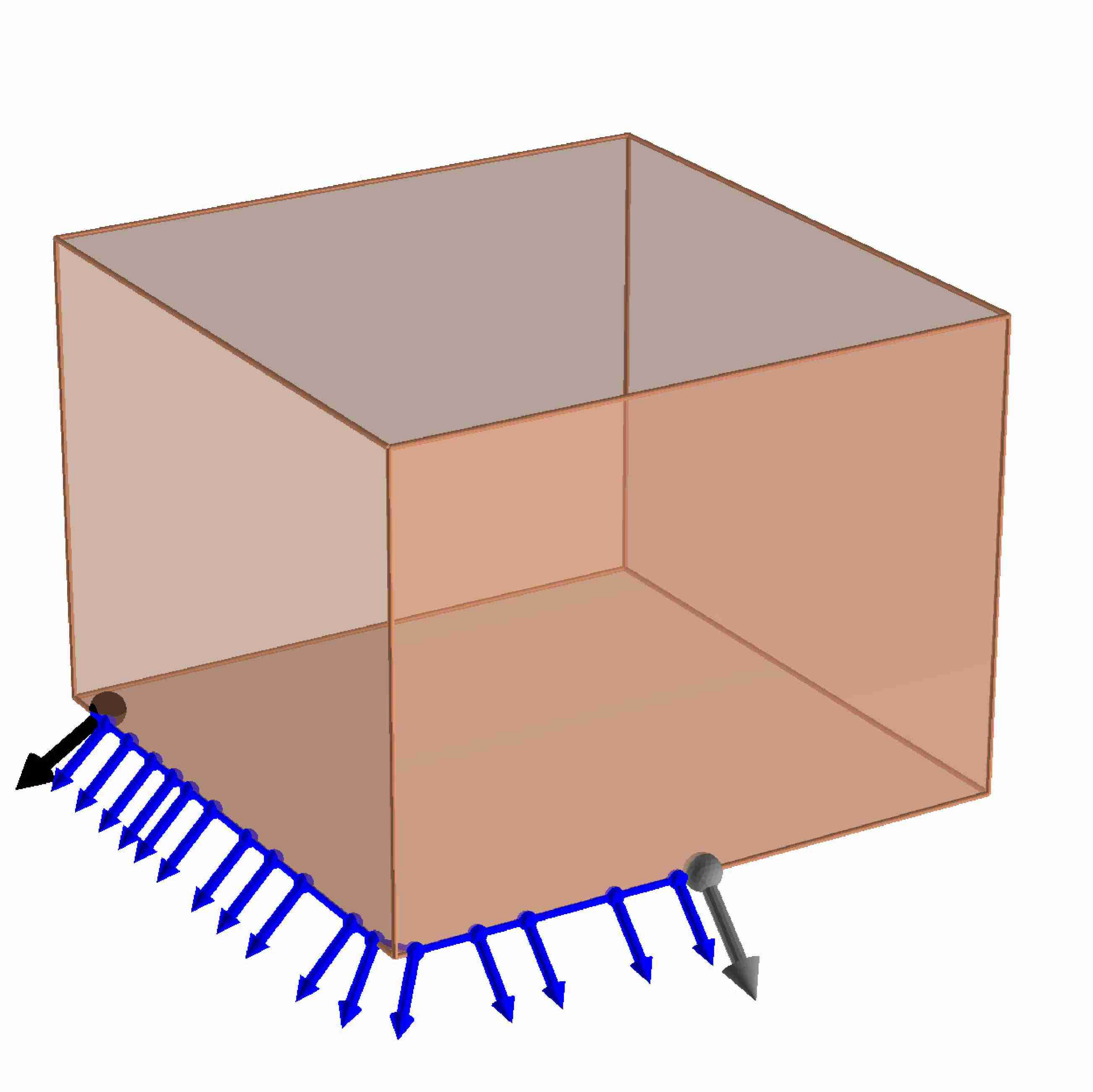}
  \includegraphics[width=0.24\columnwidth]{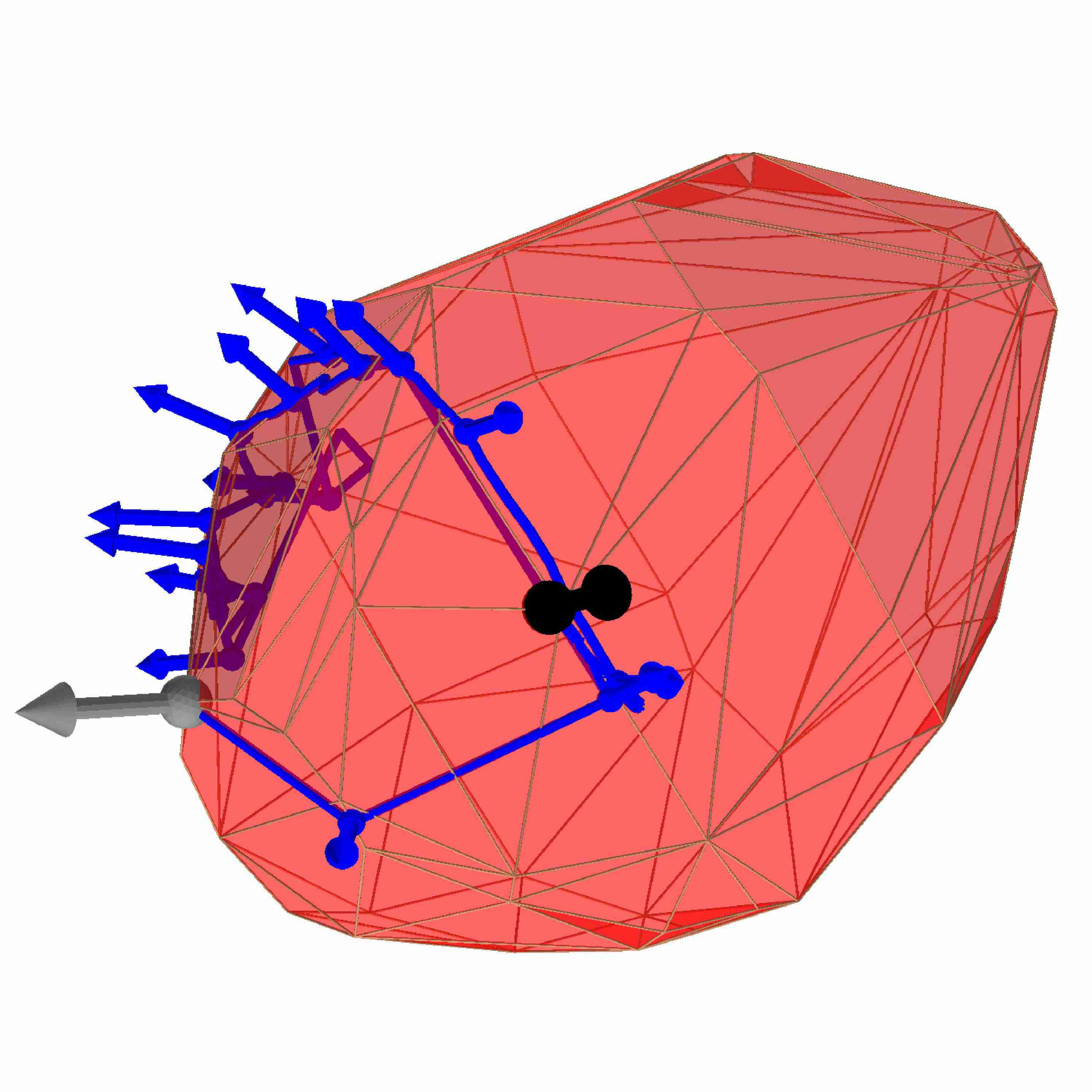}\\
  \vspace{-2.0mm}
  \begin{minipage}{0.49\columnwidth}
    \begin{center} \footnotesize (E) left forearm \end{center}
  \end{minipage}
  \begin{minipage}{0.49\columnwidth}
    \begin{center} \footnotesize (F) left gripper \end{center}
  \end{minipage}
  \caption{History of contact points in posture generation.
    \newline \footnotesize
    The figure shows the history of contact points in optimization for generating the posture of \figref{fig:demo-pr2}.
    Gray points show the initial contact points, and black points show the final contact points,
    and arrows show the smoothed normal.
  }
  \label{fig:demo-pr2-history}
\end{figure}

In this problem,
the joint position $\bm{\theta}$ represents the 17 joint angles and the base link position (x, y, yaw) relative to the object.
The task involves satisfaction of the contacts with the PN-task~\eqref{eq:pn-const} in six pairs of the robot link and object.
Six contact links of dual arms are shown in \figref{fig:demo-pr2-surface} and \figref{fig:demo-pr2-history}.
Since a two-dimensional configuration representing the surface point on the robot link and object for each of the six pairs exists,
the contact configuration $\bm{u}$ comprises 24 dimensions.
Only the lower half of the object surface is specified as the contact region, so that the object is grasped stably.
The contact point of the initial iteration is determined by a non-ad hoc rule; the surface points of the robot link and object closest to the origin of the world frame were selected (\figref{fig:demo-pr2-surface} (A),(C)).
As a constraint, the joint position range and collision avoidance between the robot and the object were specified.
At 200 iterations of the SQP, the computation time was 15~s.
The computational cost was high owing to the many contact points and complexity of the body shape.
As mentioned above,
the experimental program in this paper has been implemented in EusLisp, which is a scripting language;
by implementing it in a compiled language,
it is expected to be considerably faster.
\figref{fig:demo-pr2}-\ref{fig:demo-pr2-history} demonstrate that
appropriate posture and contact points are automatically generated and stably executed by the real robot.

We also generated the motion of receiving an object from a human.
We represent this motion with two postures ($T=2$ in eq.~\eqref{eq:problem-def});
one is a pre-grasping posture that holds a 10\% enlarged virtual object, and the other is a grasping posture.
The base link position is common to the two postures, and both were generated so that they become as close as possible.
The motion was successfully executed, and it was verified that the proposed method was effective for contact-rich motions.

\subsection{Pushing Operation with Knee}
\begin{figure}[thpb]
  \centering
  \includegraphics[height=0.48\columnwidth]{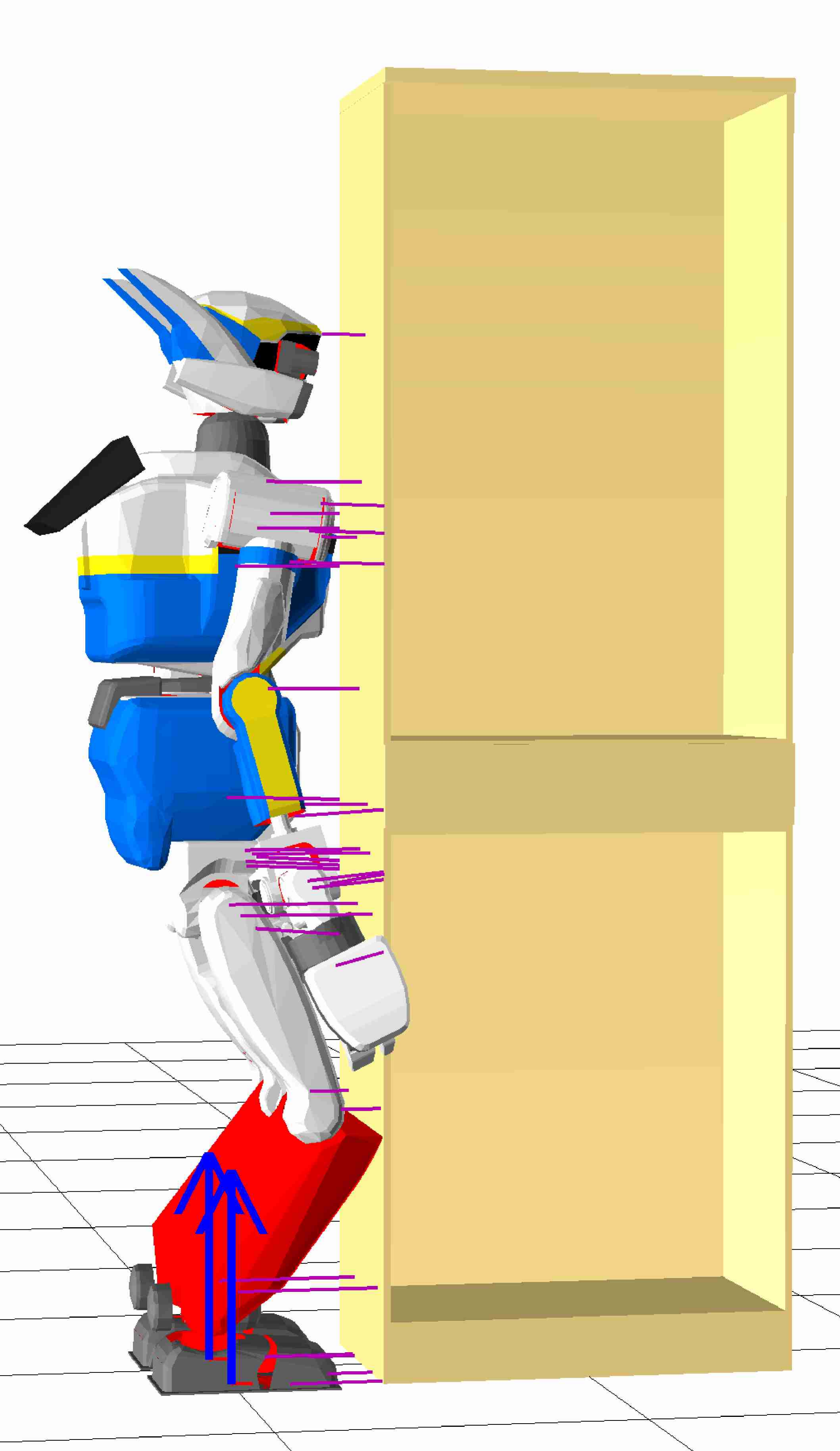}
  \includegraphics[height=0.48\columnwidth]{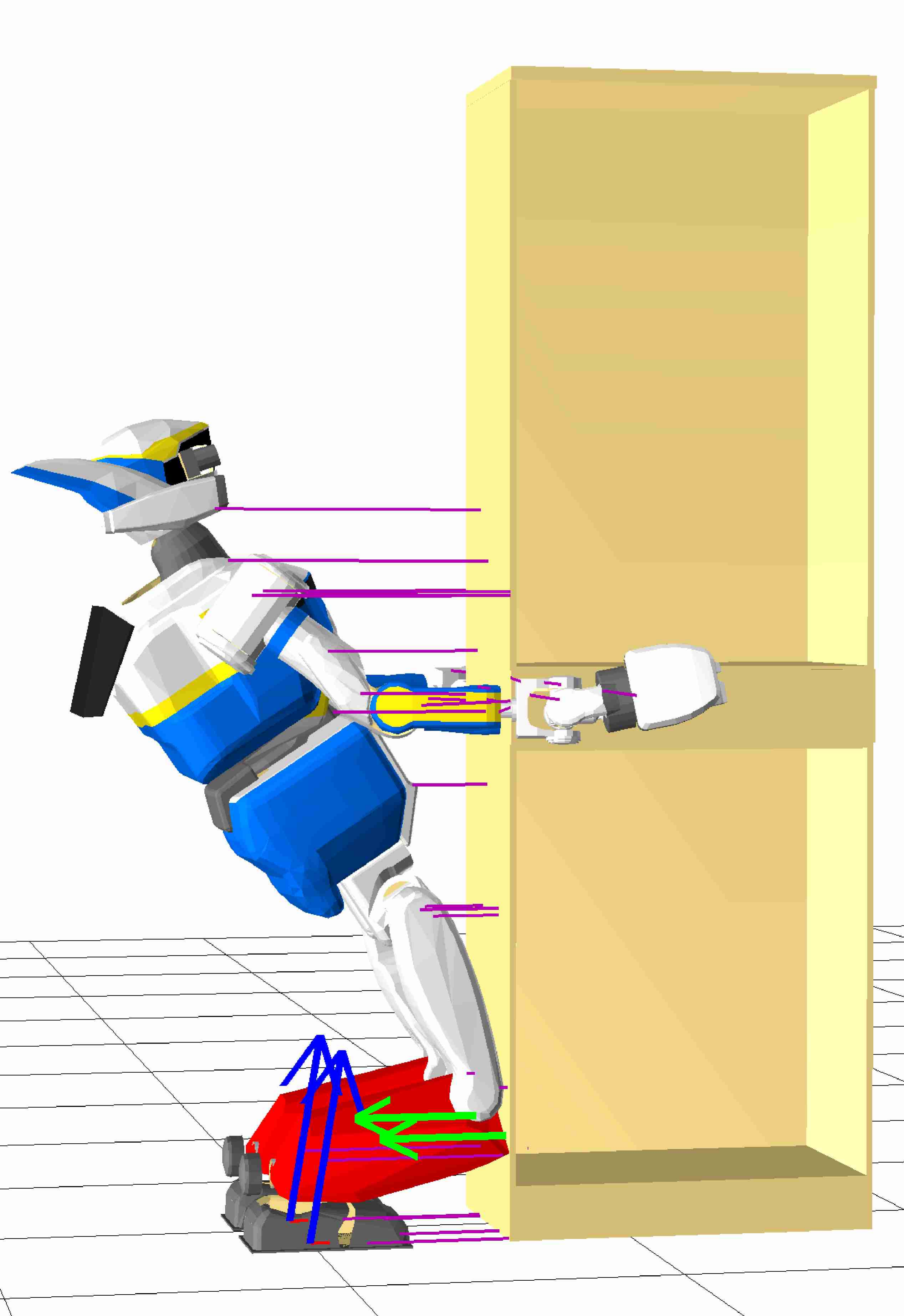}
  \hspace{0.01\columnwidth}
  \includegraphics[height=0.42\columnwidth]{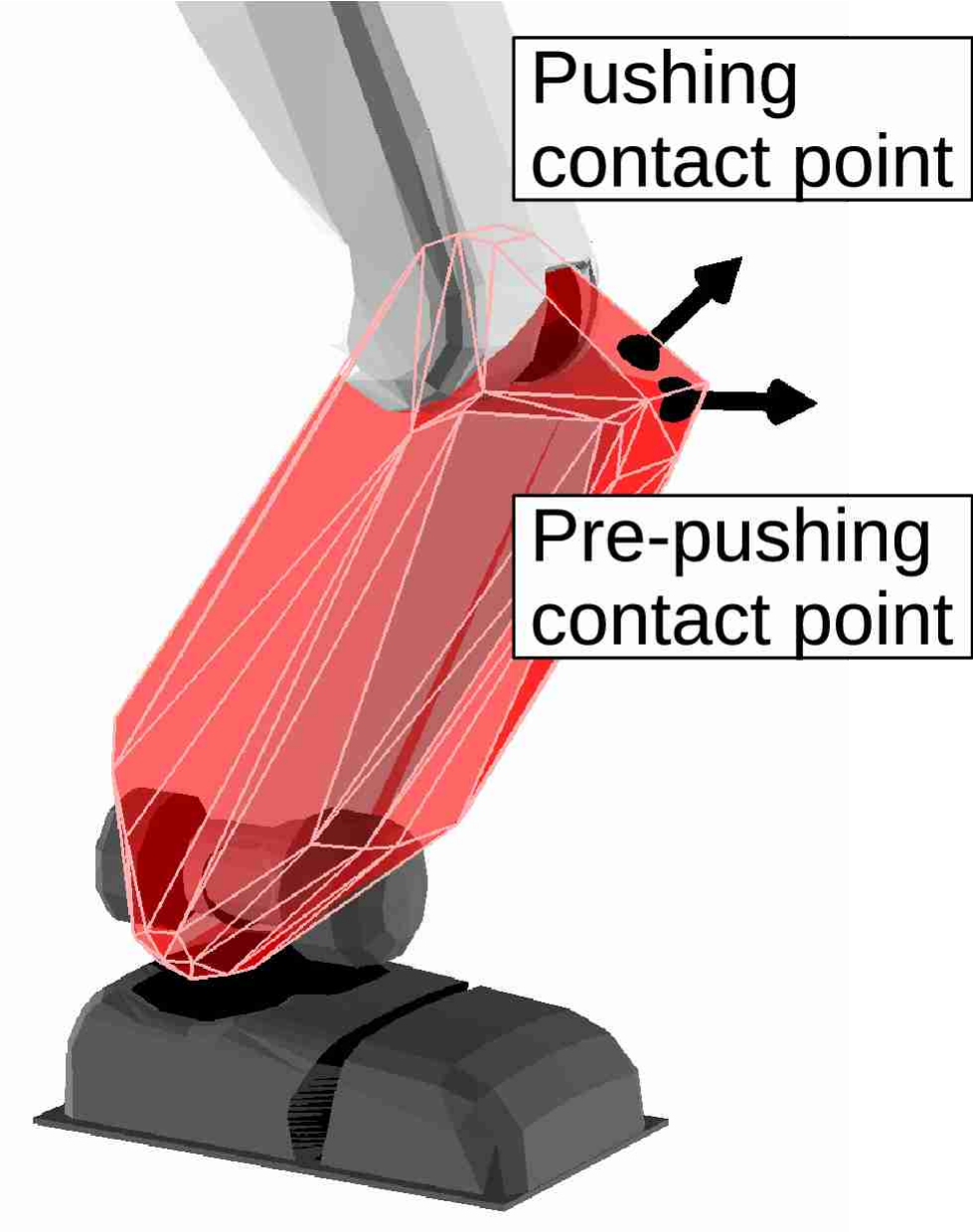}\\
  \vspace{-1mm}
  \begin{minipage}{0.26\columnwidth}
    \begin{center} \footnotesize (A1) \end{center}
  \end{minipage}
  \begin{minipage}{0.34\columnwidth}
    \begin{center} \footnotesize (A2) \end{center}
  \end{minipage}
  \begin{minipage}{0.34\columnwidth}
    \begin{center} \footnotesize (B) \end{center}
  \end{minipage}\\
  \vspace{1mm}
  \includegraphics[height=0.48\columnwidth]{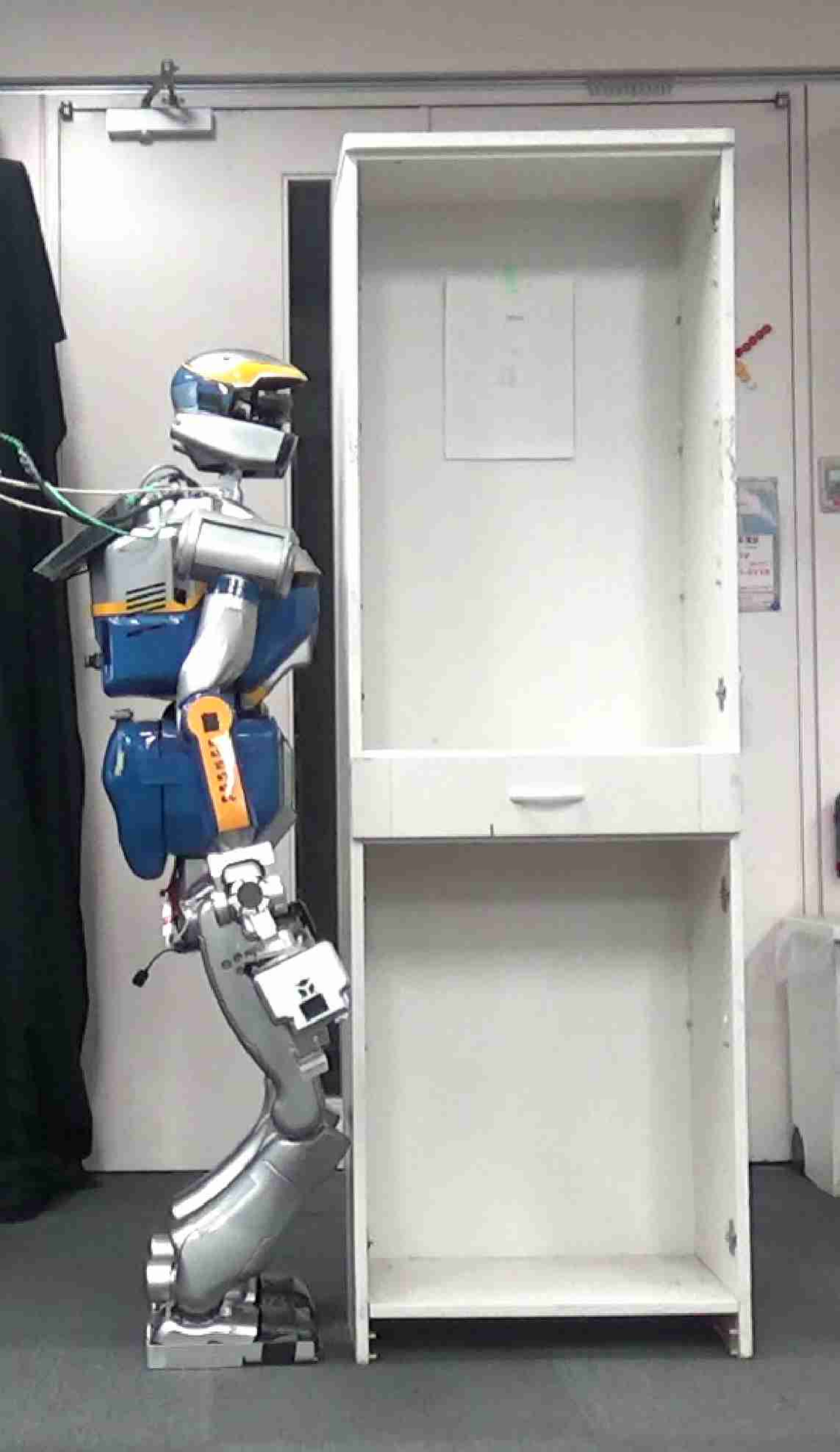}
  \includegraphics[height=0.48\columnwidth]{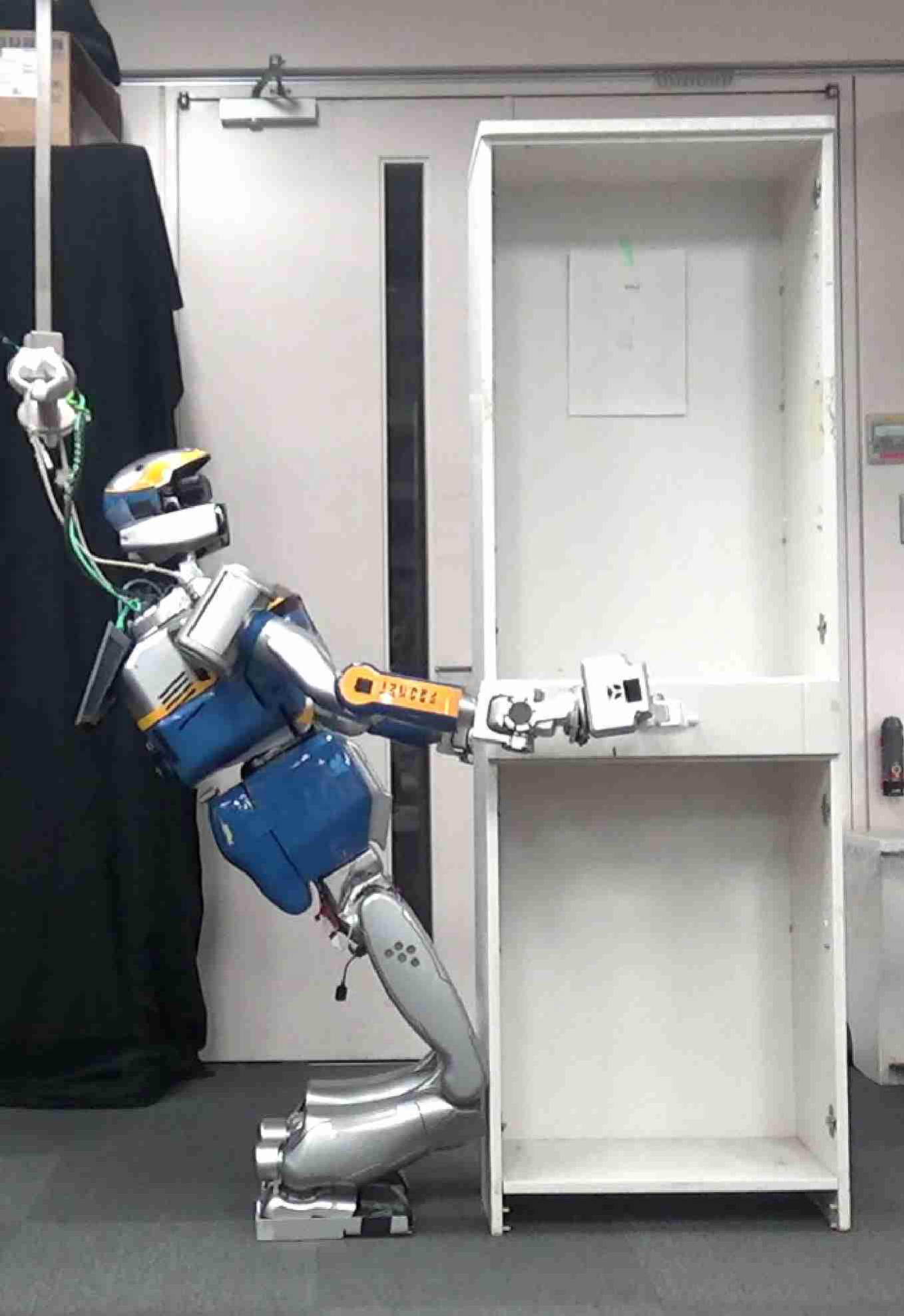}
  \includegraphics[height=0.48\columnwidth]{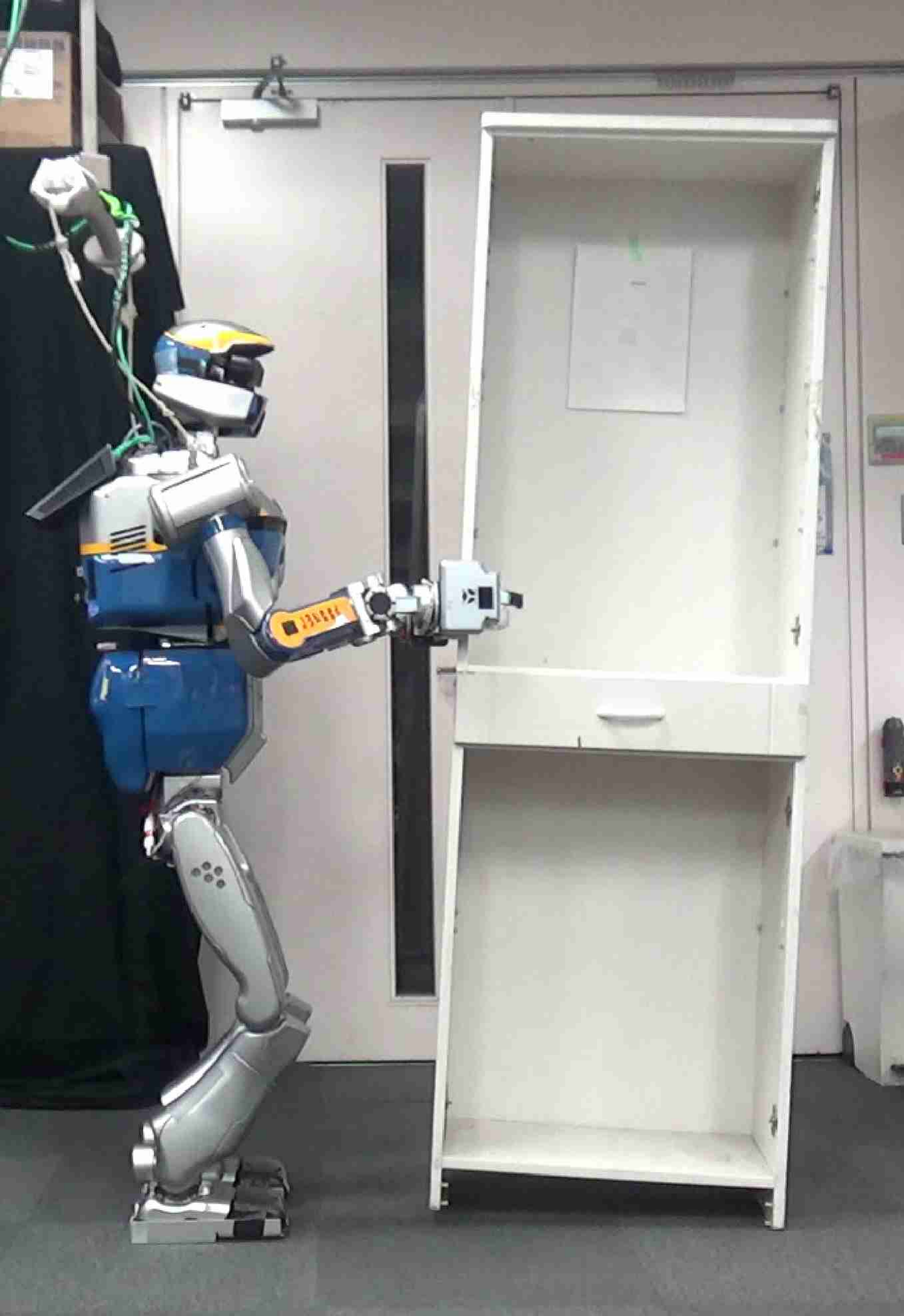}\\
  \vspace{-1mm}
  \begin{minipage}{0.26\columnwidth}
    \begin{center} \footnotesize (C1) \end{center}
  \end{minipage}
  \begin{minipage}{0.34\columnwidth}
    \begin{center} \footnotesize (C2) \end{center}
  \end{minipage}
  \begin{minipage}{0.34\columnwidth}
    \begin{center} \footnotesize (D) \end{center}
  \end{minipage}
  \caption{Pushing operation with knee.
    \newline \footnotesize
    A humanoid robot HRP-2 carries a shelf by pushing with knee.
    (A1,C1) and (A2,C2) show the pre-pushing posture (the knee contacts the shelf but applying no force) and pushing posture, respectively.
    In (A2), the knee pushing force (green arrows) is drawn four times larger than the foot force (blue arrows).
    The purple lines in (A1,A2) show the shortest distance between the collision check bodies.
    In pre-pushing and pushing, an appropriate knee contact point is determined according to the robot's posture as shown in (B).
    The shelf falls when pushed by hands as shown in (D).
  }
  \label{fig:demo-hrp2}
\end{figure}

The proposed method based on inverse kinematics is extended to the framework of inverse kinematics-and-statics \cite{HumanoidPlanning:Bouyarmane:AR2012}
by introducing the contact wrench configuration and wrench equilibrium task to the optimization problem \eqref{eq:opt-def} \cite{GeneratingPosture:Shigematsu:IROS2019}.
By this extended method, the pushing motion by a humanoid robot HRP-2 was generated with consideration of balance.
The target object was a $20$ kg shelf, and because the friction with the floor is high, it falls when pushed by hands (\figref{fig:demo-hrp2} (D)).
Therefore, the robot takes the operation strategy of pushing with the knee, in which, the shelf does not fall because the action point of the pushing force is low and the overturning moment is small (\figref{fig:demo-hrp2} (C2)).
The contact between the knee link and the shelf is a body-to-body contact, and therefore the proposed method can be used for generating a motion.
\figref{fig:demo-hrp2} (A),(C) show the generated motion and the results of the execution.
By repeatedly pushing and walking, the robot was able to carry the shelf forward.
The stabilization control based on the online sensor feedback \cite{Stabilizer:Kajita:IROS2010} was applied to the experiment.

We represent the pushing motion with three postures ($T=3$ in eq.~\eqref{eq:problem-def});
pre-pushing posture without and with an applied force to the shelf, and a pushing posture.
The dimension of $\bm{q}_t$ in eq.~\eqref{eq:config-t} was 62;
of which 32 were related to the joint position,
6 were related to the root link pose,
12 were related to the contact force,
and the remaining 12 were related to the contact configuration.
Each posture was calculated by a separate optimization, taking approximately 8~s for 100 iterations.
The pose of the robot's base link was represented by three linear and three rotational virtual joints between the world frame and the base link.
The motion trajectory was generated by interpolating a sequence of statically stable postures
slowly enough that the dynamic effects could be ignored \cite{DRCSystem:Nozawa:Humanoids2015}.
The pushing force for each knee is $100$ N.
The knee contact is represented as the PN-task~\eqref{eq:pn-const}, and the foot contact is represented as a position and rotation ($\rm{SE}(3)$) task.
\figref{fig:demo-hrp2} (B) shows the contact point on the right knee link.
Since the robot needs to bend the knee for pushing forward, the knee contact point moves on the link surface during pushing.
With the proposed method, automatic determination of the contact point for sliding and rolling motions on the link surface is possible.

\subsection{Non-prehensile Manipulation with Both Hands}
\begin{figure}[thpb]
  \centering
  \includegraphics[width=0.48\columnwidth]{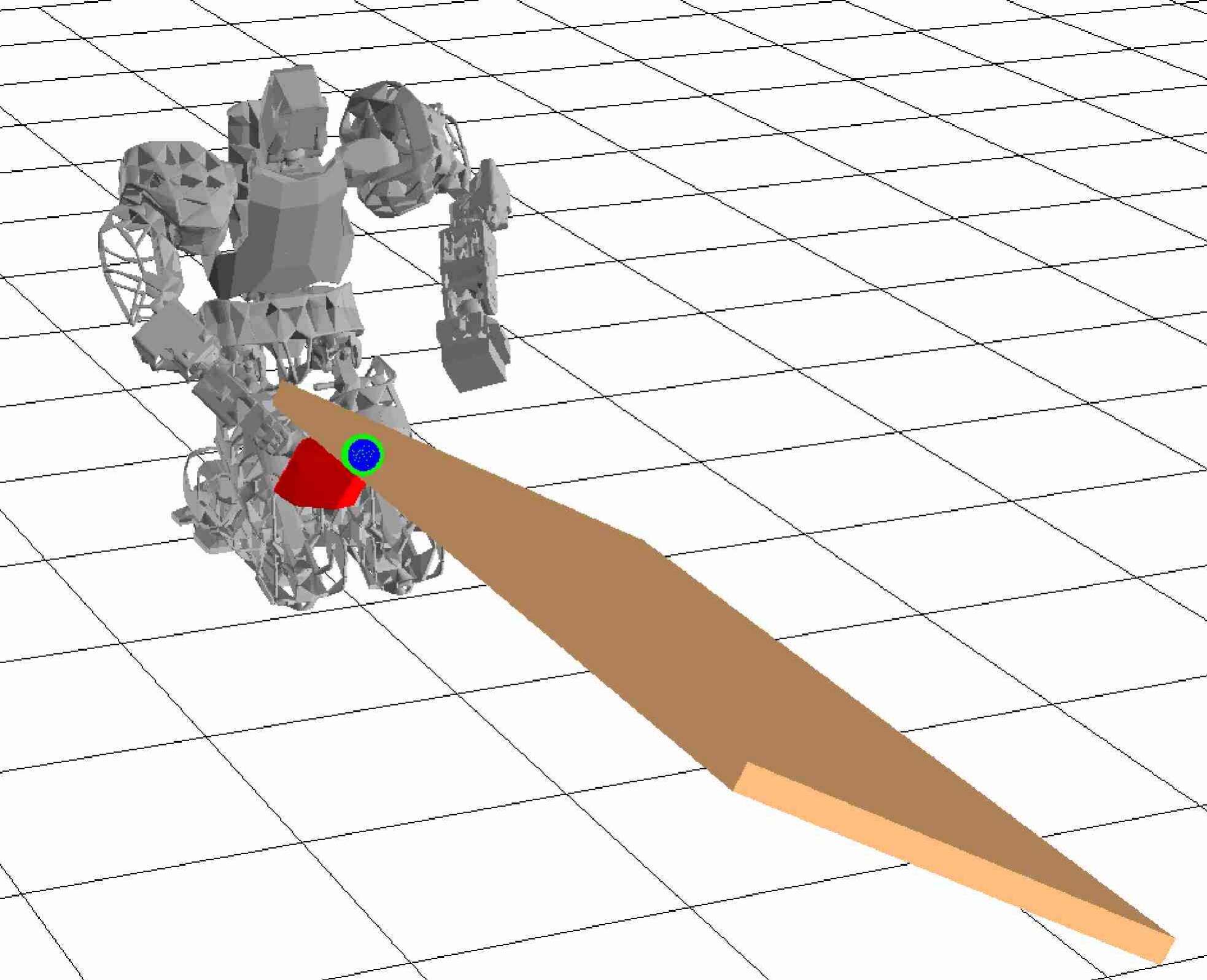}
  \includegraphics[width=0.48\columnwidth]{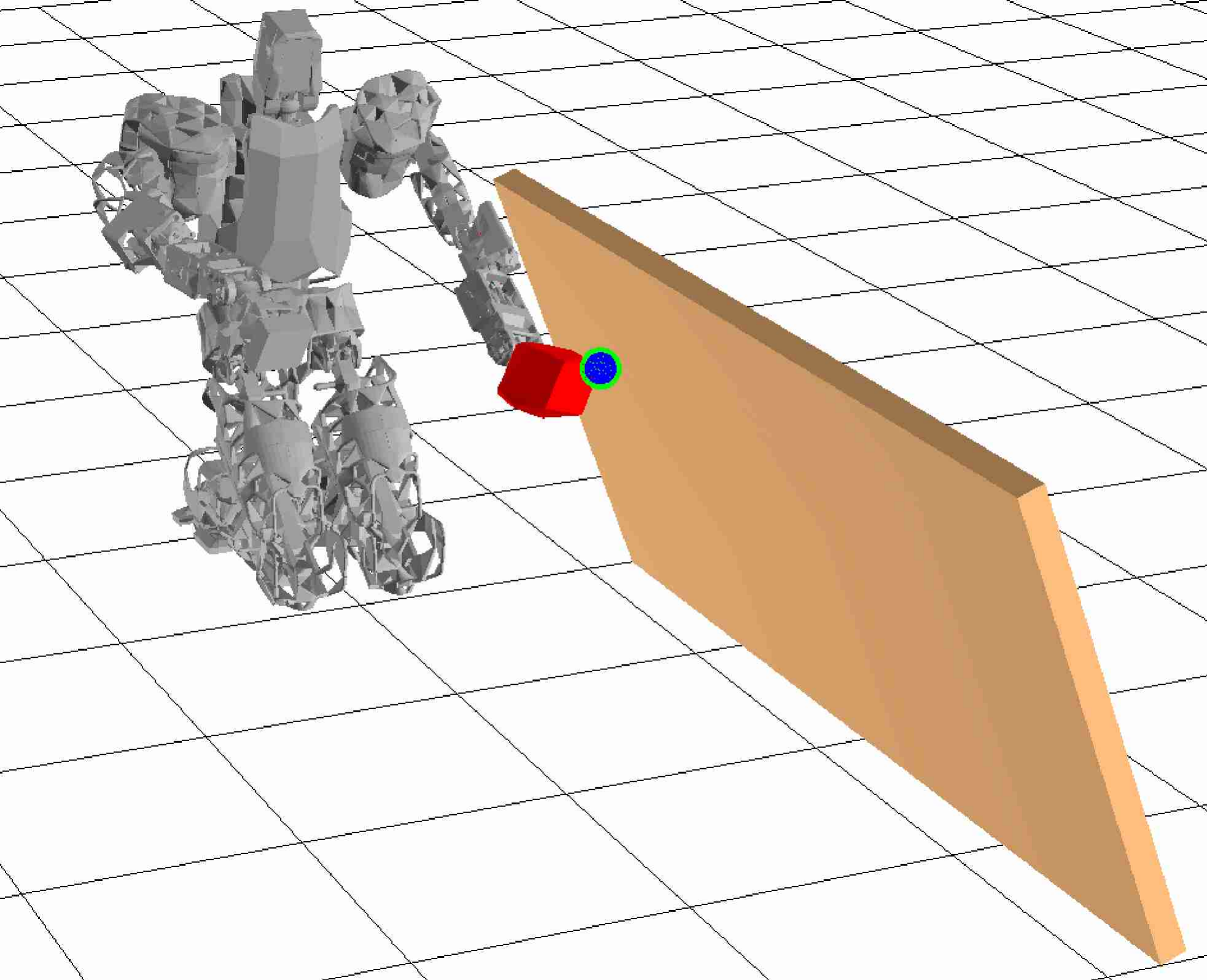}\\
  \vspace{-1mm}
  \begin{minipage}{0.48\columnwidth}
    \begin{center} \footnotesize (A1) \end{center}
  \end{minipage}
  \begin{minipage}{0.48\columnwidth}
    \begin{center} \footnotesize (A2) \end{center}
  \end{minipage}\\
  \vspace{1mm}
  \includegraphics[width=0.48\columnwidth]{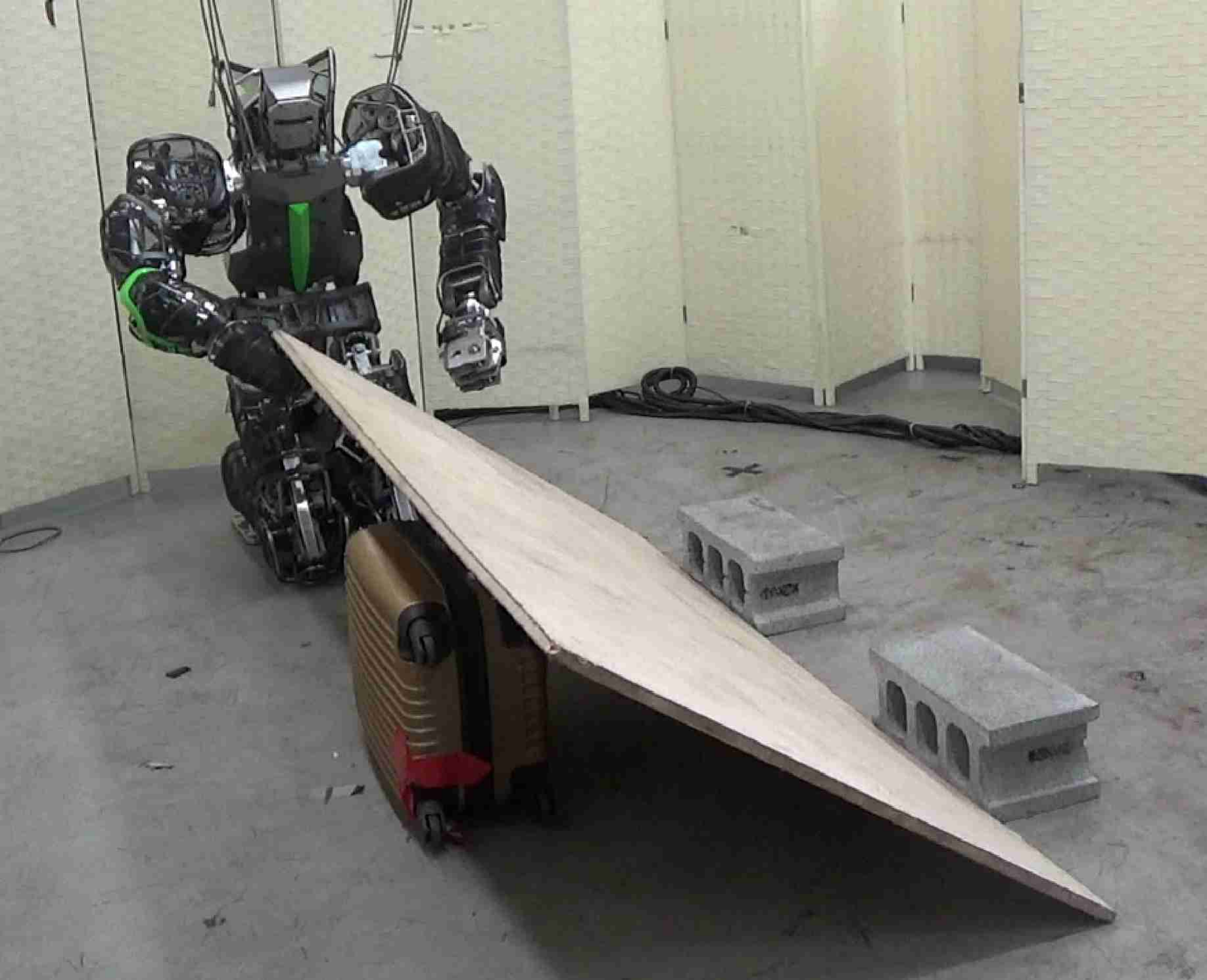}
  \includegraphics[width=0.48\columnwidth]{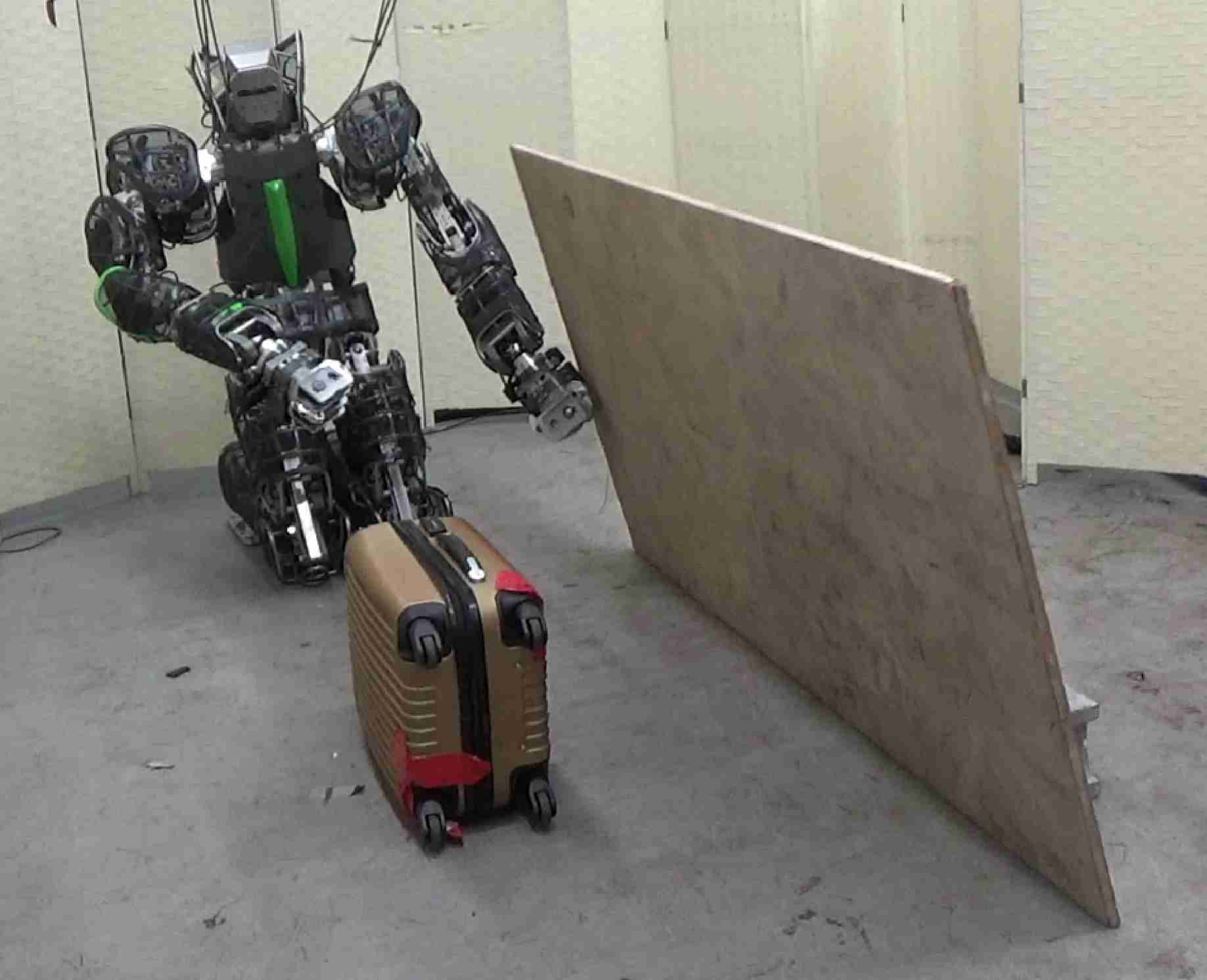}\\
  \vspace{-1mm}
  \begin{minipage}{0.48\columnwidth}
    \begin{center} \footnotesize (B1) \end{center}
  \end{minipage}
  \begin{minipage}{0.48\columnwidth}
    \begin{center} \footnotesize (B2) \end{center}
  \end{minipage}
  \caption{Tilting operation performed on a large board using both hands.
    \newline \footnotesize
    A humanoid robot RHP4B tilts the board
    from 20 degrees to 100 degrees with respect to the horizontal plane.
    In (A1,B1) and (A2,B2), the board is tilted to 40 degrees and 80 degrees, respectively.
    The blue points in (A1,A2) show the contact points between hand and board.
  }
  \label{fig:demo-rhp4b}
\end{figure}

The proposed method is also effective for non-prehensile manipulation \cite{NonprehensileManip:Mason:IJRR1999} in which the object is not completely grasped.
\figref{fig:demo-rhp4b} shows the humanoid robot RHP4B \cite{RobustHumanoid:Kakiuchi:IROS2017} tilting a large board from a kneeling posture,
which is an example of non-prehensile manipulation where an object is supported by both the environment and robot.

The motion was generated as a sequence of 12 postures ($T=12$ in eq.~\eqref{eq:problem-def});
the board was tilted from 20 degrees to 100 degrees in steps of 10 degrees with respect to the horizontal plane,
and a switch from right-hand contact to left-hand contact was performed at an inclination of 70 degrees to the horizontal.
The dimension of $\bm{q}_t$ in eq.~\eqref{eq:config-t} was 66;
of which 32 were related to the joint position,
6 were related to the base link pose,
24 were related to the contact force,
and remaining 4 were related to the contact configuration.
Each posture was calculated by a separate optimization, taking approximately 3~s for 30 iterations.
The contact point between the hand and the board, allowed to move freely on the surface during manipulation, was determined by optimizing the contact configuration.
Because the manipulation required a wide reach range, no valid sequence of motion steps could be generated in the case of a fixed contact point.

\section{Conclusions}

In this study,
we proposed the posture generation method to determine the joint positions and contact points simultaneously.
Various experiments on simulation and real robots showed the effectiveness of the optimization
with the local contact configurations and normal smoothing of the body surface.
As future work,
the proposed method will be extended for the dynamic motions with the continuous contact transitions (e.g., sliding and rolling).



\bibliographystyle{junsrt}
\bibliography{main}

\end{document}